\documentclass[conference,onecolumn]{IEEEtran}
\IEEEoverridecommandlockouts

\usepackage{cite}
\usepackage{xcolor}
\usepackage{amssymb}
\usepackage{amsmath}
\allowdisplaybreaks[4]
\usepackage{multicol}
\usepackage{amsfonts}
\usepackage{amsthm}
\usepackage{array}
\usepackage{bbm}
\usepackage[caption=false,font=normalsize,labelfont=sf,textfont=sf]{subfig}
\usepackage{textcomp}
\usepackage{stfloats}
\usepackage{url}
\usepackage{verbatim}
\usepackage{graphicx}
\usepackage{multirow}
\usepackage{textcase}

\usepackage[lined,linesnumbered,ruled]{algorithm2e}

\usepackage{tikz}
\newcommand*\circled[1]{\tikz[baseline=(char.base)]{
            \node[shape=circle,draw,inner sep=1pt] (char) {#1};}}

\newtheorem{lemma}{Lemma}
\newtheorem{fact}{Fact}

\newtheorem{theorem}{Theorem}

\newtheorem{definition}{Definition}
\newtheorem{assumption}{Assumption}
\newtheorem{corollary}{Corollary}

\begin{document}

\title{PDSL: Privacy-Preserved Decentralized Stochastic Learning with Heterogeneous Data Distribution}

\author{
\IEEEauthorblockN{Lina Wang$^a$, Yunsheng Yuan$^a$, Chunxiao Wang$^{b,c}$, Feng Li$^{a,b}$}
\IEEEauthorblockA{$^a$ School of Computer Science and Technology, Shandong University, Qingdao, China}
\IEEEauthorblockA{$^b$ Key Laboratory of Computing Power Network and Information Security, Ministry of Education, \\Shandong Computer Science Center (National Supercomputer Center in Jinan), \\Qilu University of Technology (Shandong Academy of Sciences), Jinan, China}
\IEEEauthorblockA{$^c$ Shandong Provincial Key Laboratory of Computing Power Internet and Service Computing, \\Shandong Fundamental Research Center for Computer Science, Jinan, China}
\IEEEauthorblockA{Email: linawang425@mail.sdu.edu.cn, ysyuan1028@mail.sdu.edu.cn, wangchx@sdas.org, fli@sdu.edu.cn}
}

\maketitle

\begin{abstract} 
  In the paradigm of decentralized learning, a group of agents collaborates to learn a global model using distributed datasets without a central server. However, due to the heterogeneity of the local data across the different agents, learning a robust global model is rather challenging. Moreover, the collaboration of the agents relies on their gradient information exchange, which poses a risk of privacy leakage. In this paper, to address these issues, we propose PDSL, a novel privacy-preserved decentralized stochastic learning algorithm with heterogeneous data distribution. On one hand, we innovate in utilizing the notion of Shapley values such that each agent can precisely measure the contributions of its heterogeneous neighbors to the global learning goal; on the other hand, we leverage the notion of differential privacy to prevent each agent from suffering privacy leakage when it contributes gradient information to its neighbors. We conduct both solid theoretical analysis and extensive experiments to demonstrate the efficacy of our PDSL algorithm in terms of privacy preservation and convergence.
\end{abstract}


\section{Introduction} \label{sec:intro}
  Distributed machine learning refers to a family of algorithms aiming at learning from data distributed among multiple agents~\cite{VerbraekenWKKVR-CSUR21, McMahanMRHA-AISTATS17, ShamsabadiGHC-TIFS20}. Based on communication topology, approaches to designing distributed learning algorithms include \textit{centralized learning} and \textit{decentralized learning}. As a popular approach for centralized learning, \textit{Federated Learning} (FL) enables the central server to organize the collaborative training process through the interaction of model parameters while training data can be stored locally at the agents~\cite{YangLCT-TIST19, JinRZLLD-ICML22}. However, this setup requires constant communication with the central server, potentially creating a bottleneck. To mitigate this concern, some decentralized learning algorithms have been proposed such that agents communicate their updates with their neighbors in a communication network without the assistance of a central server.

  One of the fundamental challenges for decentralized learning is data heterogeneity \cite{LiDCH-ICDE22}. For example, due to diverse users behaviors, data across different agents are distributed non-independently and identically (Non-IID), while such a Non-IID data distribution may result in a degradation in terms of convergence and accuracy~\cite{LiSZSTS-MLSys20}. Although the problem of data heterogeneity has been extensively investigated for centralized learning (e.g., FL)~\cite{LuoCHZLF-NeurIPS21, SunLZXLLQR-SIGKDD23, TanLLJLZ-AAAI23, ChenL-INFOCOM22}, there are only a very handful of studies on decentralized learning. Recently, decentralized learning algorithms based on cross-gradient information have been proposed in \cite{EsfandiariTJBHHS-ICML21, AketiKR-TMLR23}. Each agent aggregates cross-gradient information, i.e., derivatives of its model with respect to its neighbors’ datasets, to update its local model. \cite{ZhangFLYLZ-MobiHoc22} incorporates a recursive gradient correction technique to address data heterogeneity. Unfortunately, these algorithms cannot fairly evaluate the contribution of each agent.

  Another significant challenge in decentralized learning is privacy preservation. Decentralized learning entails information exchange among the agents. For example, in \cite{EsfandiariTJBHHS-ICML21, AketiKR-TMLR23}, each agent calculates cross-gradients for each of its neighbors based on its local dataset, while numerous recent studies have shown that sharing the cross-gradient information can potentially result in the leakage of private local datasets~\cite{ShokriSSS-SP17,FredriksonJR-CCS15,ZhuLH-NeurIPS19}. To tackle this problem, one popular choice is \textit{Differential Privacy} (DP). For example, \cite{XuZW-TPAMI22} incorporates DP into an asynchronous decentralized parallel SGD framework. \cite{CyffersEBM-NeurIPS22} proposes a differentially private decentralized optimization algorithm that alternates between local gradient descent and gossip averaging. Unfortunately, these state-of-the-art proposals do not take into account the heterogeneous data distribution.

  In this paper, we propose PDSL, a novel privacy-preserved decentralized stochastic learning algorithm with heterogeneous data distribution. Specifically, in each round of our PDSL algorithm, each agent first computes its local gradient and cross-gradient, i.e., the derivative of its local model the ones of its neighbors' local models with respect to its local datasets. The agent then adds noise to obtain the perturbed gradients based on DP mechanism. Next, each agent exchanges the perturbed cross-gradient information with its neighbors. Finally, each agent updates its model in a momentum-like manner based on the received perturbed cross-gradient and its perturbed local gradient. To properly weight the perturbed gradients in the aggregation, each agent utilizes Shapley values to fairly quantify the contributions of its perturbed local gradient and the received perturbed cross-gradient. In summary, our main contributions are as follows:
  \begin{itemize}
    \item We propose PDSL, an innovative and robust decentralized learning algorithm with DP to address heterogeneous data distribution and privacy leakage risks.
    \item We conduct rigorous theoretical analysis to prove the DP guarantee of our algorithm and verify its convergence under the DP framework.
    \item We carry out an extensive experimental assessment of PDSL across diverse real-world datasets. The outcomes validate both its practical viability and efficacy.
  \end{itemize}

  The remaining of this paper is organized as follows. Related literature and preliminaries are first introduced in Sec.~\ref{sec:relwork} and Sec.~\ref{sec:pre}, respectively. We then present our PDSL algorithm in detail in Sec.~\ref{sec:pdsl}. Furthermore, we conduct solid theoretical analysis on the privacy protection and convergence of our PDSL algorithm in Sec.~\ref{sec:analysis}. We also perform extensive experiments to verify the efficacy of our PDSL algorithm in Sec.~\ref{sec:exp} and finally conclude our paper in Sec.~\ref{sec:con}.

\section{Related Work} \label{sec:relwork}
  The paradigm of decentralized learning has been attracting increasing attention in recent years. \cite{LianZZHZL-NeurIPS17} integrates \textit{Stochastic Gradient Descent} (SGD) with gossip-averaging as proposed in \cite{BoydGPS-TIT06} to develop the \textit{Decentralized Parallel Stochastic Gradient Descent} (D-PSGD) algorithm. \cite{ScamanBBML-NeurIPS18} considers the distributed optimization of non-smooth convex function and proposes the \textit{Multi-Step Primal-Dual} (MSPD) algorithm for non-smooth decentralized optimization. Moreover, \cite{YuJY-ICML19} proposes the momentum version of D-PSGD, namely \textit{Decentralized Momentum Stochastic Gradient Descent} (DMSGD) algorithm. \cite{YeLL-JSTSP22} presents Soft-DSGD for communication networks featuring lightweight and unreliable communication protocols.
  
  

  These studies usually assume that data is independently and identically distributed across different agents. However, this assumption may not always hold in real-world applications. Therefore, there exist some state-of-the-art proposals that considering the challenge of data heterogeneity. \cite{LinKSJ-ICML21} combines quasi-global momentum with local stochastic gradients to mitigate the difficulties of decentralized learning on heterogeneous data. \cite{LiLV-IOTJ21} incorporates \textit{Mutual Knowledge Transfer} (MKT) into decentralized learning to alleviate the negative impact of data heterogeneity. \cite{ShiSWSYWT-ICML23} indicates that model inconsistency caused by heterogeneous data and network topology in decentralized FL may result in over-fitting in local models and performance degradation. Therefore, DFedSAM and DFedSAM-MGS are proposed in \cite{ShiSWSYWT-ICML23} to enhance the performance of decentralized learning. Specifically, DFedSAM employs \textit{Sharpness Aware Minimization} (SAM) to generate flat model in each client, DFedSAM-MGS further improves DFedSAM by utilizing \textit{Multiple Gossip Steps} (MGS) for better model consistency. 
  Moreover, various tracking approaches, such as gradient tracking~\cite{PuA-MP21} and momentum tracking~\cite{TakezawaBNSY-TMLR23}, have been introduced to address the issue of heterogeneous data in decentralized learning. \cite{AketiHR-NeurIPS24} introduces \textit{Global Update Tracking} (GUT) algorithm which applies the tracking mechanism with respect to the model updates instead of the gradients. Recently, \cite{EsfandiariTJBHHS-ICML21} and \cite{AketiKR-TMLR23} adopt the concept of cross-gradients to reduce the impact of heterogeneous data distribution in decentralized learning. Specifically, \cite{EsfandiariTJBHHS-ICML21} collects the gradient information of each agent's model on all its neighbors' datasets and uses quadratic programming to project them into a single gradient which is then used to update the model. In \cite{AketiKR-TMLR23}, each agent aggregates two types of cross-gradients, i.e., the derivatives computed locally using its local data on the neighbors’ model parameters, and the gradients of its local model with respect to its neighbors’ datasets.

  While decentralized learning algorithms enable agents to avoid sharing raw data through local training and exchanging model parameters or gradients, \cite{ShokriSSS-SP17, MelisSDS-SP19, ZhuLH-NeurIPS19} have shown that malicious adversaries can infer private information from them. \textit{Differential privacy} (DP) \cite{DworkR-FTTCS14} has been widely used to prevent privacy leakage in machine learning. Despite extensive studies in centralized distributed learning (e.g., FL) \cite{ChenDBZJ-TKDE24, LinWLSHJ-TDSC23}, only a few works investigates the application of DP in decentralized learning algorithms. \cite{HuangG-arXiv20} applies DP to ADMM-based decentralized updates. \cite{CyffersEBM-NeurIPS22} proposes Muffliato, a privacy amplification mechanism consisting of local Gaussian noise injection at the agent level followed by a gossiping process for averaging private values. \cite{XuZW-TPAMI22} presents a DP-based asynchronous decentralized parallel SGD. 
  However, these proposals do not consider the data heterogeneity across the agents. In this paper, we leverage the notion of Shapley value to design a DP-based decentralized learning algorithm on heterogeneous data distribution.

\section{Preliminaries} \label{sec:pre}
  We first introduce the basics of decentralized learning and DP in Sec.~\ref{ssec:dsl} and Sec.~\ref{ssec:dp}, respectively. We then introduce the concept of Shapley value in Sec.~\ref{ssec:shapley}.

  \subsection{Decentralized Stochastic Learning} \label{ssec:dsl}
    Consider a set of agents $ \mathcal{M} = \{ 1,2, \cdots, M\}$. Each agent $i \in \mathcal{M}$ owns a private local dataset $\mathcal{D}_i$. The aim of decentralized stochastic learning is to let the agents collaborate to train a global model with their local datasets. Specifically, the objective can be formulated as follows:
    \begin{align}  \label{eq:decentstochprob}
      \min_{x \in \mathbb{R}^d} & \mathcal{F}(x) := \frac{1}{M} \sum^M_{i=1} f_i(x)
    \end{align}
    where $x \in \mathbb{R}^d$ denotes model parameter, and 
    \begin{align} \label{eq:localloss}
      f_i (x) = \mathbb{E}_{\xi_i \sim \mathcal{D}_i} \left[ F_i (x; \xi_i) \right]
    \end{align}
    is the expectation of local loss function $F_i$ associated with agent $i$ on its local data distribution $\mathcal{D}_i$.

    We assume the agents communicate with each other over a graph $\mathcal{G} = ( \mathcal{M}, \mathbf{W} )$, where $\mathbf{W} \in [0,1]^{M \times M}$ is the weighted adjacent matrix of $\mathcal G$. We suppose $\mathbf{W}$ is a symmetric doubly stochastic matrix such that $\sum_{j\in\mathcal{M}} \omega_{i,j} = \sum_{i\in\mathcal{M}} \omega_{i,j} = 1$ and $\omega_{i,j} = \omega_{j,i}, \forall i, j \in \mathcal{M}$, where $\omega_{i,j}$ denotes the $(i,j)$-th element of $\mathbf{W}$. Let $\mathcal{M}_i \subseteq \mathcal{M}$ be the set of neighbors of agent $i$ including itself, i.e., $\mathcal{M}_i = \{ j \in \mathcal{M} \mid \omega_{i,j} > 0 \}$. $\omega_{i,j}$ is known for agent $i$ for any $j \in \mathcal{M}_i$.
    
    In our decentralized learning algorithm, we employ a validation dataset $\mathcal{Q}$ to address the problem of data heterogeneity. Specifically, each agent measures the data relevances (or contributions) of its local model and the ones of its peers to the global training goal. Let $\mathcal{Q}$ denote the global validation dataset, which is loaded in each agent in advance.

  \subsection{Differential Privacy} \label{ssec:dp}
    In this paper, we leverage the notion of \textit{Differential Privacy} (DP) for the purpose of privacy preservation. It is defined in terms of adjacent datasets. Specifically, two datasets are said to be adjacent if they differ in a single entry. A formal definition of DP is given in \textbf{Definition}~\ref{def:dp}.
    \begin{definition}[$(\epsilon, \delta)$-Differential Privacy] \label{def:dp}
      A randomized mechanism $\mathcal{A}: \mathsf{D} \rightarrow \mathsf{R}$ satisfies $(\epsilon, \delta)$-differential privacy, if for any two adjacent datasets $\mathcal{D}, \mathcal{D}' \in \mathsf{D}$ and any subset of outputs $\mathcal{S} \subseteq \mathsf{R}$, it holds that
      \begin{align}  \label{eq:dp-condition}
        \mathrm{Pr} \left[ \mathcal{A(\mathcal{D}) \in \mathcal{S}} \right] \leq e^{\epsilon} \mathrm{Pr} \left[ \mathcal{A(\mathcal{D}') \in \mathcal{S}}  \right] + \delta
      \end{align}
    \end{definition}
    \noindent $\epsilon$ denotes the so-called \textit{privacy budget}, which specifies the privacy loss we can afford. Smaller privacy budget implies stronger privacy preservation is necessitated. 
    %

    A widely used $(\epsilon, \delta)$-DP mechanism is Gaussian mechanism. Given any query function $q: \mathsf{D} \rightarrow \mathsf{R}$, $(\epsilon, \delta)$-DP can be ensurend by adding Gaussian noises to each dimension of $\mathsf{R}$. In particular, for any dataset $\mathcal{D} \in \mathsf{D}$, we have
    \begin{align} \label{eq:mechanA}
      \mathcal{A}(\mathcal{D}) = q(\mathcal{D}) + \mathcal{N}(0, \sigma^2 \mathbf{I}_d)
    \end{align}
    where $\mathcal{N}$ denotes a Gaussian distribution with zero mean, $\mathbf{I}_d$ is a $d$-dimensional identity matrix, and $\sigma^2 \mathbf{I}_d$ denotes the covariance matrix of the Gaussian distribution. We define the sensitivity of $q$ in \textbf{Definition}~\ref{def:l2-sensitivity}, and the noise magnitude $\sigma$ is proportional to the sensitivity of $q$.
    \begin{definition}[Sensitivity] \label{def:l2-sensitivity}
      The sensitivity of query function $q$ is the maximum absolute difference derived on the output of any two adjacent datasets $\mathcal{D}, \mathcal{D}' \in \mathsf{D}$, i.e.,
      \begin{align} \label{eq:l2-sensitivity}
          \triangle_{2} q = \max_{\mathcal{D}, \mathcal{D}'} \left\| q(\mathcal{D}) - q(\mathcal{D}') \right\|
      \end{align}
    \end{definition}

 \subsection{Cooperative Game and Shapley Value} \label{ssec:shapley}
   \begin{definition}[Cooperative Game]
      Let $\mathcal{Z}=\{1,2,\cdots,Z\}$ denote a set of $Z$ players. A cooperative game is represented by a tuple $(\mathcal{Z}, v)$, where $v: 2^{\mathcal{Z}} \rightarrow \mathbb{R}$ is a characteristic function of the game, assigning a real number to each coalition $\mathcal{Z}' \subseteq \mathcal{Z}$ and satisfying $v(\emptyset)=0$.
   \end{definition}

   The characteristic function $v(\mathcal{Z'})$ assigns the payoff attainable by coalition $\mathcal{Z'} \subseteq \mathcal{Z} $. 
   The distribution of the payoff of the coalition $\mathcal{Z'}$ among the players is a fundamental issue for cooperative game. As presented in \cite{shapley1971cores}, the payoff allocation among the players should meet the following desirable properties, i.e., balance, symmetry, additivity and zero element. Specifically,
   \begin{itemize}
      \item \textit{Balance}. The payoff obtained by each player in $\mathcal{Z}$ must add up to $v(\mathcal Z)$.
      \item \textit{Symmetry}. Two players must receive the same payoff if they contribute the same to each coalition.
      \item \textit{Additivity}. The solution to the sum of two cooperative games must be the sum of what it awards to each of the two games.
      \item \textit{Zero element}. A player should not receive payoff if it contributes nothing to each coalition.
    \end{itemize}
    
    To design a payoff allocation solution that satisfies the aforementioned properties, \cite{shapley1971cores} proposed to assign the payoff to each player  by considering its marginal contributions to all coalitions that the player could possibly participate in. Specifically, the Shapley value of a player is a weighted sum of the player's marginal contributions across all possible coalitions in the game.
    \begin{definition}[Shapley Value] \label{def:shapley}
      Let ${\mathcal{S}} \in \Omega(\mathcal{Z})$ denote a permutation of players in $\mathcal{Z}$ and $\mathcal{Z}_i ({\mathcal{S}})$ denote the coalition made of all predecessors of player $i$ in ${\mathcal{S}}$. Assuming ${\mathcal{S}}(i)$ represents the location of $i$ in $\mathcal{S}$, $\mathcal{Z}_i ({\mathcal{S}})$ can be denoted as
      \begin{equation} \label{eq:predecessor}
         \mathcal{Z}_i ({\mathcal{S}}) = \left\{ i' \in {\mathcal{S}} \mid {\mathcal{S}}(i') < {\mathcal{S}}(i) \right\}
      \end{equation}
      Given a cooperative game $(\mathcal{Z}, v)$, the Shapley value for player $i\in\mathcal{Z}$ is defined as follows 
      \begin{equation} \label{eq:shapley-1}
         \varphi_i(v) =  \frac{1}{Z !} \sum_{\mathcal{S} \in \Omega(\mathcal{Z})} \left( v \left( \mathcal{Z}_i (\mathcal{S}) \bigcup \{i\} \right) - v \left( \mathcal{Z}_i (\mathcal{S}) \right) \right)
      \end{equation}
    \end{definition}
    In another word, the payoff allocated to player $i$ in a coalition game is the average marginal contribution of $i$ to ${\mathcal{Z}}_i (\mathcal{S})$ over all $\mathcal{S} \in \Omega(\mathcal{Z})$. It is easy to show that (\ref{eq:shapley-1}) can be re-written as:
    \begin{equation} \label{eq:shapley-2}
       \varphi(i) = \sum_{\mathcal{Z}' \subseteq \mathcal{Z} \setminus \{i\}} \left(Z \dbinom{Z-1}{Z'} \right)^{-1}   \left(v \left( \mathcal{Z}' \bigcup \{i\} \right) - v \left( \mathcal{Z}' \right)\right) 
    \end{equation}
    where $Z = |\mathcal{Z}|$ and $Z' = |\mathcal{Z}'|$.

\section{Algorithm Design} \label{sec:pdsl}
  In our algorithm, cross-gradient information finely weighted by Shapley values is utilized to address the problem of data heterogeneity. Unfortunately, exchanging the cross-gradient information among the agents may result in privacy leakage. To resolve this issue, we leverage the notion of DP, such that the cross-gradient information can be perturbed through Gaussian mechanism for the purpose of privacy preservation.
  Specifically, the time span of our algorithm is divided into $T$ rounds. In each round $t$, each agent $i \in \mathcal{M}$ first calculates its local stochastic gradient $g^{[t]}_{i,i}$ according to its local dataset, and perturbs $g^{[t]}_{i,i}$ through the Gaussian mechanism. Then, each agent $i \in \mathcal{M}$ exchanges its current local model $x^{[t-1]}_i$ with its neighbors. Once receiving the local model $x^{[t-1]}_j$ from any neighbor $j\in\mathcal{M}_i$, agent $i$ calculates stochastic cross-gradient $g^{[t]}_{i,j}$ according to its local dataset for agent $j$. $g^{[t]}_{i,j}$ is perturbed by a Gaussian noise $n^{[t]}_{i,j}$, and the perturbed cross-gradient $\hat{g}^{[t]}_{i,j}$ is then sent back to agent $j$. When receiving $\hat{g}^{[t]}_{j,i}$ from each neighbor $j\in\mathcal{M}_i$, agent $i$ calculates an update of its local model $x^{[t-1]}_{i}$ according to $\hat{g}^{[t]}_{j,i}$. Based on the updates $\left\{ x^{[t]}_{i,j} \right\}_{j\in\mathcal{M}_i}$, agent $i$ calculates Shapley value $\varphi^{[t]}_{i,j}$ for each $j\in\mathcal{M}_i$, and the Shapley values are utilized to compute the weight parameters for the aggregation of the received (perturbed) stochastic gradients. The local model $x^{[t]}_i$ at each agent $i$ is finally updated in a monemtum-like manner.

  The pseudo-code of our PDSL algorithm is given in \textbf{Algorithm}~\ref{alg:pdsl}. For each agent $i \in \mathcal{M}$, in each round $t$, agent $i$ first computes its local stochastic gradient using a sample $\xi_{i,t}$ uniformly sampled from its local dataset $\mathcal{D}_i$
  \begin{equation} \label{eq:selfgrad}
    g^{[t]}_{i,i} = \nabla F_i \left( x^{[t-1]}_i; \xi_{i,t} \right)
  \end{equation}
  as shown in Line~\ref{ln:selfgrad}. Then, as presented in Lines~\ref{ln:clipselfgrad} $\sim$~\ref{ln:adnoiseselfgrad}, to limit gradient sensitivity and enable a quantifiable privacy assessment, agent $i$ clips its local stochastic gradient $g^{[t]}_{i,i}$ 
  \begin{equation} \label{eq:clipselfgrad}
    \tilde{g}^{[t]}_{i,i} = \frac{g^{[t]}_{i,i}}{\max \left\{ 1, C^{-1} \left\| g^{[t]}_{i,i} \right\| \right\}} 
  \end{equation}
  and adds random Gaussian noise $n^{[t]}_{i,i} \sim \mathcal{N}(0, \sigma^2 \mathbf{I}_d)$ to the clipped gradient such that
  \begin{equation} \label{eq:adnoiseselfgrad}
    \hat{g}^{[t]}_{i,i} = \tilde{g}^{[t]}_{i,i} + n^{[t]}_{i,i}
  \end{equation}
  where $C$ is the clipping threshold. 
  Next, agent $i$ broadcasts its current local model $x^{[t-1]}_i$ to its neighbors $\mathcal{M}_i \setminus \{i\}$ (see Line~\ref{ln:sendmodel01}). 
  As shown in Lines~\ref{ln:sendnoiselocalgradstart}$\sim$\ref{ln:sendnoiselocalgradend}, once receiving the local model $x^{[t-1]}_j$ from each neighbor $j \in \mathcal{M}_i \setminus \{i\}$, agent $i$ computes the stochastic cross-gradient 
  \begin{equation} \label{eq:localgrad}
    g^{[t]}_{i,j} = \nabla F_i \left( x^{[t-1]}_j; \xi_{i,t} \right)
  \end{equation}
  $g^{[t]}_{i,j}$ is then clipped by
  \begin{equation} \label{eq:cliplocalgrad}
     \tilde{g}^{[t]}_{i,j} = \frac{g^{[t]}_{i,j}}{\max \left\{ 1, C^{-1} \left\| g^{[t]}_{i,j} \right\| \right\}} 
  \end{equation}
  and perturbed by Gaussian noise $n^{[t]}_{i,j} \sim \mathcal{N}(0, \sigma^2 \mathbf{I}_d)$ as follows
  \begin{equation} \label{eq:adnoiselocalgrad}
    \hat{g}^{[t]}_{i,j} = \tilde{g}^{[t]}_{i,j} + n^{[t]}_{i,j}
  \end{equation}
  Thereafter, agent $i$ sends the perturbed gradient $\hat{g}^{[t]}_{i,j}$ back to agent $j$.

  Based on the gradient $\hat{g}^{[t]}_{j,i}$ returned by each neighbor, agent $i$ updates its local model according to
  \begin{equation} \label{eq:up4neighbor}
    x^{[t]}_{i,j} = x^{[t-1]}_{i} - \gamma \cdot \hat{g}^{[t]}_{j,i}, ~\forall j \in \mathcal{M}_i
  \end{equation}
  where $\gamma$ denotes the learning rate, as illustrated in Lines~\ref{ln:recenoiselocalgrad}$\sim$\ref{ln:model4shapley}. These local model updates are utilized to calculate the Shapley values (see Line~\ref{ln:compshapley}). Specifically, we define the characteristic function $v: 2^{\mathcal{M}_i} \rightarrow \mathbb{R}$ as follows
  \begin{align} \label{eq:charfunc}
    v \left(\mathcal{M}'; \mathcal{Q} \right) = \frac{1}{|\mathcal{Q}|} \sum_{\xi \in \mathcal{Q}} J \left( \xi; x^{[t]}_{i, \mathcal{M}'}\right), ~\forall \mathcal{M}' \subseteq \mathcal{M}_i
  \end{align}
  where 
  \begin{equation}
    x^{[t]}_{i, \mathcal{M}'} = \frac{1}{|\mathcal{M}'|} \sum_{j \in \mathcal{M}'} x^{[t]}_{i, j}
  \end{equation}
  and $J \left( \xi; x^{[t]}_{i, \mathcal{M}'}\right)$ denotes the accuracy of $x^{[t]}_{i, \mathcal{M}'}$ on the data sample $\xi \in \mathcal{Q}$. The Shapley value of any agent $j \in \mathcal{M}_i$ then can be calculated as
  \begin{equation} \label{eq:compshapley}
    \varphi^{[t]}_{i,j} = \sum_{\mathcal{M}' \subseteq \mathcal{M}_i \setminus \{j\}}  \frac{ v \left( \mathcal{M}' \bigcup \{j\}; {\mathcal{Q}} \right) - v \left( \mathcal{M}'; {\mathcal{Q}} \right) }{ |\mathcal{M}_i| \binom{ |\mathcal{M}_i| - 1}{|\mathcal{M}'|} }
  \end{equation}
  and normalized by 
  \begin{equation} \label{eq:compnormshapley}
    \hat{\varphi}^{[t]}_{i,j} = \frac{\varphi^{[t]}_{i,j} - \min_{k \in \mathcal{M}_i} \varphi^{[t]}_{i,k}}{\max_{k \in \mathcal{M}_i} \varphi^{[t]}_{i,k} - \min_{k \in \mathcal{M}_i} \varphi^{[t]}_{i,k}} 
  \end{equation}
  as illustrated in Line~\ref{ln:compnormshapley}. Based on these normalized Shapley values, agent $i$ calculates a weight parameter for $\forall j \in \mathcal{M}_i$ 
  \begin{equation} \label{eq:weight}
    \pi^{[t]}_{i,j} = \frac{\hat{\varphi}^{[t]}_{i,j}}{\omega_{i,j} \sum_{k \in \mathcal{M}_i} \hat{\varphi}^{[t]}_{i,k}}
  \end{equation}
  as demonstrated in Line~\ref{ln:compweight}.
  Next, as shown in Line~\ref{ln:avgnoisegrad}, agent $i$ calculates a weighted average across the received perturbed stochastic gradients
  \begin{equation}  \label{eq:avggrad}
    \bar{g}^{[t]}_i = \sum_{j\in \mathcal{M}_i} \pi^{[t]}_{i,j} \hat{g}^{[t]}_{j, i}
  \end{equation}
  Finally, the local model $x^{[t]}_i$ is updated in a momentum-like manner. Specifically, agent $i$ first calculates
  \begin{equation} \label{eq:imupmomentum}
    \hat{u}^{[t]}_i = \alpha {u}^{[t-1]}_i + \bar{g}^{[t]}_i
  \end{equation}
  and
  \begin{equation} \label{eq:imupmodel}
    \hat{x}^{[t]}_i = {x}^{[t-1]}_i - \gamma \hat{u}^{[t]}_i
  \end{equation}
  as shown in Lines~\ref{ln:uplocmoment01} and \ref{ln:uplocpara01}, respectively, where $\alpha$ denotes the momentum coefficient. It then broadcasts $\hat{u}^{[t]}_i$ and $\hat{x}^{[t]}_i$ to its neighbors (see Line~\ref{ln:sendmomentlocpara}). After receiving $\hat{u}^{[t]}_j$ and $\hat{x}^{[t]}_j$ from $\forall j \in \mathcal{M}_i \setminus \{ i \}$, agent $i$ updates its momentum and local model by
  \begin{equation} \label{eq:updatemomentum}
    u^{[t]}_i = \sum_{j\in \mathcal{M}_i} \omega_{i,j} \hat{u}^{[t]}_j
  \end{equation}
  and
  \begin{equation} \label{eq:updatemodel}
    x^{[t]}_i =  \sum_{j\in \mathcal{M}_i} \omega_{i,j} \hat{x}^{[t]}_j
  \end{equation}

  \begin{algorithm}[htb!]
    \KwIn{Initial point $x^{[0]}_i$, learning rate $\gamma$, momentum buffer $u^{[0]}_i = 0$, momentum coefficient $\alpha$, doubly stochastic matrix $\mathbf{W}$, the number of iterations $T$, clipping threshold $C$, variance of noise $\sigma^2$, $d$-dimensional identity matrix $\mathbf{I}_d$.}
    \KwOut{Local model $x^{[T]}_i$ of agent $i$}
    \ForEach{$t=1,2,\cdots,T$}{
      Compute local gradient $g^{[t]}_{i,i}$ according to (\ref{eq:selfgrad}); \label{ln:selfgrad} \\
      Clip local stochastic gradient $\tilde{g}^{[t]}_{i,i} = \frac{g^{[t]}_{i,i}}{\max \left\{1, C^{-1} \left\| g^{[t]}_{i,i} \right\| \right\}} $; \label{ln:clipselfgrad} \\
      Randomly generate Gaussian noise $n^{[t]}_{i,i} \in \mathbb{R}^d$ from $\mathcal{N}(0, \sigma^2 \mathbf{I}_d)$ and add the noise to local stochastic gradient: $\hat{g}^{[t]}_{i,i} = \tilde{g}^{[t]}_{i,i} + n^{[t]}_{i,i}$;
      \label{ln:adnoiseselfgrad}\\
      Send $x^{[t-1]}_i$ to each neighbor $j \in \mathcal{M}_i \setminus \{i\}$; \label{ln:sendmodel01}\\
      \ForEach{$j \in \mathcal{M}_i \setminus \{i\}$}{ \label{ln:sendnoiselocalgradstart}
        Receive $x^{[t-1]}_j$ from $j \in \mathcal{M}_i \setminus \{i\}$; \label{ln:recemodel01}\\
        Compute gradient $g^{[t]}_{i,j}$ according to (\ref{eq:localgrad}); \label{ln:localgrad} \\
        Clip stochastic gradient $\tilde{g}^{[t]}_{i,j} =\frac{g^{[t]}_{i,j}}{\max \left\{1, C^{-1} \left\| g^{[t]}_{i,j} \right\| \right\}}$;  \label{ln:cliplocalgrad}\\
        Randomly generate Gaussian noise $n^{[t]}_{i,j} \in \mathbb{R}^d$ from $\mathcal{N}(0, \sigma^2 \mathbf{I}_d)$ and add the noise to stochastic gradient: $\hat{g}^{[t]}_{i,j} = \tilde{g}^{[t]}_{i,j} + n^{[t]}_{i,j}$;    \label{ln:adnoiselocalgrad}\\
        Send $\hat{g}^{[t]}_{i,j}$ back to agent $j$; \label{ln:sendnoiselocalgrad}\\
      } \label{ln:sendnoiselocalgradend}
      Receive $\hat{g}^{[t]}_{j,i}$ from each neighbor $j \in \mathcal{M}_i \setminus \{i\}$; \label{ln:recenoiselocalgrad}\\
      Compute $x^{[t]}_{i,j} = x^{[t-1]}_{i} - \gamma \cdot \hat{g}^{[t]}_{j,i}$ for $\forall j \in \mathcal{M}_i$; \label{ln:model4shapley}\\
      Compute Shapley value $\varphi^{[t]}_{i,j}$ for $\forall j\in\mathcal{M}_i$ according to \textbf{Algorithm}~\ref{alg:shapeyest};   \label{ln:compshapley}\\
      Compute normalized Shapley value $\hat{\varphi}^{[t]}_{i,j}$ for $\forall j\in\mathcal{M}_i$ according to (\ref{eq:compnormshapley}); \label{ln:compnormshapley} \\
      Compute weight $\pi^{[t]}_{i,j}$ for $\forall j \in \mathcal{M}_i$ based on (\ref{eq:weight}); \label{ln:compweight}\\
      $\bar{g}^{[t]}_i = \sum_{j\in \mathcal{M}_i} \pi^{[t]}_{i,j} \hat{g}^{[t]}_{j, i}$;  \label{ln:avgnoisegrad}\\
      $\hat{u}^{[t]}_i = \alpha u^{[t-1]}_i + \bar{g}^{[t]}_i$;  \label{ln:uplocmoment01} \\
      $\hat{x}^{[t]}_i = x^{[t-1]}_i - \gamma \hat{u}^{[t]}_i$;  \label{ln:uplocpara01} \\
      Send $\hat{u}^{[t]}_i$ and $\hat{x}^{[t]}_i$ to each neighbor $j \in \mathcal{M}_i \setminus \{i\}$; \label{ln:sendmomentlocpara}\\
      Receive $\hat{u}^{[t]}_j$ and  $\hat{x}^{[t]}_j$ from $\forall j \in \mathcal{M}_i \setminus \{i\}$; \label{ln:recemomentlocpara}\\
      $u^{[t]}_i = \sum_{j\in \mathcal{M}_i} \omega_{i,j} \hat{u}^{[t]}_j$; \label{ln:uplocmoment02}  \\
      $x^{[t]}_i =  \sum_{j\in \mathcal{M}_i} \omega_{i,j} \hat{x}^{[t]}_j$;  \label{ln:uplocpara02}  
    }
  \caption{Our PDSL algorithm (at each agent $i$).} \label{alg:pdsl}
  \end{algorithm}

  Based on Sec.~\ref{ssec:shapley}, for each agent $i \in \mathcal{M}$, the computation of Shapley values according to (\ref{eq:compshapley}) primarily depends on the permutations of $\mathcal{M}_i$. This process can lead to high computational complexity, particularly when agent $i$ has a large number of neighbors. To tackle this issue, we employ the Monte Carlo method to compute Shapley values with lower complexity~\cite{CastroGT-COR09, FatimaWJ-AI08}. The pseudo-code of our Monte Carlo-based estimation algorithm (for agent $i$ in round $t$) is presented in \textbf{Algorithm}~\ref{alg:shapeyest}. Specifically, we initialize $\varphi^{[t]}_{i,j} = 0$ for each agent $j \in \mathcal{M}_i$. In each iteration $r = 1,2,\cdots, R$, we randomly generate a permutation $\phi_r$ of $\mathcal{M}_i$. Then, as shown in Line~\ref{ln:shapleyest-prede}, we construct $\mathcal{Z}_j(\phi_r)$ for each agent $j \in \mathcal{M}_i$, i.e., the set of predecessors of $j$ in $\phi_r$, and finally compute $v\left(\mathcal{Z}_j\left(\phi_r\right) \bigcup \{j\}; \mathcal{Q} \right)$ and $v\left(\mathcal{Z}_j\left(\phi_r\right); \mathcal{Q} \right)$ using (\ref{eq:charfunc}), and calculate $\varphi^{[t]}_{i,j}$ by
  \begin{equation} \label{eq:updateshap}
    \varphi^{[t]}_{i,j} \leftarrow \varphi^{[t]}_{i,j} + \frac{v\left(\mathcal{Z}_j\left(\phi_r\right) \bigcup \{j\}; \mathcal{Q} \right) - v\left(\mathcal{Z}_j\left(\phi_r\right); \mathcal{Q} \right)} {R}
  \end{equation}
  instead of (\ref{eq:compshapley}) as illustrated in Line~\ref{ln:shapleyest-shapley}.
  \begin{algorithm}[htb!]
    \KwIn{Neighbors $\mathcal{M}_i$, local models $\left\{ x^{[t]}_{i,j} \right\}_{j \in \mathcal{M}_{i}}$ and validation dataset $\mathcal{Q}$, and the maximum number of iterations $R$.}
    \KwOut {Shapley value $\varphi^{[t]}_{i,j}$ of each agent $j \in \mathcal{M}_i$ in round $t$.}
    %
    %
    %
    $\varphi^{[t]}_{i,j} = 0$ for $\forall j \in \mathcal{M}_i$;  \label{ln:shapleyest-init}\\
    \ForEach{$r = 1, 2, \cdots, R$}{ 
      Let $\phi_r$ be a random permutation of  $\mathcal{M}_i$;  \label{ln:shapleyest-perm}\\
      \ForEach{$j \in \mathcal{M}_i $}{
        $\mathcal{Z}_j \left( \phi_r \right) = \left\{ j' \in \mathcal{M}_i ~\big{|}~ \phi_r(j') < \phi_r(j) \right\}$;  \label{ln:shapleyest-prede}\\
        $\varphi^{[t]}_{i,j} \leftarrow \varphi^{[t]}_{i,j} + \frac{v\left(\mathcal{Z}_j\left(\phi_r\right) \bigcup \{j\}; \mathcal{Q} \right) - v\left(\mathcal{Z}_j\left(\phi_r\right); \mathcal{Q} \right)} {R}$
            \label{ln:shapleyest-shapley} \\
        }
    }
  \caption{Shapley value estimation (for agent $i$ in round $t$).} 
  \label{alg:shapeyest}
  \end{algorithm}

\section{Theoretical Analysis} \label{sec:analysis}
  In this section, theoretical analysis is conducted to reveal the performance of our PDSL algorithm in terms of privacy preservation and convergence in Sec.~\ref{ssec:privacy} and Sec.~\ref{ssec:mainresult}, respectively. All proofs of the \textbf{Theorems}~\ref{thm:privacy}$\sim$\ref{thm:convergence-main} and \textbf{Corollary}~\ref{cor:main} are given in the  appendices (see Appendix A and Appendix B).

  \subsection{Privacy Preservation} \label{ssec:privacy}
    The notion of DP implies that we have better privacy preservation and thus smaller privacy budget if stronger noise (with higher variance of the Gaussian noise in our case) is adopted. \textbf{Theorem}~\ref{thm:privacy} reveals a lower bound of $\sigma$ (which determines the minimum magnitude of the Gaussian noise) to ensure $(\epsilon, \delta)$-DP.
    \begin{theorem} \label{thm:privacy}
       For any $\epsilon > 0, \delta \in (0,1)$, \textbf{Algorithm}~\ref{alg:pdsl} can guarantee $(\epsilon,\delta)$-DP in each round, if
      \begin{align} \label{eq:privacy-stand-dev}
        \sigma \geq \max_{i \in \mathcal{M}} \left\{\frac{2C\left( \frac{1}{ \omega_{\min}} + \sum_{j \in \mathcal{M}_i} \frac{1}{\omega_{i,j}} \right) \sqrt{2 \ln(1.25/\delta)} }{\hat{\varphi}_{\min}  \epsilon \sqrt{\sum_{j \in \mathcal{M}_i} \omega^{-2}_{i,j}}} \right\}
      \end{align}
      where $C$ is the clipping threshold of any gradient, $\omega_{\min} = \min_{i \in \mathcal{M}, j \in \mathcal{M}_i} \omega_{i,j}$, and $\hat{\varphi}_{\min} = \min_{j \in \mathcal{M}_i, t\in T} \frac{\hat{\varphi}^{[t]}_{i,j}}{\sum_{k \in \mathcal{M}_i} \hat{\varphi}^{[t]}_{i,k}}$.
    \end{theorem}

   \subsection{Convergence Analysis} \label{ssec:mainresult}
    Before presenting the main result regarding the convergence of our PDSL algorithm, we first introduce \textbf{Assumptions}~\ref{ass:L-smooth}$\sim$\ref{ass:ds_mat} which are commonly employed in the analysis of distributed machine learning in existing state-of-the-art proposals \cite{YuJY-ICML19, EsfandiariTJBHHS-ICML21}.
    \textbf{Assumption}~\ref{ass:L-smooth} indicates that the gradient variations of each function $f_i(x)$ are smooth, and $\mathcal{F}(x)$ is also $L$-smooth. \textbf{Assumption}~\ref{ass:bd-varia} bounds the variance of stochastic gradient, and  \textbf{Assumption}~\ref{ass:ds_mat} assumes the adjacency matrix of the communication graph is doubly stochastic.
    \begin{assumption} \label{ass:L-smooth}
      Each function $f_i(x)$ is $L$-smooth for any $i\in\mathcal{M}$, i.e., there exists a constant $L>0$ such that 
      \begin{equation} \label{eq:L-smooth}
        \left \| \nabla f_i(x) - \nabla f_i(y) \right\| \leq L\|x-y\|, ~~\forall x,y \in \mathbb{R}^d
      \end{equation}
    \end{assumption}
    \begin{assumption} \label{ass:bd-varia}
      There exist $\zeta > 0$, $\kappa > 0$ such that
      \begin{equation} \label{eq:bd-varia0}
        \mathbb{E}_{\xi \sim \mathcal{D}_i} \Big\| \nabla F_i(x; \xi) - \nabla f_i(x) \Big\|^2 \leq \zeta^2, ~~\forall i \in \mathcal{M}
      \end{equation}
      and
      \begin{equation} \label{eq:bd-varia1}
        \Big\| \nabla f_i(x) - \nabla \mathcal F(x) \Big\|^2 \leq \kappa^2,~~\forall i \in \mathcal{M}
      \end{equation}
    \end{assumption}
    \begin{assumption} \label{ass:ds_mat}
      Given the symmetric doubly stochastic matrix $\mathbf{W}$, let $\lambda_i(\mathbf{W})$ denote the $i$-th largest eigenvalue of $\mathbf{W}$. we have $\lambda_1(\mathbf{W}) = 1$ and there exists constant $\rho < 1$ such that $ \max \{ |\lambda_2(\mathbf{W})|, |\lambda_N(\mathbf{W})| \} \leq \sqrt{\rho}$.
    \end{assumption}

    The main results about the convergence of PDSL is given in \textbf{Theorem}~\ref{thm:convergence-main}. In particular, the upper bound of the average gradient magnitude achieved by our PDSL algorithm mainly depends on the initial value of the objective function and the optimal value. In addition, the strength of the Gaussian noise, represented by $\sigma$, also exerts impact on the convergence of our PDSL algorithm. It is a minor polynomial factor and thus cannot disturb the convergence of our algorithm.
    \begin{theorem} \label{thm:convergence-main}
      Suppose \textbf{Assumptions}~\ref{ass:L-smooth}$\sim$\ref{ass:ds_mat} hold, and learning rate $\gamma$ satisfies the following condition
      \begin{align} \label{main-rate}
        \frac{(1-\alpha)^2}{\alpha} < \gamma \leq \min \left\{ \begin{aligned} & \frac{(1-\alpha)(1-\sqrt{\rho})}{2\sqrt{26}L},   \alpha(1-\sqrt{\rho})^2 (2\sqrt{13}L)^{-2} \bigg( -1 +  \sqrt{\frac{52L^2(1-\alpha)^2}{\alpha^2(1-\sqrt{\rho})^2} + 1} \bigg)\end{aligned}  \right\}
      \end{align}
      For any $T \geq 1$, we have
      \begin{align} \label{eq:main-bound}
        & \frac{1}{T} \sum^{T}_{t=1} \mathbb{E} \left[ \left\| \nabla \mathcal{F} \left( \bar{x}^{[t-1]} \right) \right\|^2 \right]  \nonumber\\
        \leq & \frac{1}{m_1 T} \left( \mathcal{F} \left( \bar{x}^{[0]} \right) - \mathcal{F}^* \right) \nonumber\\
        & + \left( m_2 + m_3 \frac{\gamma^2 \alpha^2}{(1-\alpha)^4} + m_4 \right) \left( \frac{4 C^2}{\omega^4_{\min}} + \frac{4 \sigma^2 d}{\omega^4_{\min}} + \frac{2 \zeta^2}{M} \right) \nonumber\\
        & + m_5 \left( \frac{ 16\gamma^2 \left( C^2 + \sigma^2 d \right)}{\omega^4_{\min} (1-\alpha)^2 (1-\sqrt{\rho})^2} + \frac{4 \gamma^2  \left( 7\zeta^2 + 13\kappa^2 \right)}{(1-\alpha)^2 (1-\sqrt{\rho})^2} \right)
      \end{align}
      for \textbf{Algorithm}~\ref{alg:pdsl}, where $\mathcal{F}^*$ denotes the optimal value of the objective function, and
      \begin{align} \label{eq:main-constants}
      \begin{cases}
        \bar{x}^{[t]} = \frac{1}{M} \sum^M_{i=1} x^{[t]}_i \vspace{1ex}\\
        \omega_{\min} = \min_{i \in \mathcal{M}, j \in \mathcal{M}_i} \omega_{i,j} \vspace{1ex}\\
        m_1 = \frac{\gamma}{2(1-\alpha)} - \frac{1-\alpha}{2\alpha} \vspace{1ex} \\
        m_2 = \frac{1}{m_1} \left( \frac{\alpha L \gamma^2}{2(1-\alpha)^3} + \frac{L \gamma^2}{2(1-\alpha)^2} \right) \vspace{1ex}\\
        m_3 = \frac{L(1-\alpha)}{2 m_1 \alpha}  \vspace{1ex}\\
        m_4 = \frac{\alpha \gamma^2}{2 m_1 (1-\alpha)^3}   \vspace{1ex}\\
        m_5 = \frac{L^2 \gamma}{2 m_1 (1-\alpha)}
      \end{cases}
     \end{align}
    \end{theorem}

   Then, \textbf{Corollary}~\ref{cor:main} provides a more explicit demonstration of the convergence property of our PDSL algorithm.
   \begin{corollary} \label{cor:main}
    Let learning rate $\gamma = O\left(\sqrt{\frac{1}{T}}\right)$. When $T$ is sufficiently large such that
    \begin{align} \label{eq:cormain-t}
        T \geq \max \left\{ \begin{aligned} & \frac{104L^2}{(1-\alpha)^2 (1-\sqrt{\rho})^2},  \frac{52^2L^4}{\left( (1-\sqrt{\rho})\sqrt{52L^2(1-\alpha)^2 + \alpha^2 (1-\sqrt{\rho})^2} - \alpha (1-\sqrt{\rho})^2 \right)^2 }
        \end{aligned} \right\}
    \end{align}
    we have
    \begin{align} \label{eq:cor}
        \frac{1}{T} \sum^{T}_{t=1} \mathbb{E} \left[ \left\| \nabla \mathcal{F} \left( \bar{x}^{[t-1]} \right) \right\|^2 \right] \leq K \left( \frac{1}{\sqrt{T}} + \frac{\sigma^2 d}{\sqrt{T}} + \frac{1}{M\sqrt{T}} + \frac{1}{T} + \frac{\sigma^2 d}{T} \right)
    \end{align}
    for some constant $K > 0$.
  \end{corollary}
  This corollary indicates that, when $T$ is sufficiently large, the convergence rate of our PDSL algorithm mainly depends on $\mathcal{O} \left( \frac{\sigma^2}{\sqrt{T}} \right)$. When 'stronger' Gaussian noise is adopted for higher differential privacy preservation, we have to be content with slower convergence. Furthermore, $\sigma$ is usually a constant and it is revealed that our PDSL algorithm has a linear speedup (with respect to $T$). To the best of our knowledge, this convergence rate is comparable to the well-known results in decentralized SGD algorithms (e.g., \cite{EsfandiariTJBHHS-ICML21, XuZW-TPAMI22}).

\section{Experiments} \label{sec:exp}
  In this section, we evaluate the performance of our PDSL algorithm through extensive experiments. We introduce our experimental setup in Sec.~\ref{ssec:setup}, and then introduce the state-of-the-art reference algorithms in Sec.~\ref{ssec:refalgo}. We finally report and analyze the experiment results in Sec.~\ref{ssec:results}.

  \subsection{Datasets and Experiment Setup} \label{ssec:setup}
    We adopt two standard image datasets, i.e., MNIST and CIFAR-10, in our experiments. The MNIST dataset consists of $70,000$ handwritten digits from ten classes, with $60,000$ training images and $10,000$ test images. The CIFAR-10 dataset consists of $60,000$ images in ten classes, such as airplane, automobile, bird, cat, deer, dog, frog, horse, ship, and truck, $50,000$ images are used for training and the remaining $10,000$ images are used for testing. For both MNIST and CIFAR-10 datasets, we randomly select $2,000$ images ($20\%$) from the original test dataset to serve as the global validation dataset.

    We utilize two \textit{Convolutional Neural Network} (CNN) models to assess the efficacy of our PDSL algorithm on MNIST and CIFAR-10 datasets, respectively. The CNN model for MNIST dataset is composed of two $3 \times 3$ convolutional layers, each succeeded by a $2 \times 2$ max pooling layer, then followed by a fully connected layer. The one for CIFAR-10 dataset incorporates two $5 \times 5$ convolutional layers, each succeeded by a $2 \times 2$ max pooling layer, and then two fully connected layers. We adopt ReLU activation functions in the convolutional layers in the above CNN models. The mini-batch size is set to be $250$ for both MNIST and CIFAR-10 datasets. For MNIST dataset, Let momentum coefficient $\alpha=0.5$ and learning rate $\gamma=0.001$. For CIFAR-10 dataset, Let $\alpha=0.7$ and $\gamma=0.01$.

    To obtain heterogeneous data distribution across different agents, we adopt Dirichlet distribution $\mathsf{Dir}(\mu p)$ to compute probability vectors for data labels, where $\mu$ is a parameter that influences the data distribution of agents, $p$ is a $Y$-dimensional vector with all components are $1$, and $Y$ denotes the number of different labels. As $\mu \rightarrow 0$, the data distribution across different agents tends to be more diverse. In our experiments, we let $\mu = 0.25$ for both MNIST and CIFAR-10 datasets.

    Moreover, we study the performance of our PDSL algorithm with three different communication graph topologies in our experiments, i,e., fully connected graph, ring graph, and bipartite graph. The fully connected graph is a dense topology, while the bipartite and ring graphs are sparser topologies. 

  \subsection{Reference Algorithms}  \label{ssec:refalgo}
    In this section, we compare our PDSL algorithm with the following state-of-the-art ones. 
    \begin{itemize}
      \item \textbf{DP-DPSGD}: In \cite{XuZW-TPAMI22}, the notion of DP is leveraged to enable privacy preservation in decentralized learning. In the proposed algorithm, each agent uses perturbed gradient information to update its local model. Unfortunately, it does not take into account data heterogeneity.
      \item \textbf{MUFFLIATO}: MUFFLIATO is a privacy amplification mechanism, consisting of local Gaussian noise injection at the individual agents followed by gossiping for averaging the private values \cite{CyffersEBM-NeurIPS22}. However, MUFFLIATO either does not consider the issue of data heterogeneity.
      \item \textbf{DP-CGA}: \textit{Cross-Gradient Aggregation} (CGA) algorithm \cite{EsfandiariTJBHHS-ICML21} is a decentralized learning algorithm with non-IID data distributions. We combine it with DP to ensure both learning efficacy of privacy preservation. Specifically, before sharing the cross-gradient information with its neighbors, each agent perturbs it with Gaussian noise.
      %
      %
      \item \textbf{DP-NET-FLEET}: In the \textit{Decentralized Networked Federated Learning with Recursive Gradient Correction} (NET-FLEET) algorithm \cite{ZhangFLYLZ-MobiHoc22}, a recursive gradient correction technique is adopted to handle heterogeneous data distribution across the agents, such that each agent runs multiple local updates between two consecutive communication rounds with their neighbors. In our experiments, we enable privacy preservation in NET-FLEET algorithm through Gaussian mechanism. 
    \end{itemize}

  \subsection{Experiment Results}  \label{ssec:results}
    \subsubsection{Experiment Results on MNIST Dataset}  \label{ssec:res-minist}
\begin{figure*}[htb!]
          \begin{center}
            \parbox{.3\textwidth}{\center\includegraphics[width=.25\textwidth]{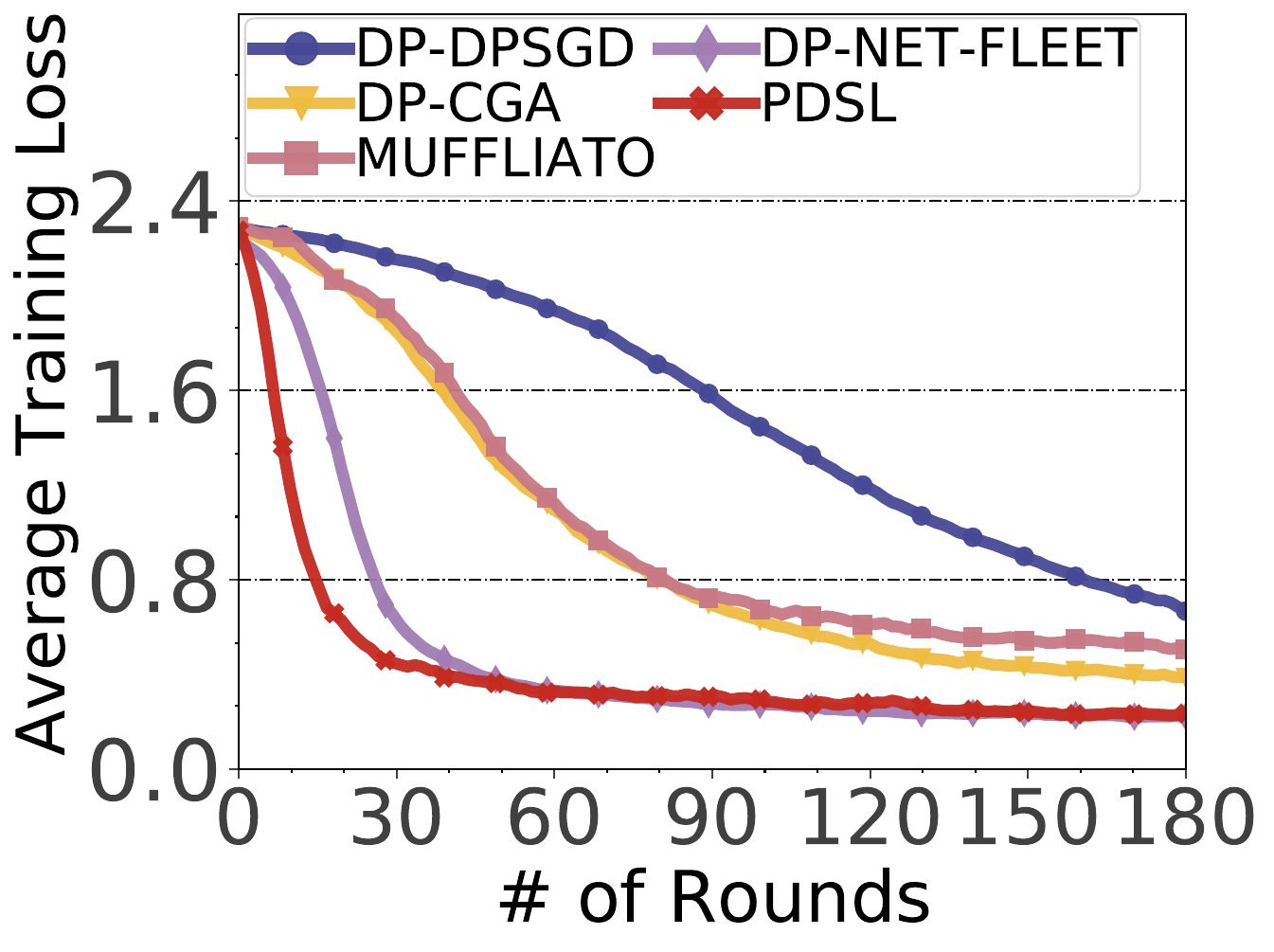}}
            \parbox{.3\textwidth}{\center\includegraphics[width=.25\textwidth]{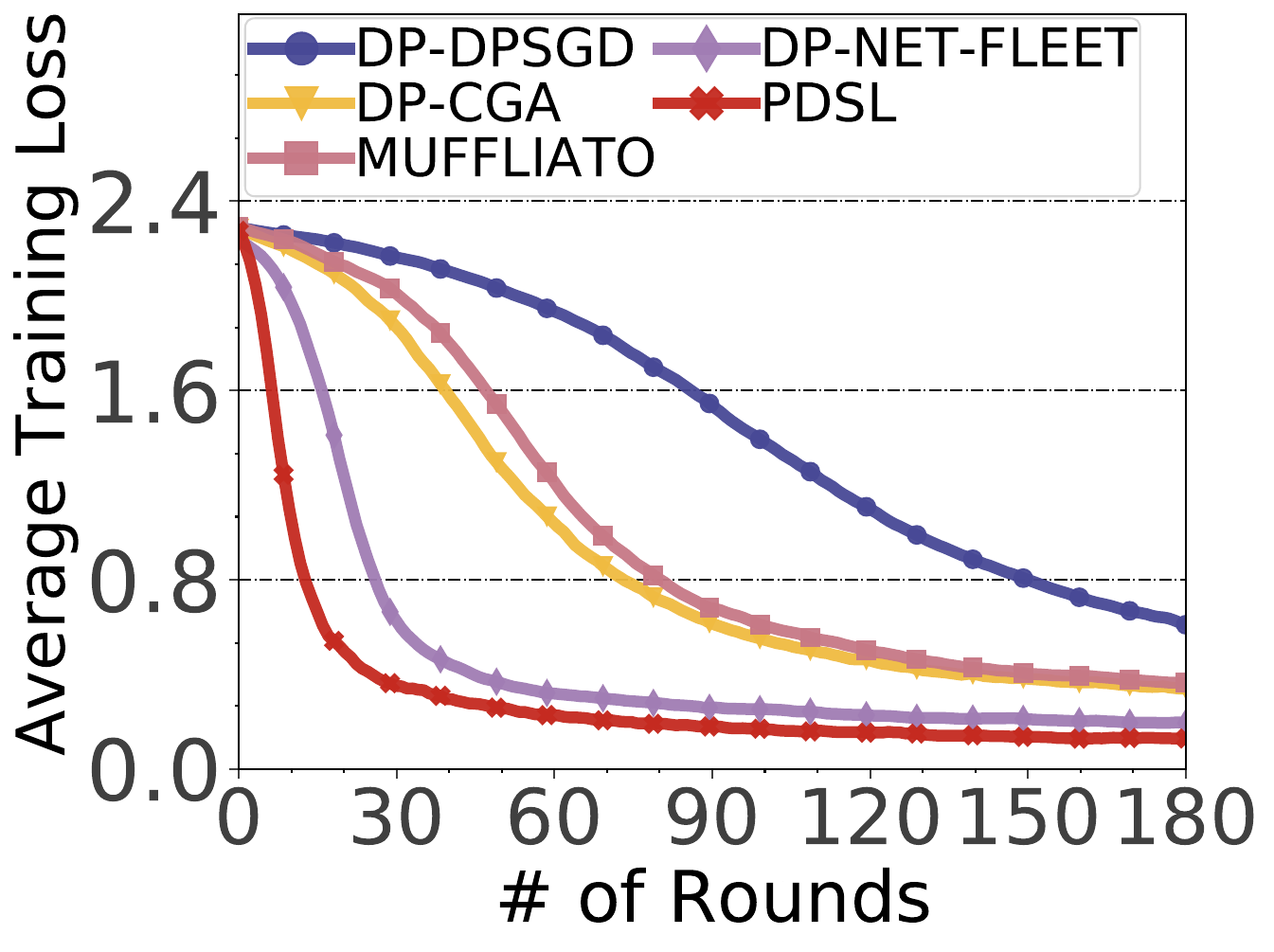}}
            \parbox{.3\textwidth}{\center\includegraphics[width=.25\textwidth]{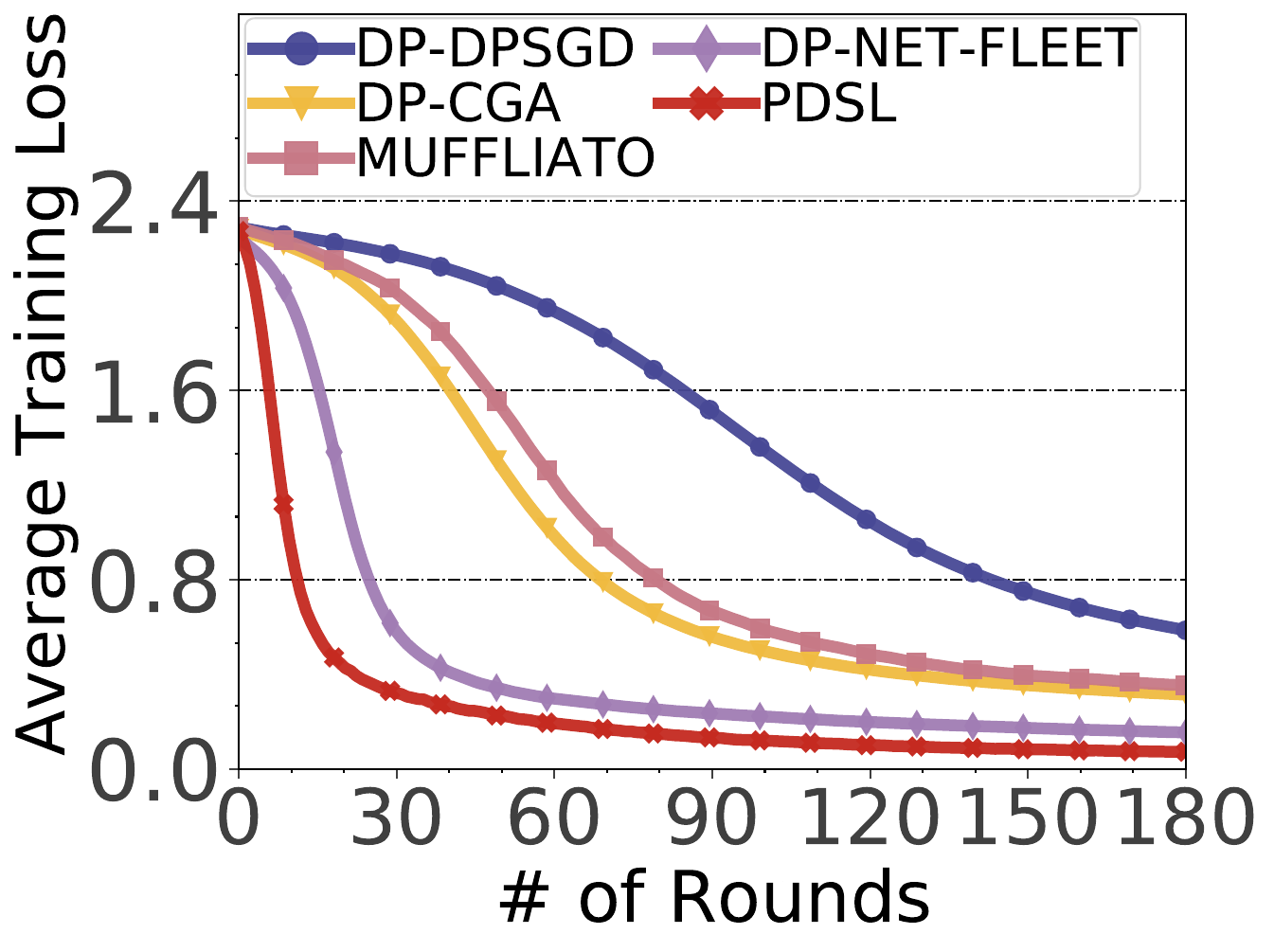}}
            \parbox{.3\textwidth}{\center\scriptsize(a1) $\epsilon=0.08$ with $M=10$}
            \parbox{.3\textwidth}{\center\scriptsize(a2) $\epsilon=0.1$ with $M=10$}
            \parbox{.3\textwidth}{\center\scriptsize(a3) $\epsilon=0.3$ with $M=10$}
            \parbox{.3\textwidth}{\center\includegraphics[width=.25\textwidth]{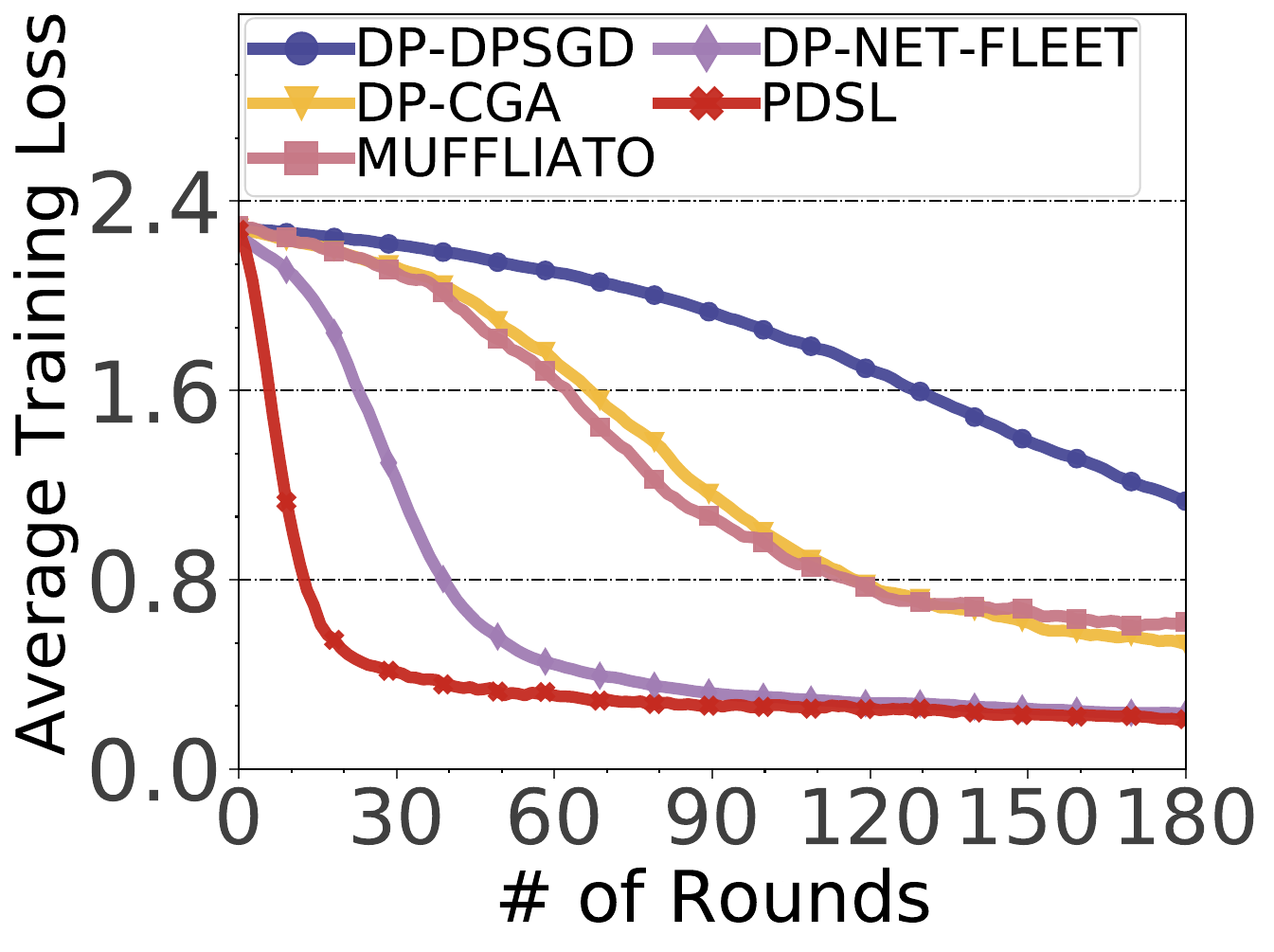}}
            \parbox{.3\textwidth}{\center\includegraphics[width=.25\textwidth]{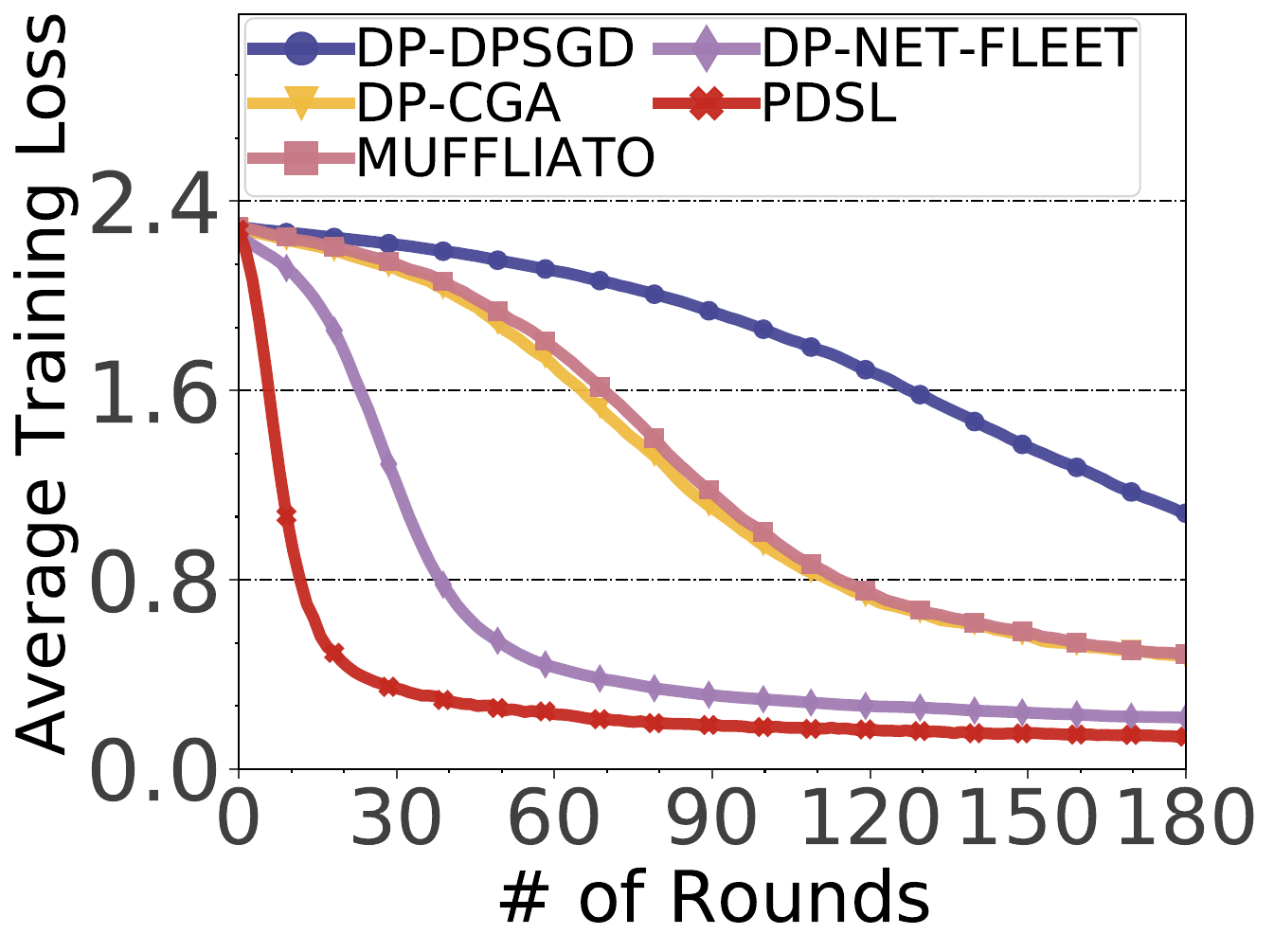}}
            \parbox{.3\textwidth}{\center\includegraphics[width=.25\textwidth]{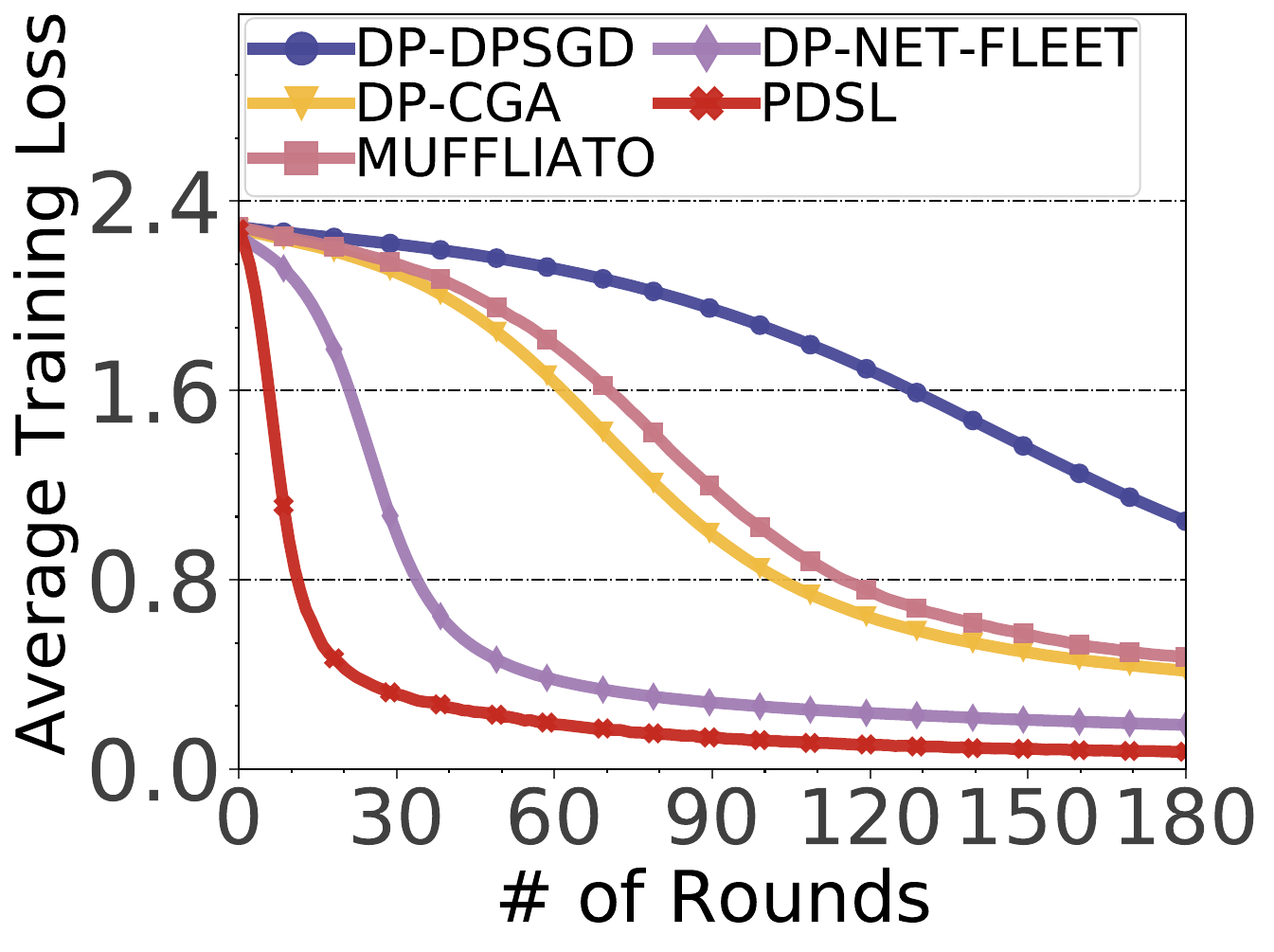}}
            \parbox{.3\textwidth}{\center\scriptsize(b1) $\epsilon=0.08$ with $M=15$}
            \parbox{.3\textwidth}{\center\scriptsize(b2) $\epsilon=0.1$ with $M=15$}
            \parbox{.3\textwidth}{\center\scriptsize(b3) $\epsilon=0.3$ with $M=15$}
            \parbox{.3\textwidth}{\center\includegraphics[width=.25\textwidth]{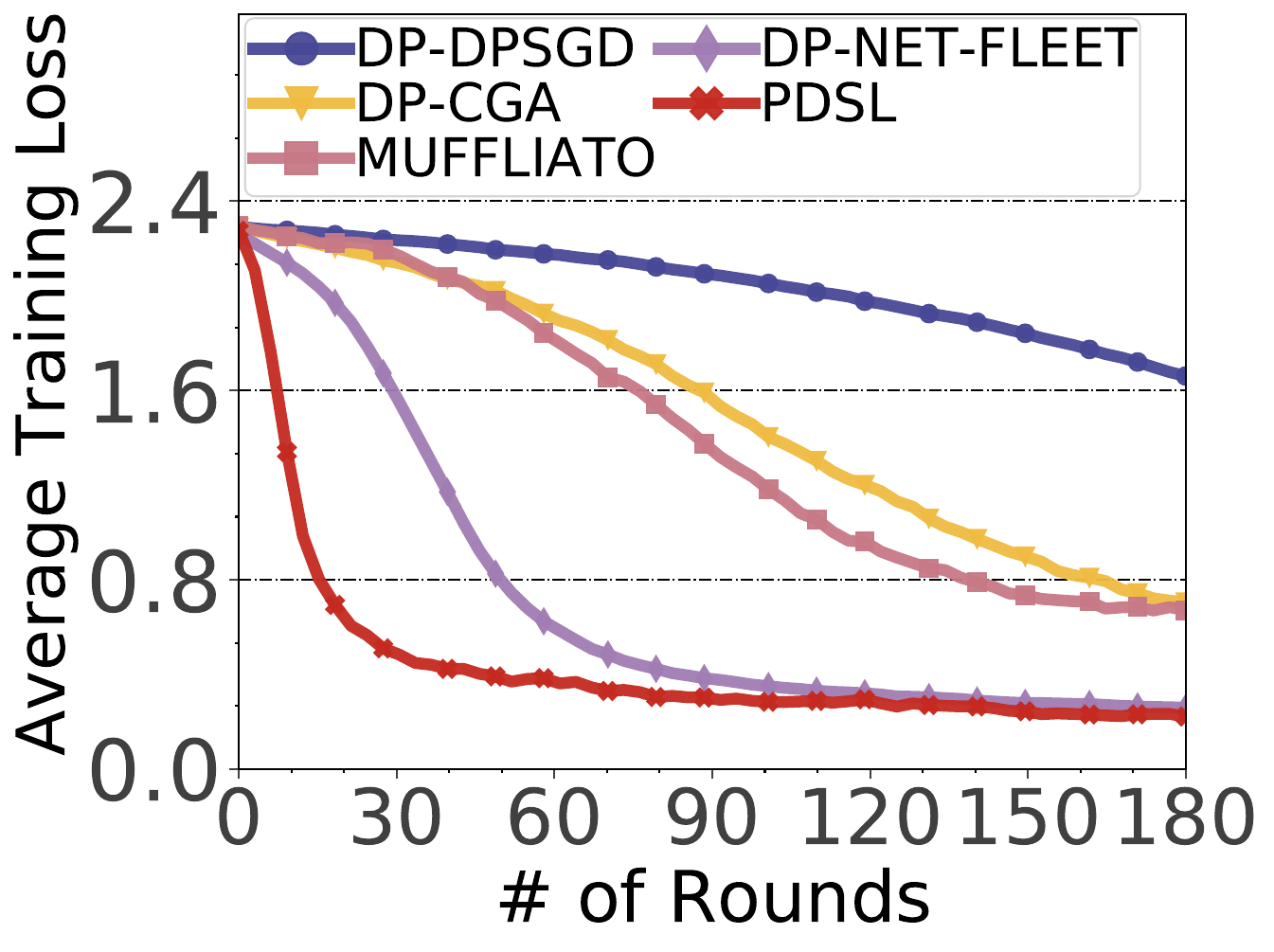}}
            \parbox{.3\textwidth}{\center\includegraphics[width=.25\textwidth]{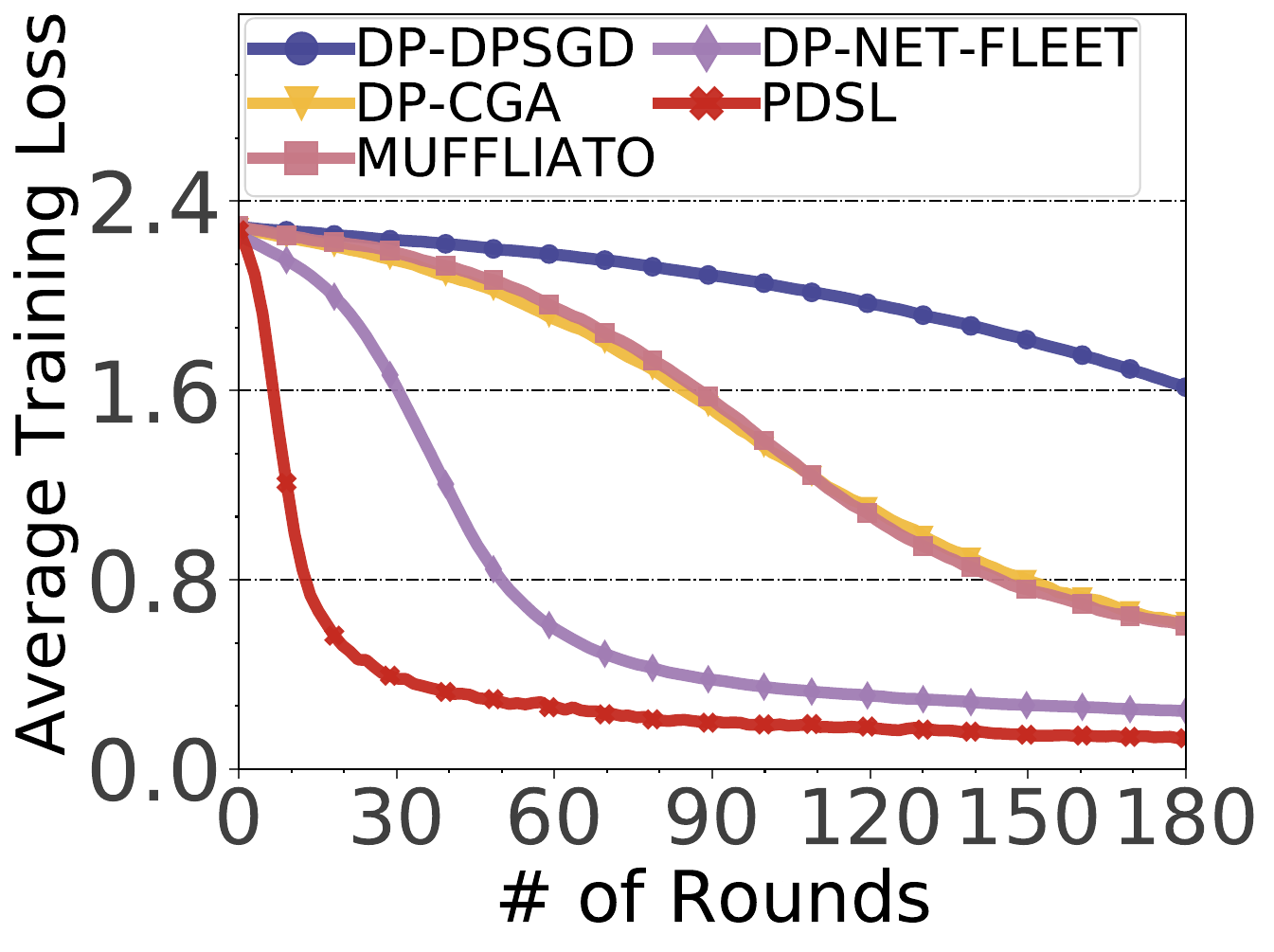}}
            \parbox{.3\textwidth}{\center\includegraphics[width=.25\textwidth]{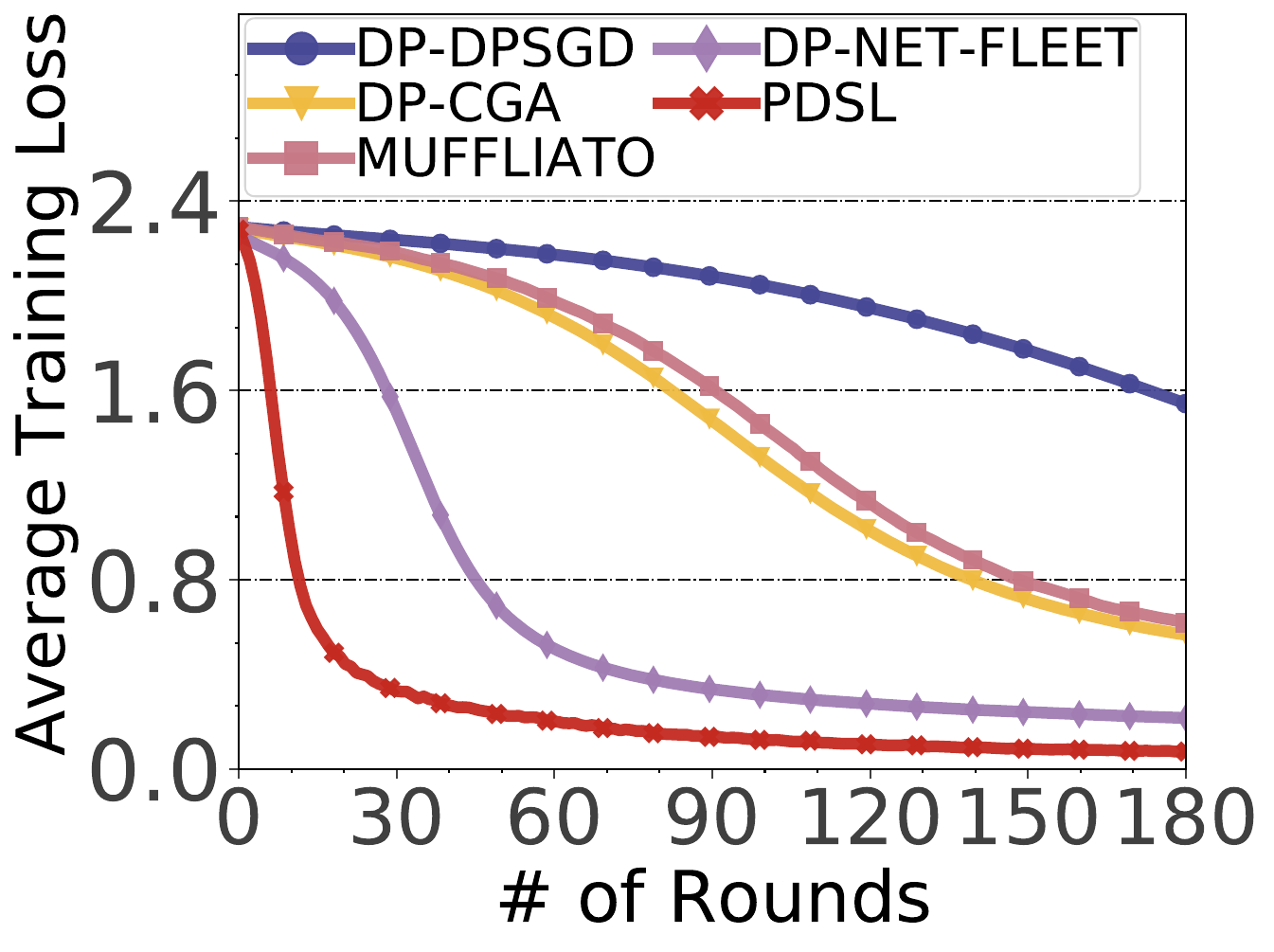}}
            \parbox{.3\textwidth}{\center\scriptsize(c1) $\epsilon=0.08$ with $M=20$}
            \parbox{.3\textwidth}{\center\scriptsize(c2) $\epsilon=0.1$ with $M=20$}
            \parbox{.3\textwidth}{\center\scriptsize(c3) $\epsilon=0.3$ with $M=20$}
          \caption{Comparison results on MNIST dataset over fully connected graphs.}
          \label{fig:mnist-loss-full}
          \end{center}
       \end{figure*}

      In Fig.~\ref{fig:mnist-loss-full}, we show the average losses of the different algorithms with different numbers of agents $M=10, 15, 20$ on fully connected communication graphs. We also vary privacy budget $\epsilon = 0.08, 0.1, 0.3$. The results indicate that our PDSL algorithm exhibits a faster convergence rate than the other four reference algorithms across the different privacy budgets. Specifically, our PDSL algorithm can reach its minimum loss value after about $80$ rounds. Even with an increased number of agents, the convergence rate of our PDSL algorithm remains consistent, ensuring that $80$ rounds are still sufficient for convergence. However, when the number of agents increases (e.g., $M=15, 20$), the other four reference algorithms require more communication rounds to achieve convergence. Specifically, when $M=20$, DP-NET-FLEET algorithm reaches convergence within about $100$ rounds, while DP-DPSGD, DP-CGA and MUFFLIATO algorithms fail to guarantee convergence even after $180$ communication rounds.

      Moreover, as demonstrated in Fig.~\ref{fig:mnist-loss-full}, compared to other reference algorithms, our PDSL algorithm achieves a lower average loss across the different values of $\epsilon$. For example, when $M=10$ and $\epsilon=0.3$, the average loss of PDSL is about $0.1$ when the convergence is reached; it is $1.6$ times lower than that of DP-NET-FLEET, $3.5$ times lower than that of MUFFLIATO, $3.2$ times lower than that of DP-CGA, and $5.8$ times lower than that of DP-DPSGD. As the privacy budget $\epsilon$ decreases (e.g., $\epsilon = 0.1, 0.08$) such that more noise is injected, our PDSL algorithm still exhibits a lower average loss compared to other reference algorithms. For instance, when $M=10, \epsilon=0.08$, the final average loss of PDSL is $2.5$ times lower than that of MUFFLIATO algorithm, $2$ times lower than that of DP-CGA algorithm, and $3.5$ times lower than that of DP-DPSGD algorithm. Additionally, it is noteworthy that the average loss of PDSL remains nearly unchanged with an increased number of agents (e.g., $M =15, 20$) when the privacy budget $\epsilon$ remains constant, unlike the other reference algorithms which show a significant increase. For example, when $M=20, \epsilon=0.3$, the final average loss of our PDSL algorithm is $6 \sim 15$ times lower than the ones of the reference algorithms.

\begin{figure*}[htb!]
          \begin{center}
            \parbox{.3\textwidth}{\center\includegraphics[width=.25\textwidth]{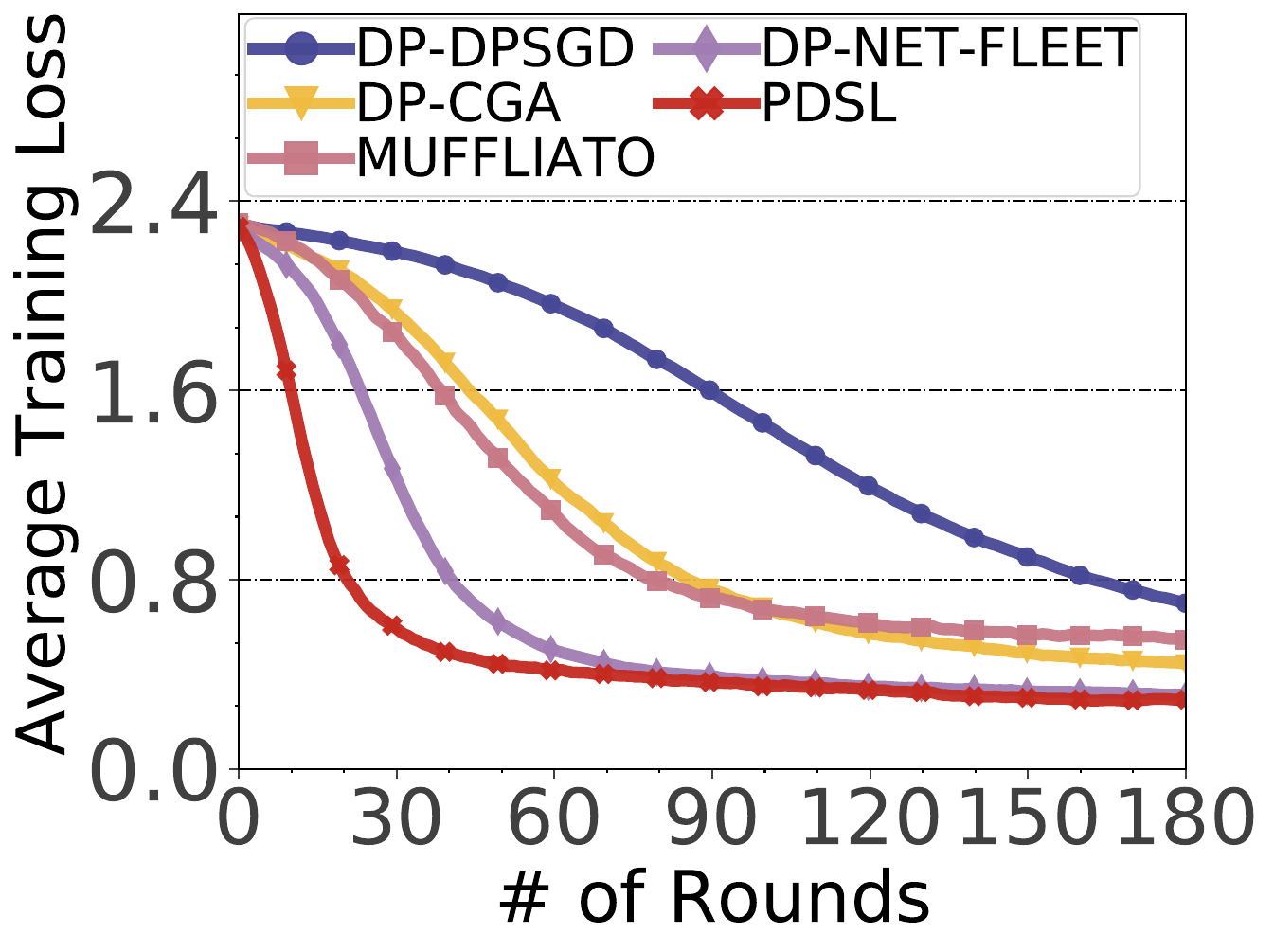}}
            \parbox{.3\textwidth}{\center\includegraphics[width=.25\textwidth]{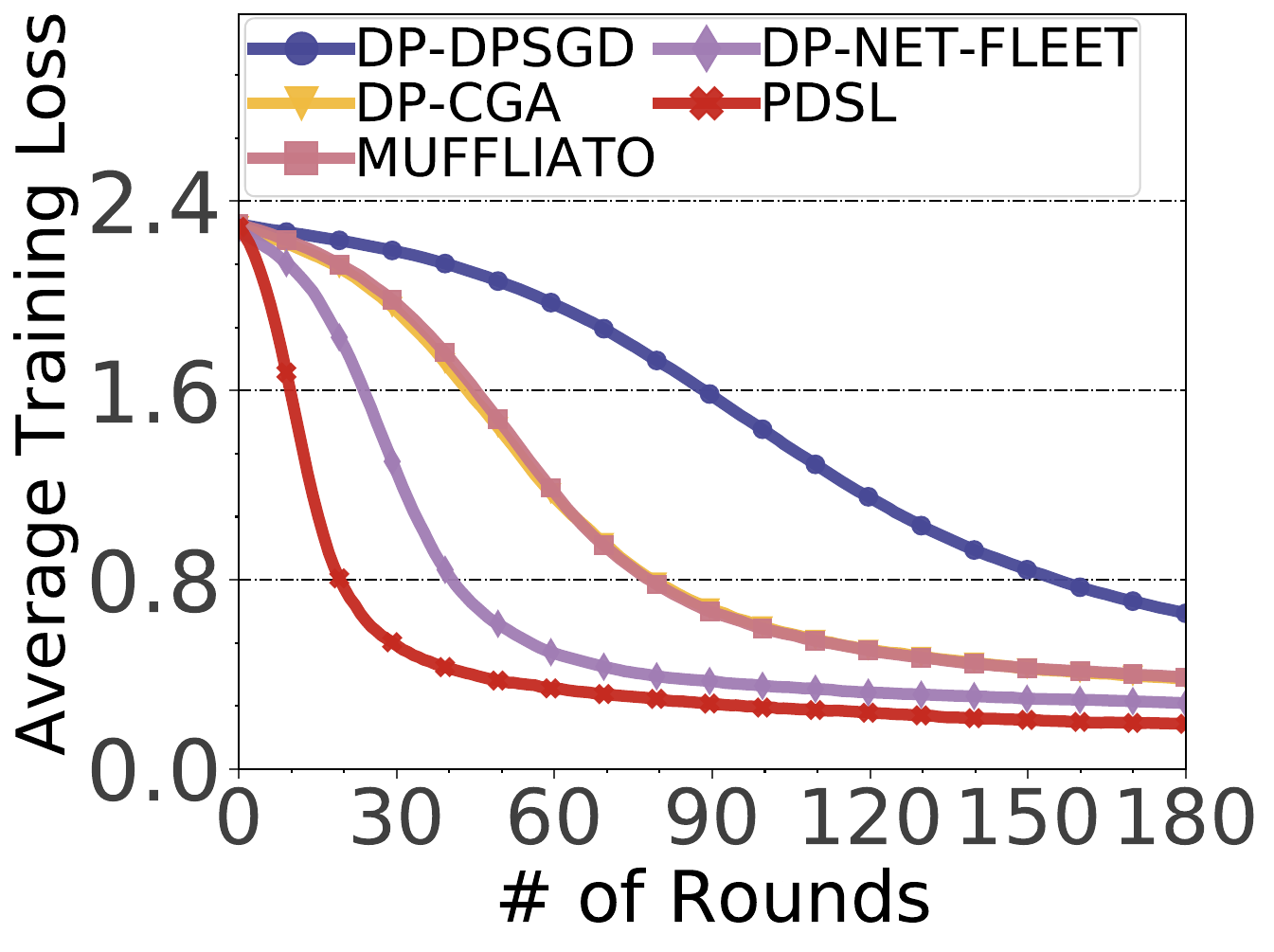}}
            \parbox{.3\textwidth}{\center\includegraphics[width=.25\textwidth]{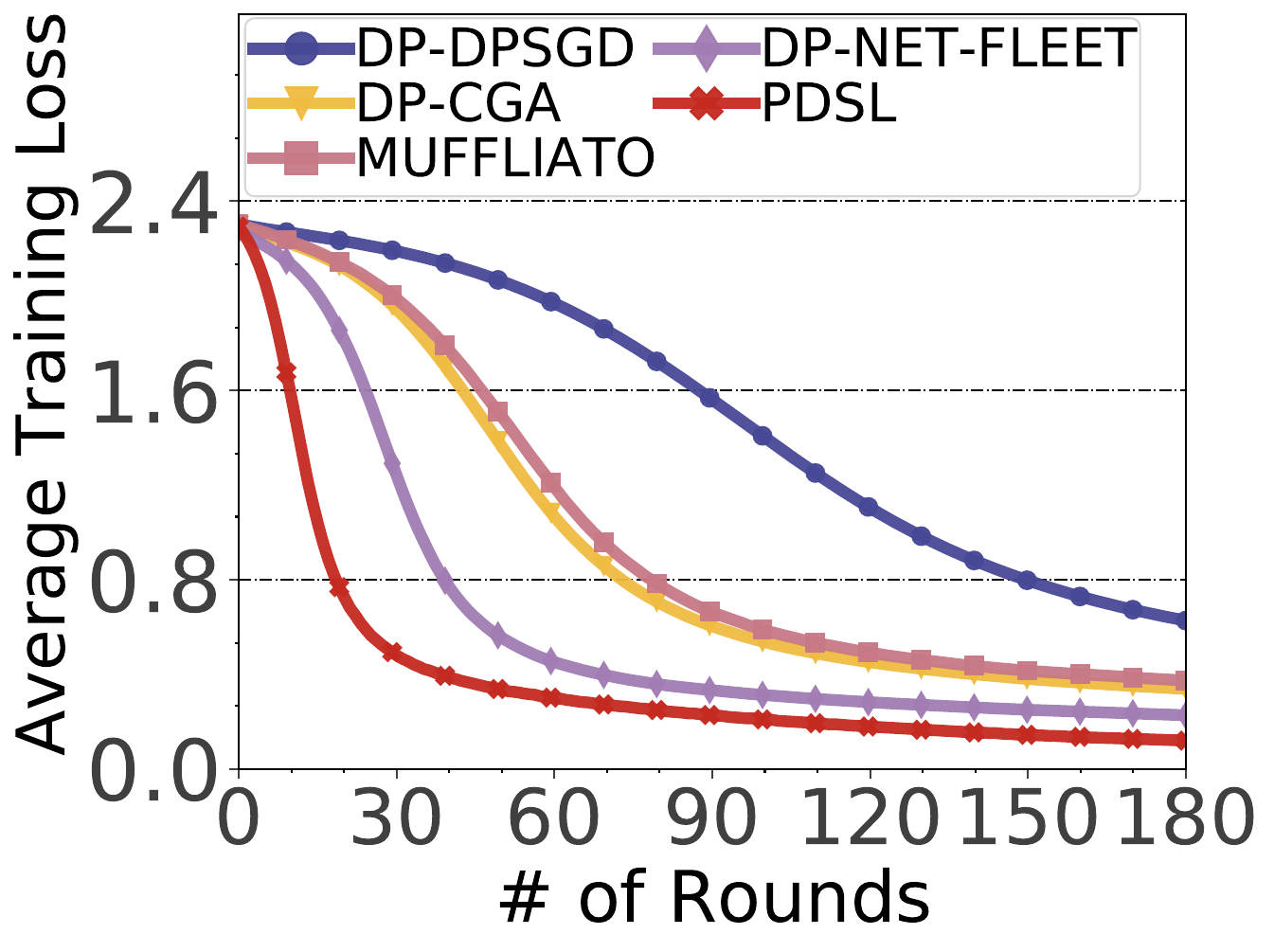}}
            \parbox{.3\textwidth}{\center\scriptsize(a1) $\epsilon=0.08$ with $M=10$}
            \parbox{.3\textwidth}{\center\scriptsize(a2) $\epsilon=0.1$ with $M=10$}
            \parbox{.3\textwidth}{\center\scriptsize(a3) $\epsilon=0.3$ with $M=10$}
            \parbox{.3\textwidth}{\center\includegraphics[width=.25\textwidth]{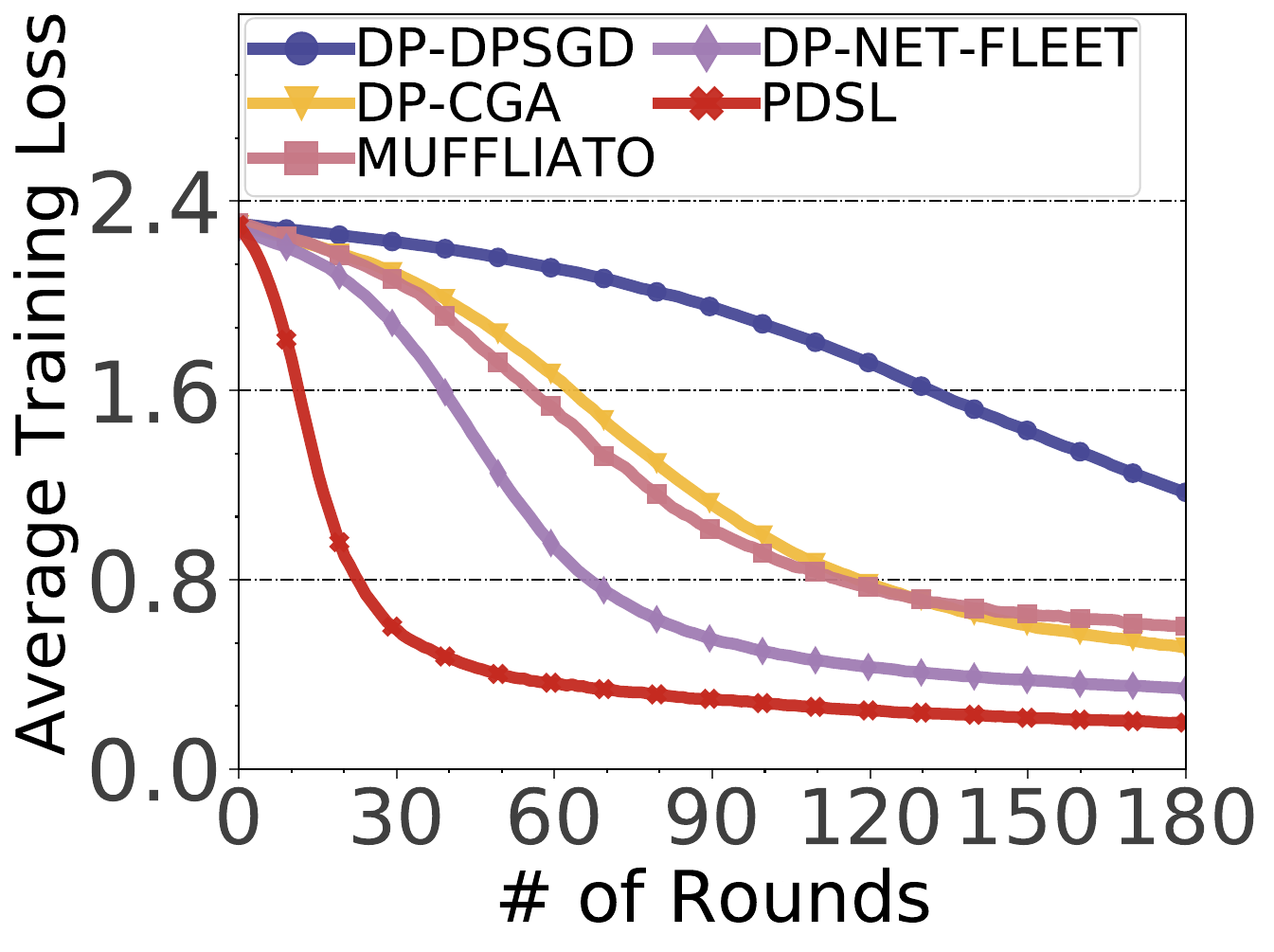}}
            \parbox{.3\textwidth}{\center\includegraphics[width=.25\textwidth]{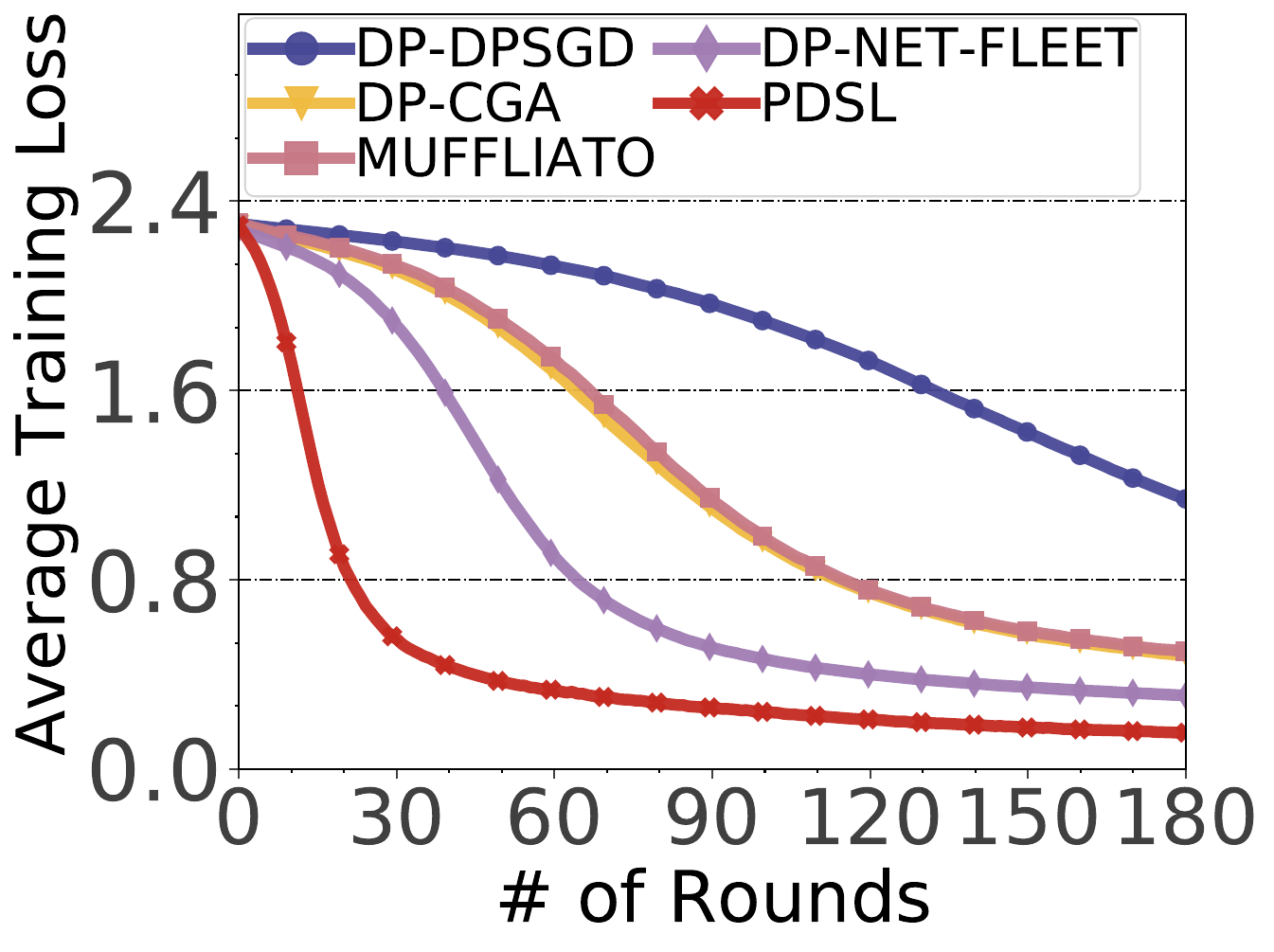}}
            \parbox{.3\textwidth}{\center\includegraphics[width=.25\textwidth]{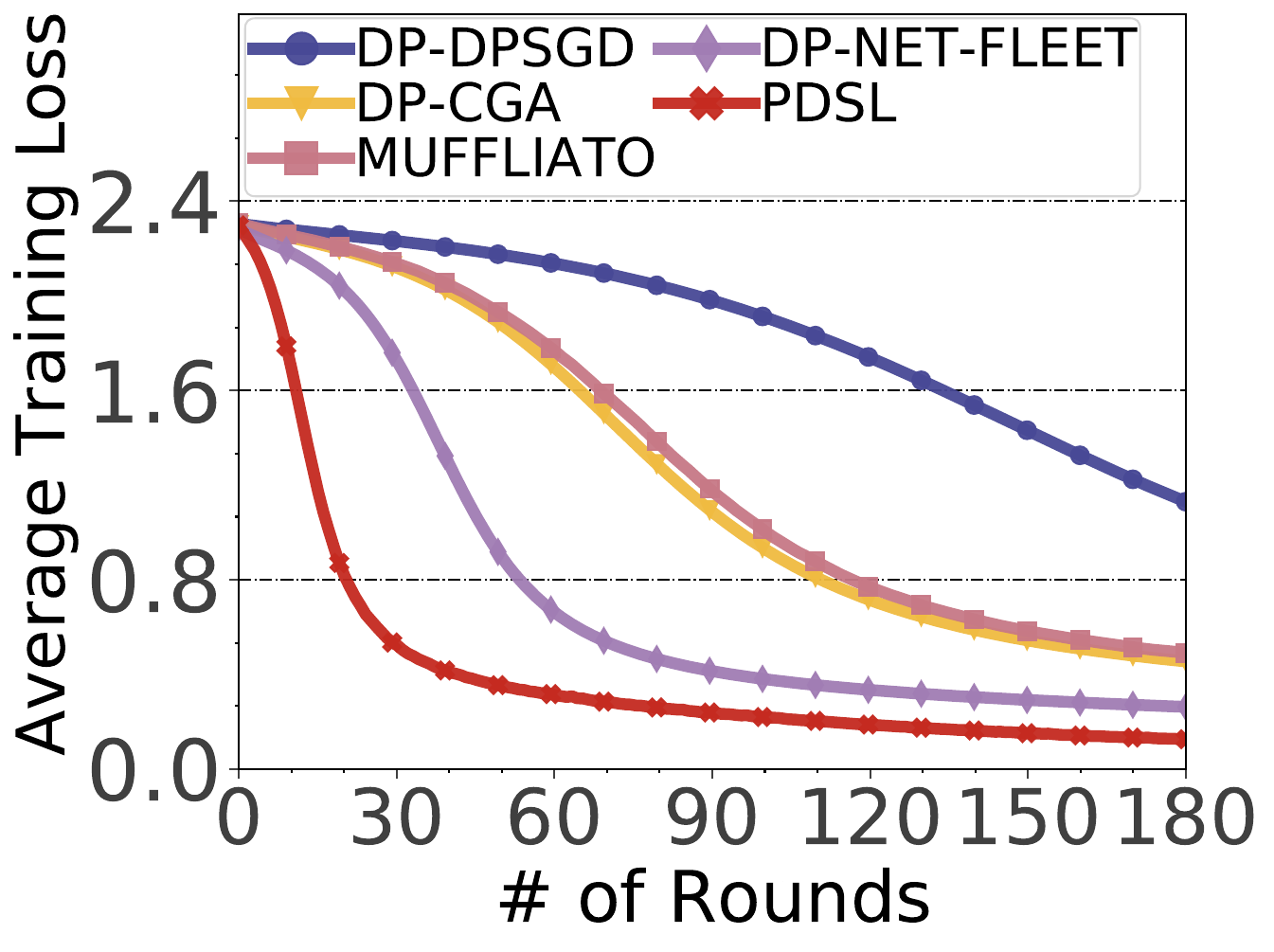}}
            \parbox{.3\textwidth}{\center\scriptsize(b1) $\epsilon=0.08$ with $M=15$}
            \parbox{.3\textwidth}{\center\scriptsize(b2) $\epsilon=0.1$ with $M=15$}
            \parbox{.3\textwidth}{\center\scriptsize(b3) $\epsilon=0.3$ with $M=15$}
            \parbox{.3\textwidth}{\center\includegraphics[width=.25\textwidth]{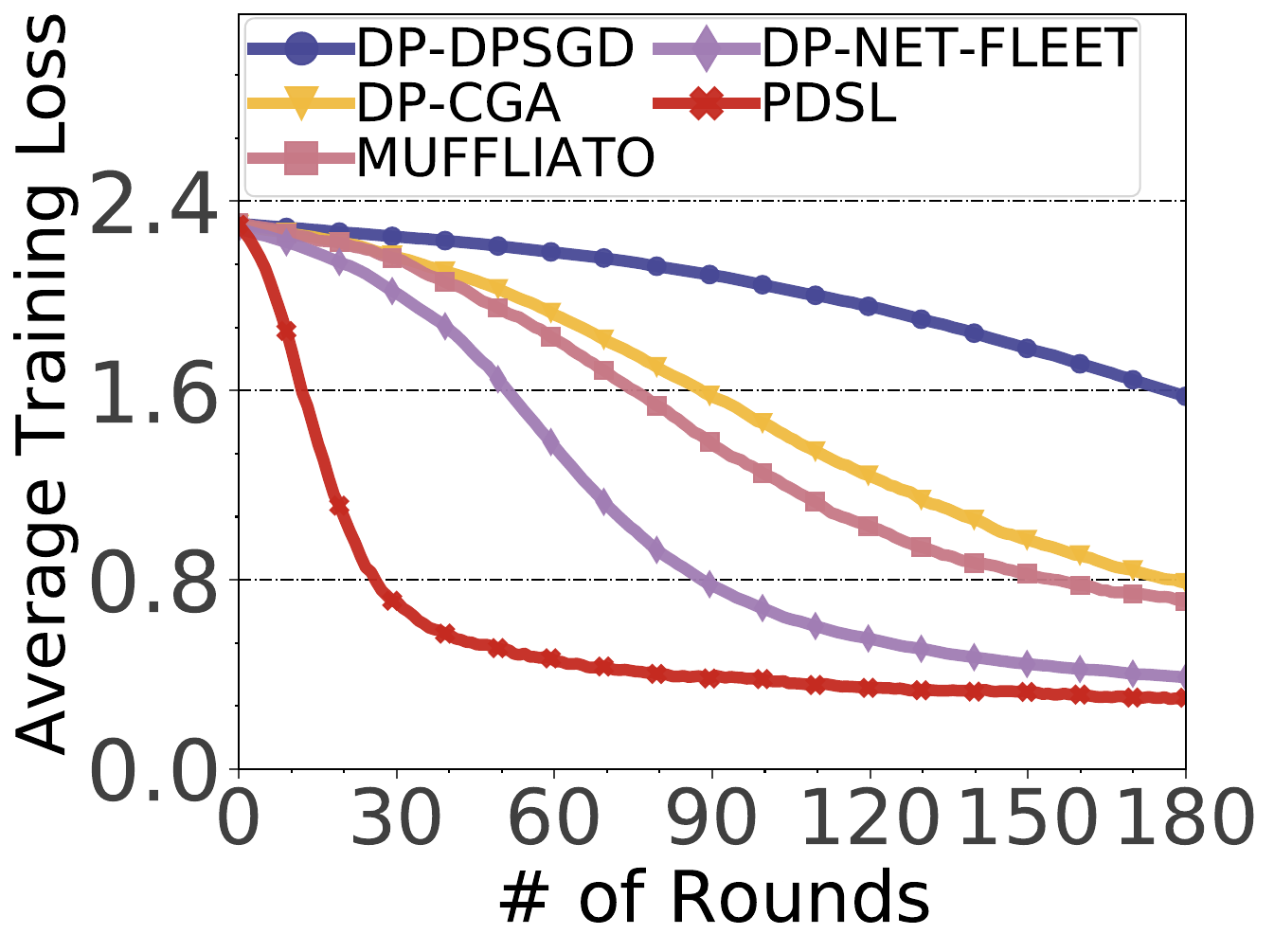}}
            \parbox{.3\textwidth}{\center\includegraphics[width=.25\textwidth]{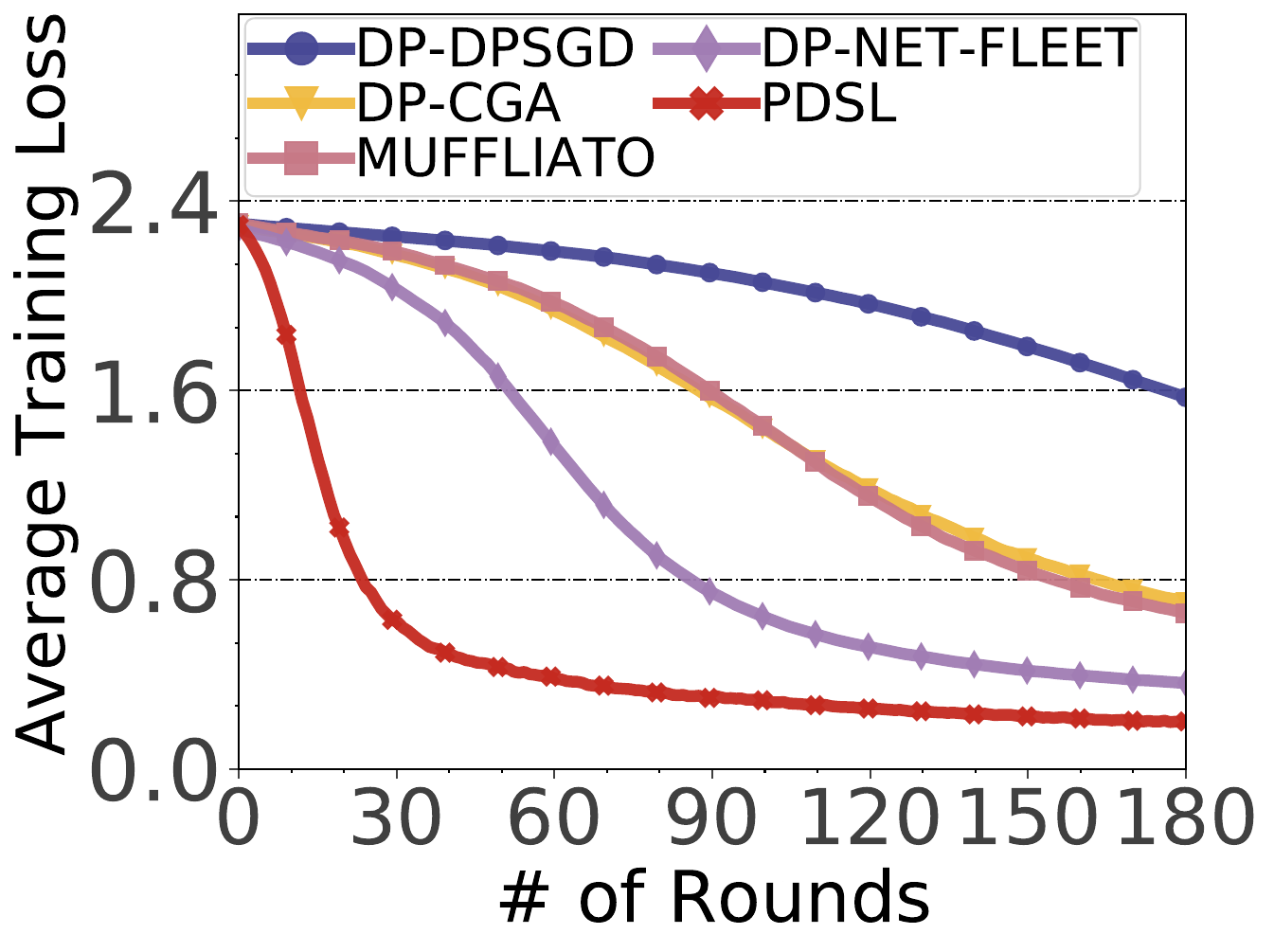}}
            \parbox{.3\textwidth}{\center\includegraphics[width=.25\textwidth]{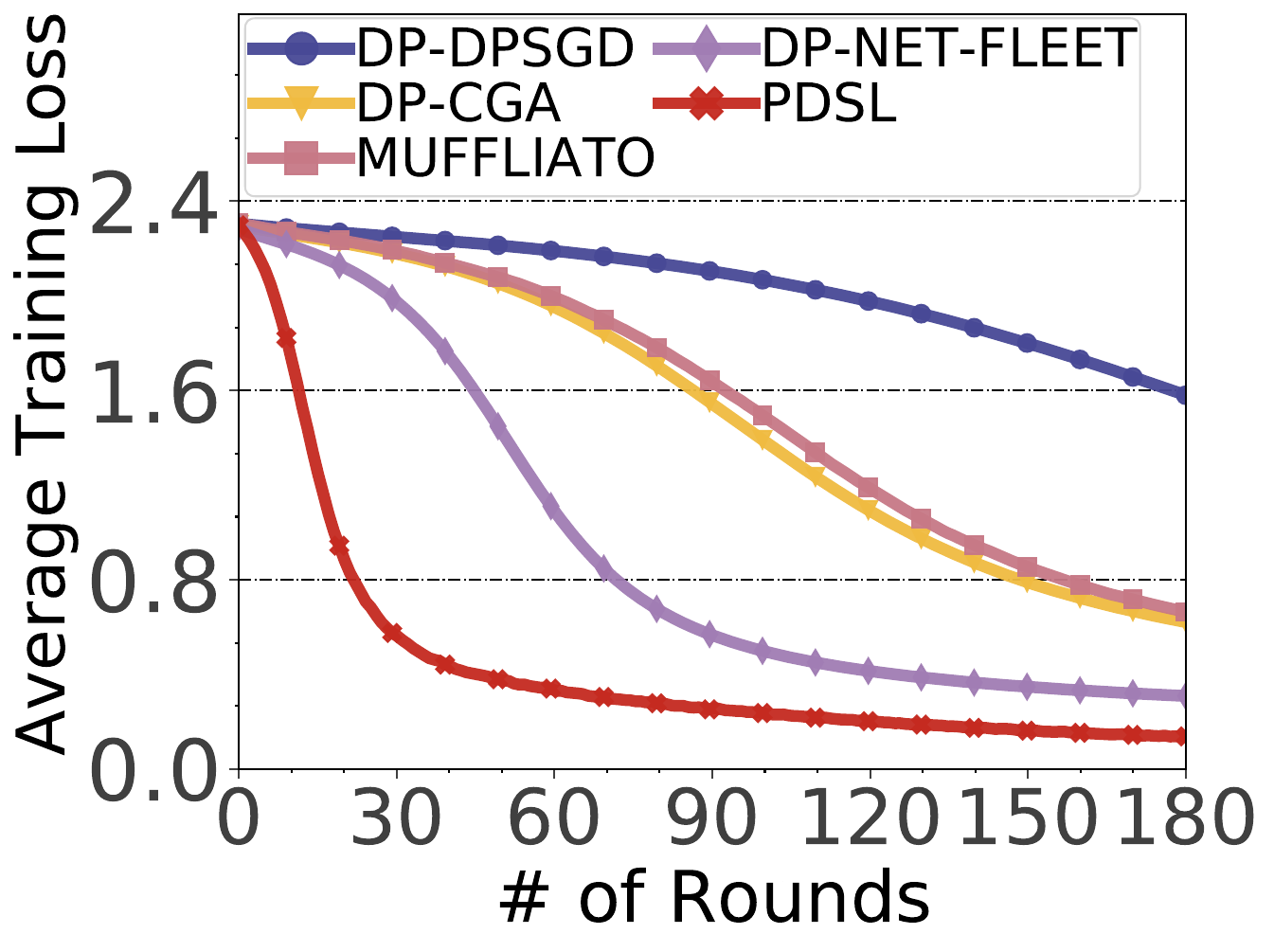}}
            \parbox{.3\textwidth}{\center\scriptsize(c1) $\epsilon=0.08$ with $M=20$}
            \parbox{.3\textwidth}{\center\scriptsize(c2) $\epsilon=0.1$ with $M=20$}
            \parbox{.3\textwidth}{\center\scriptsize(c3) $\epsilon=0.3$ with $M=20$}
          \caption{Comparison results on MNIST dataset over bipartite graphs.}
          \label{fig:mnist-loss-bipartite}
          \end{center}
       \end{figure*}

\begin{figure*}[htb!]
          \begin{center}
            \parbox{.3\textwidth}{\center\includegraphics[width=.25\textwidth]{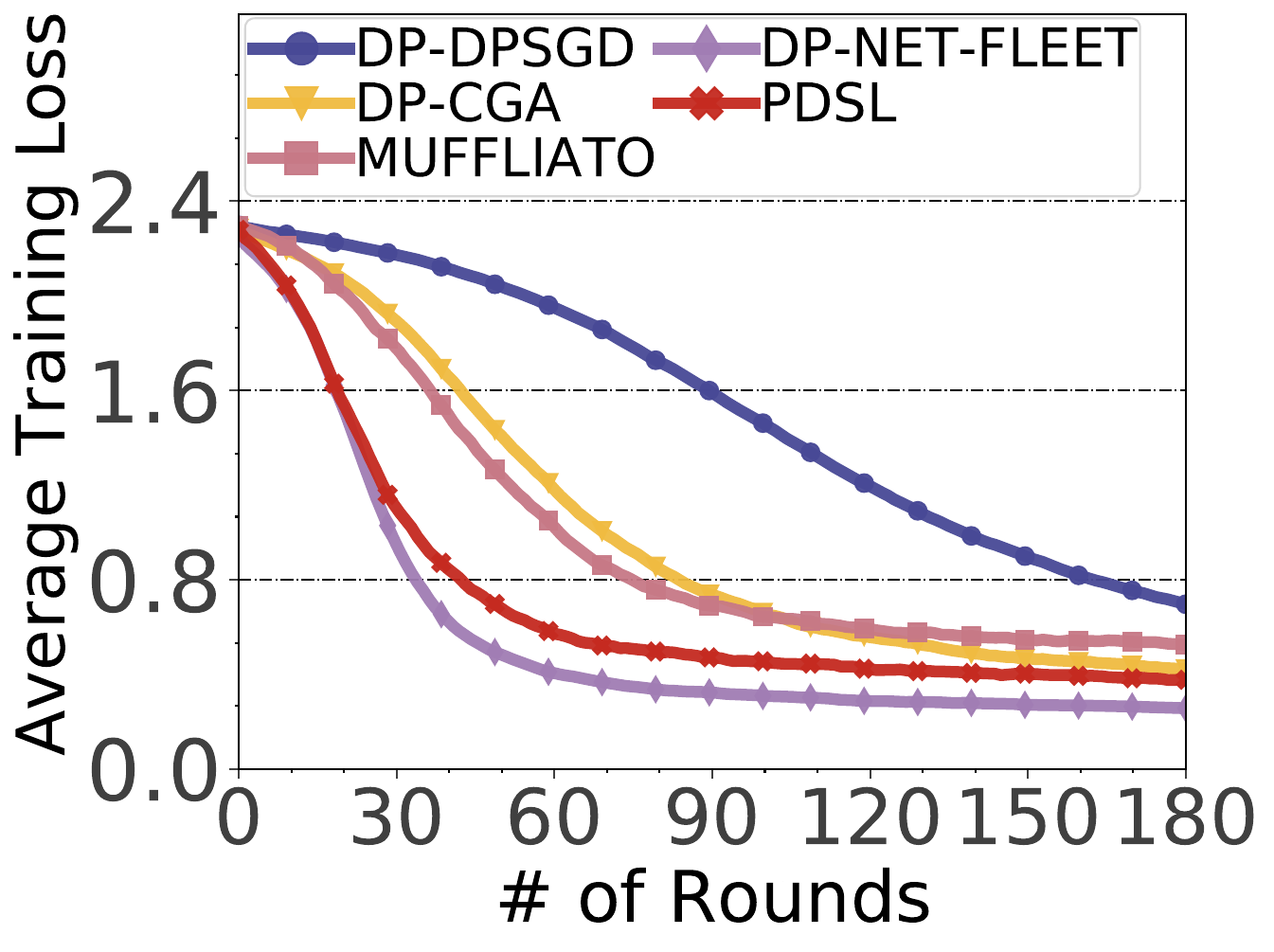}}
            \parbox{.3\textwidth}{\center\includegraphics[width=.25\textwidth]{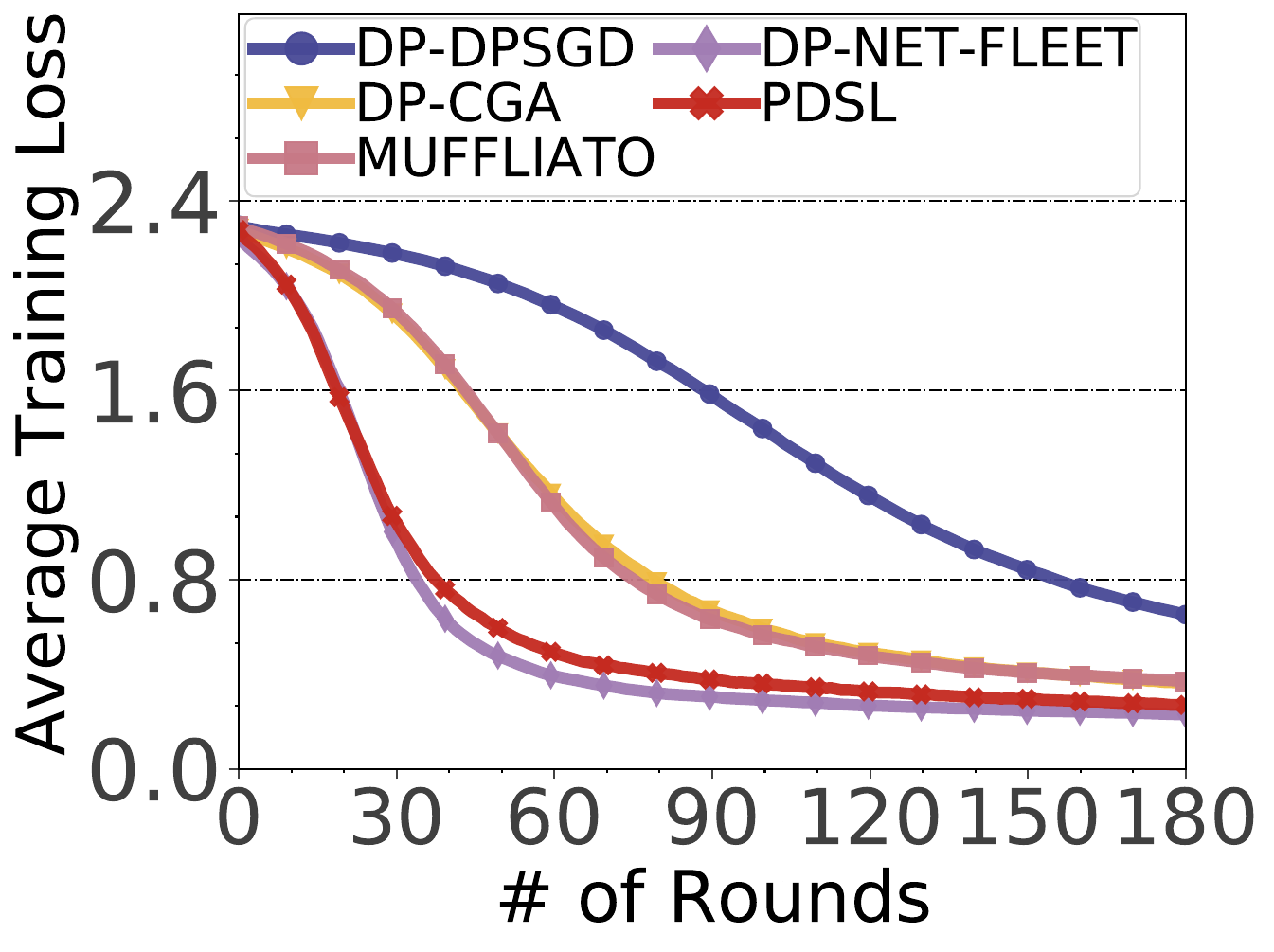}}
            \parbox{.3\textwidth}{\center\includegraphics[width=.25\textwidth]{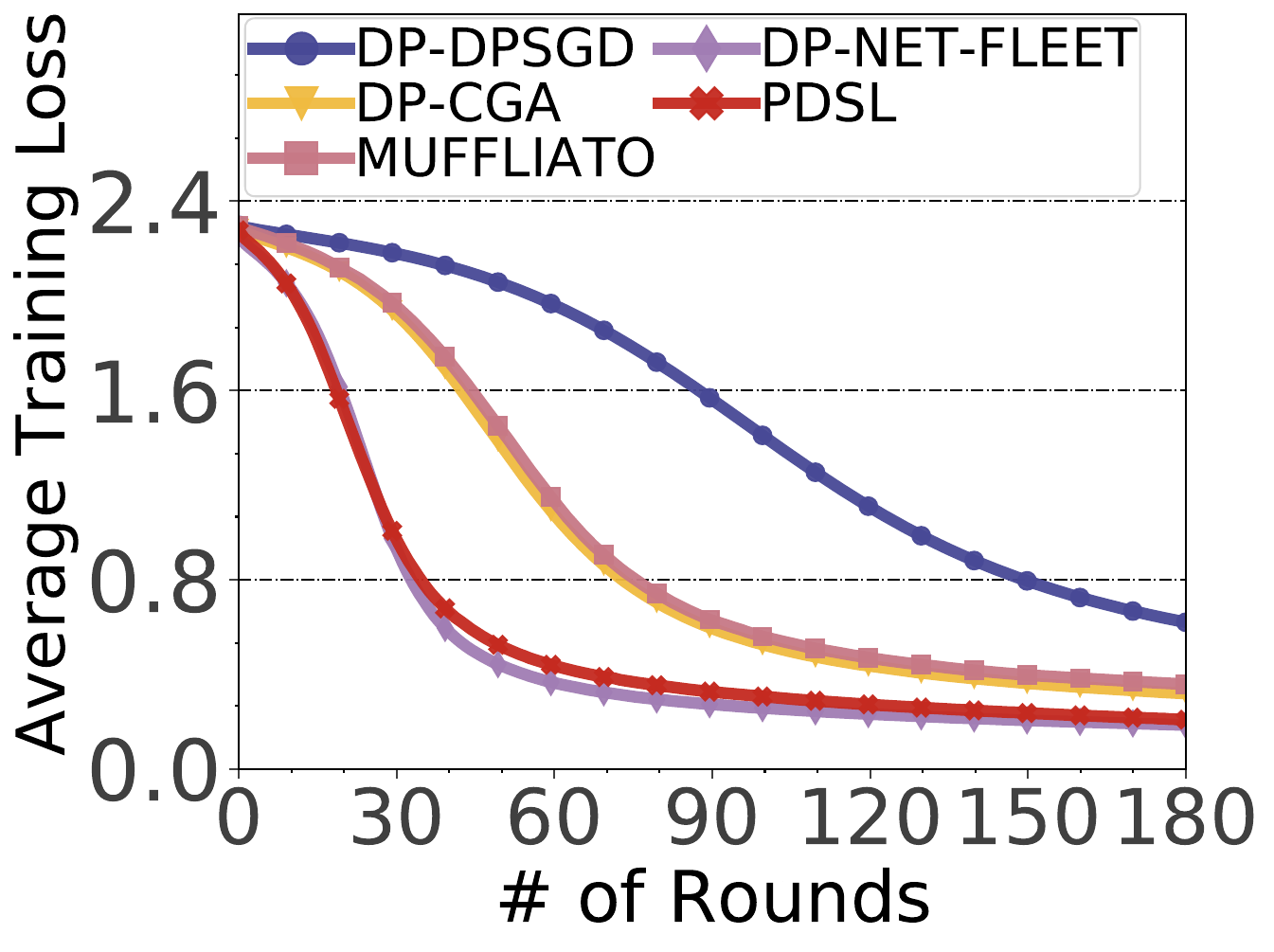}}
            \parbox{.3\textwidth}{\center\scriptsize(a1) $\epsilon=0.08$ with $M=10$}
            \parbox{.3\textwidth}{\center\scriptsize(a2) $\epsilon=0.1$ with $M=10$}
            \parbox{.3\textwidth}{\center\scriptsize(a3) $\epsilon=0.3$ with $M=10$}
            \parbox{.3\textwidth}{\center\includegraphics[width=.25\textwidth]{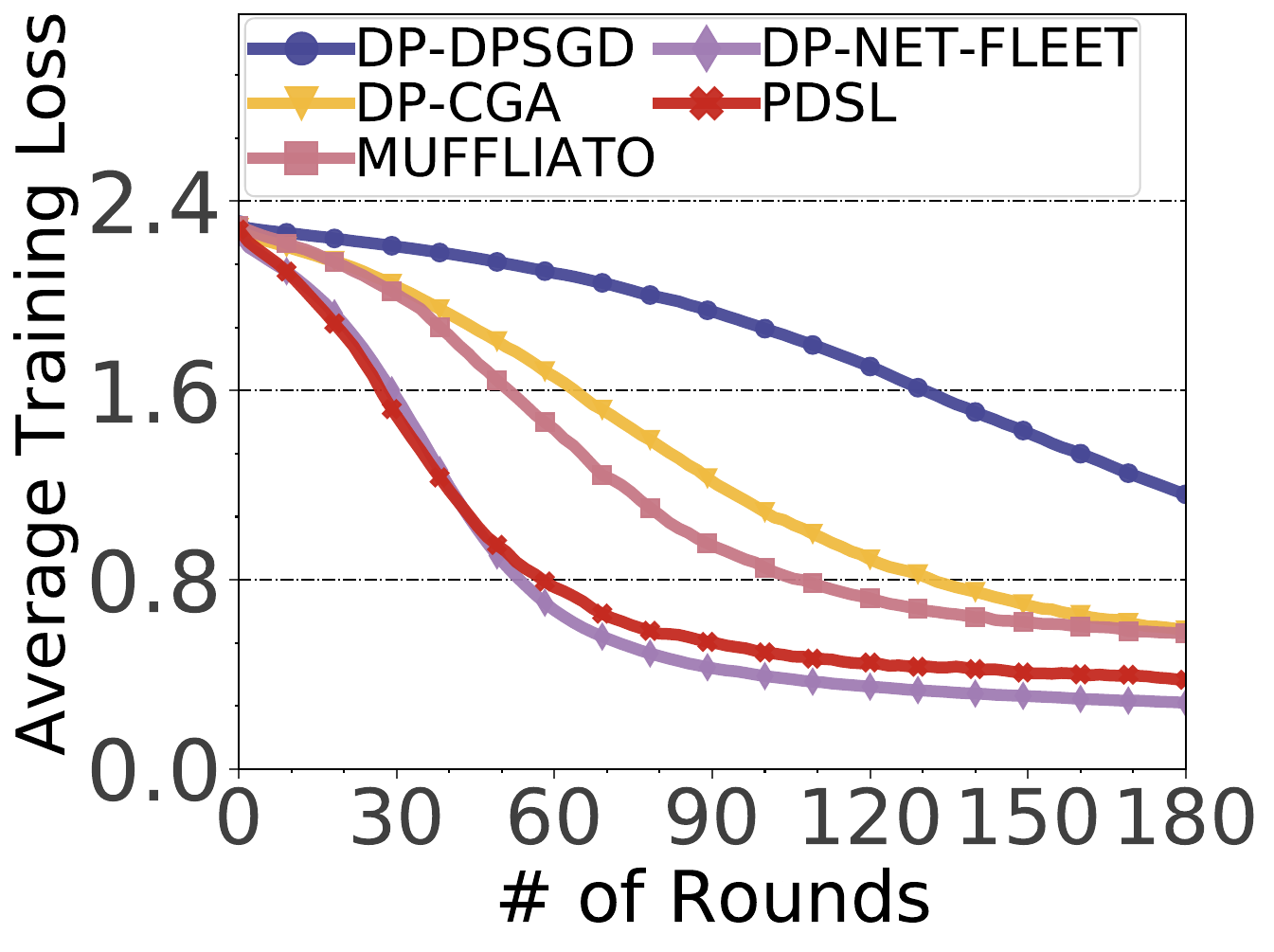}}
            \parbox{.3\textwidth}{\center\includegraphics[width=.25\textwidth]{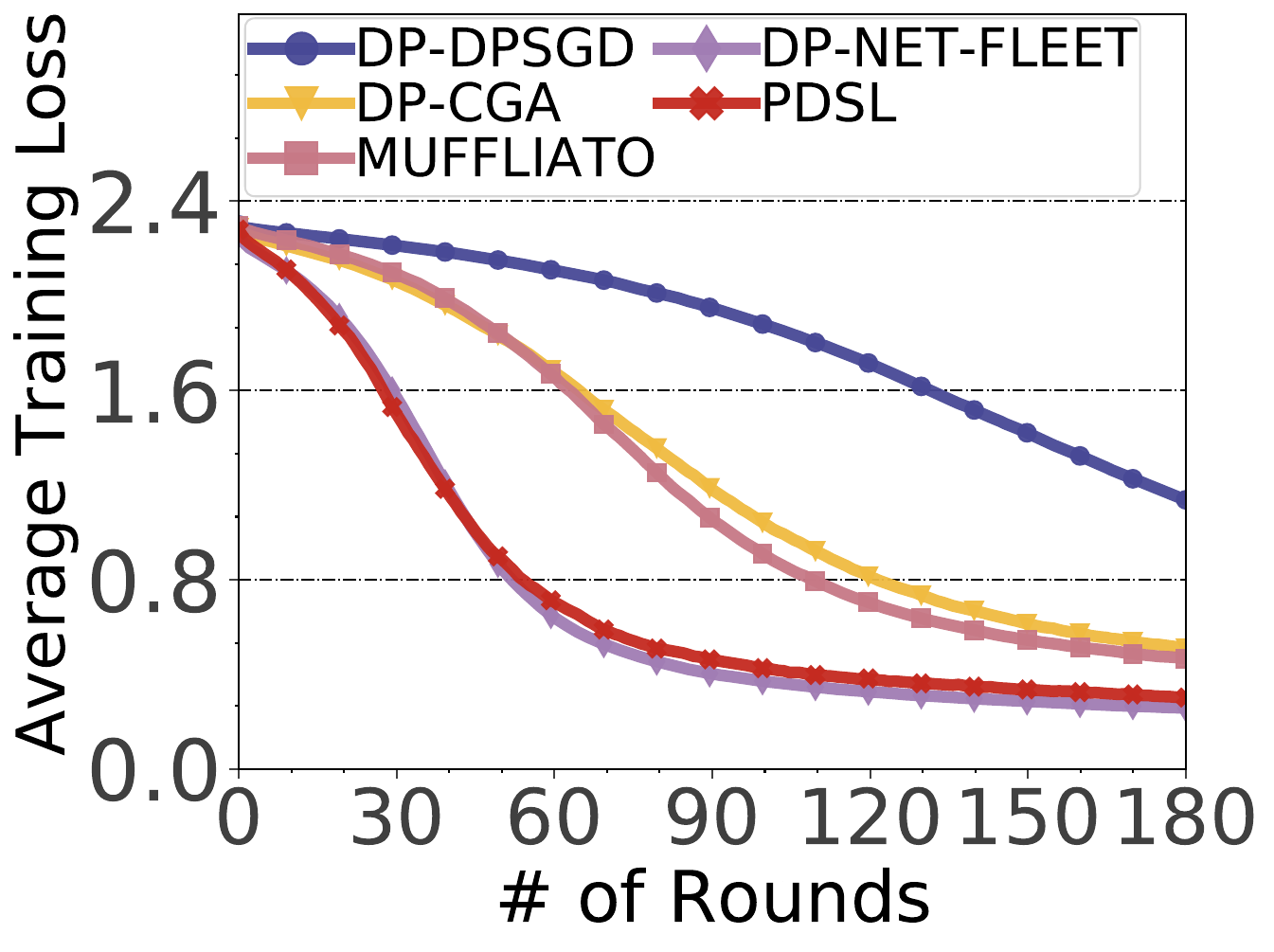}}
            \parbox{.3\textwidth}{\center\includegraphics[width=.25\textwidth]{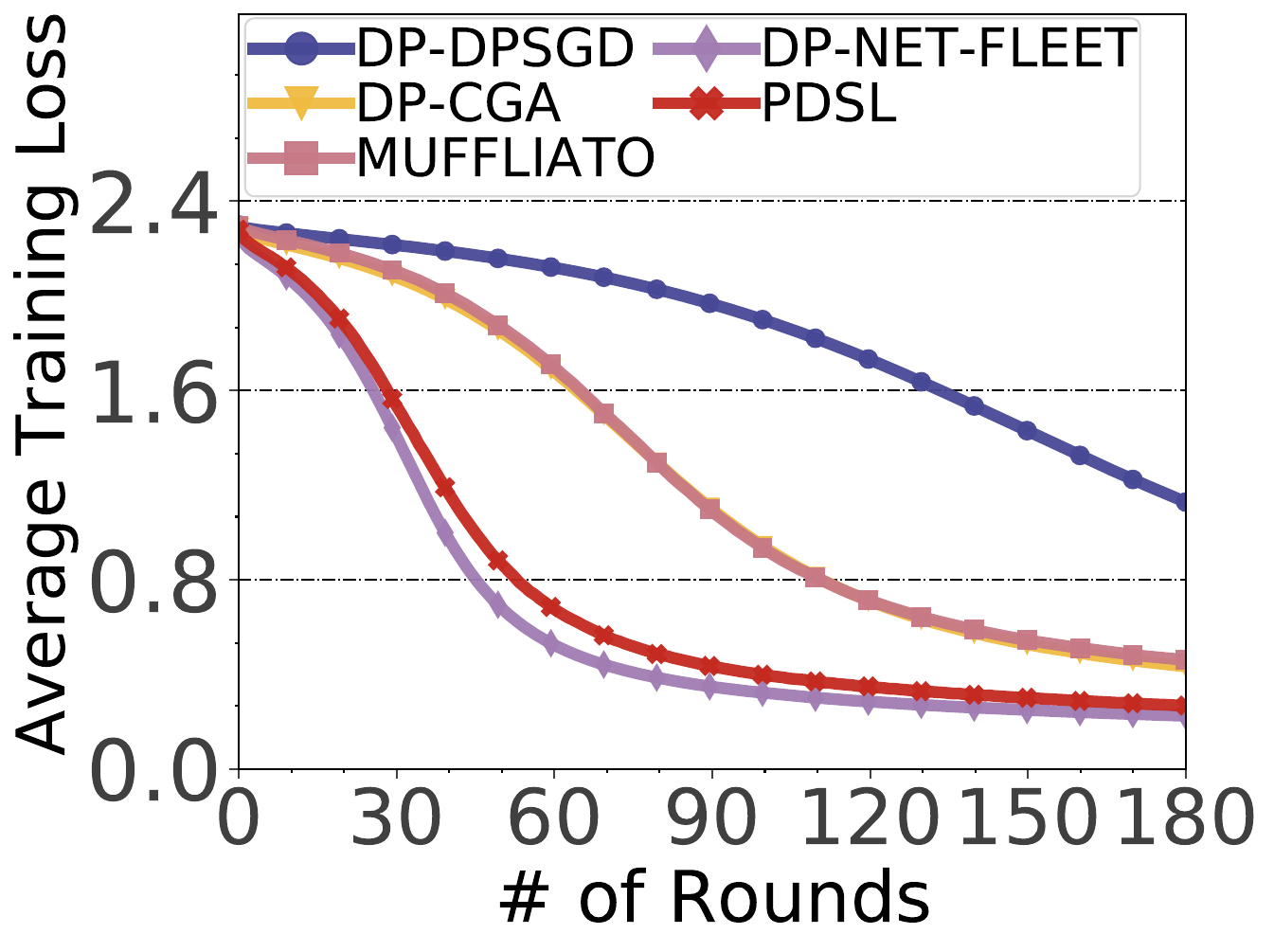}}
            \parbox{.3\textwidth}{\center\scriptsize(b1) $\epsilon=0.08$ with $M=15$}
            \parbox{.3\textwidth}{\center\scriptsize(b2) $\epsilon=0.1$ with $M=15$}
            \parbox{.3\textwidth}{\center\scriptsize(b3) $\epsilon=0.3$ with $M=15$}
            \parbox{.3\textwidth}{\center\includegraphics[width=.25\textwidth]{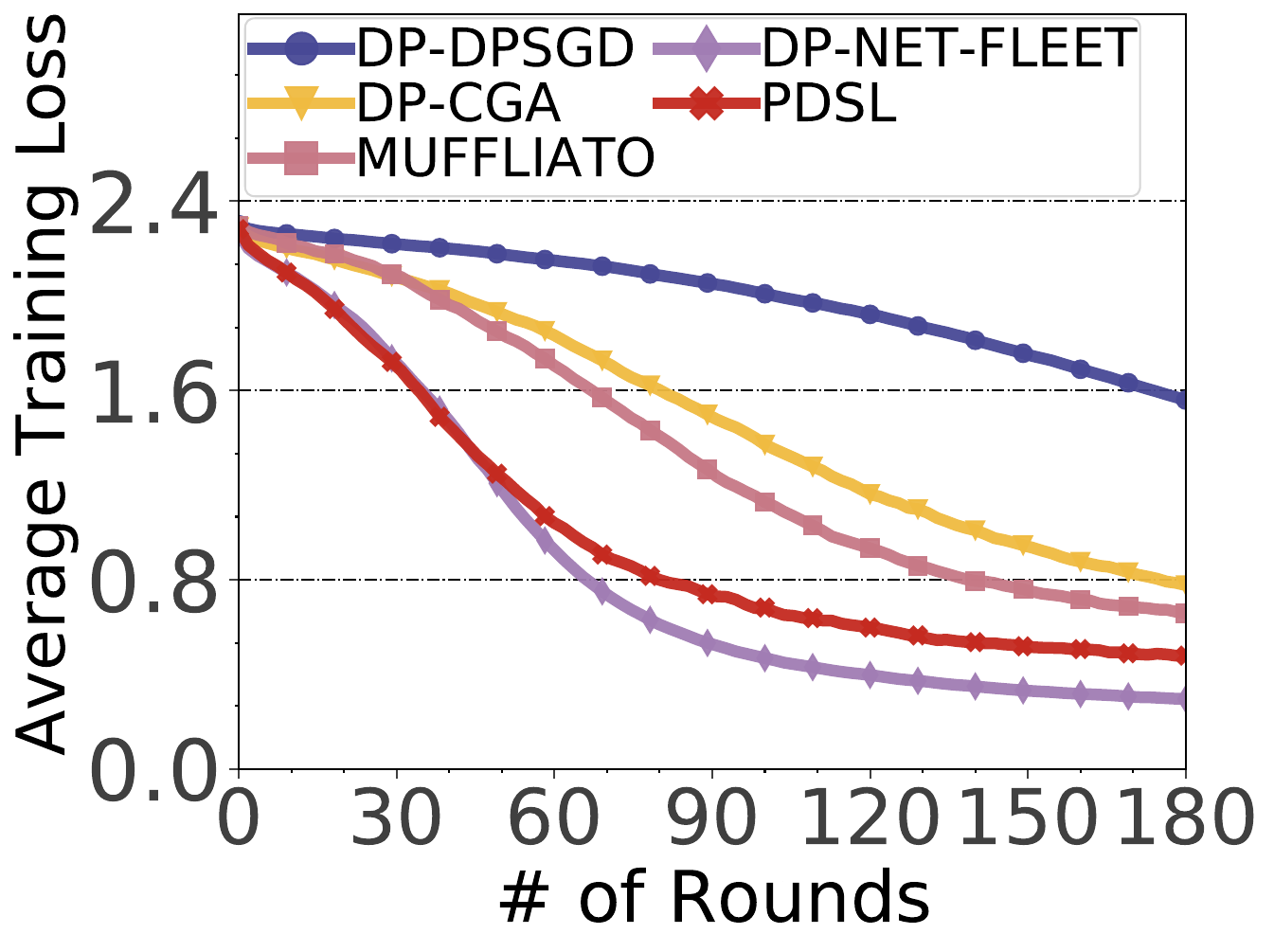}}
            \parbox{.3\textwidth}{\center\includegraphics[width=.25\textwidth]{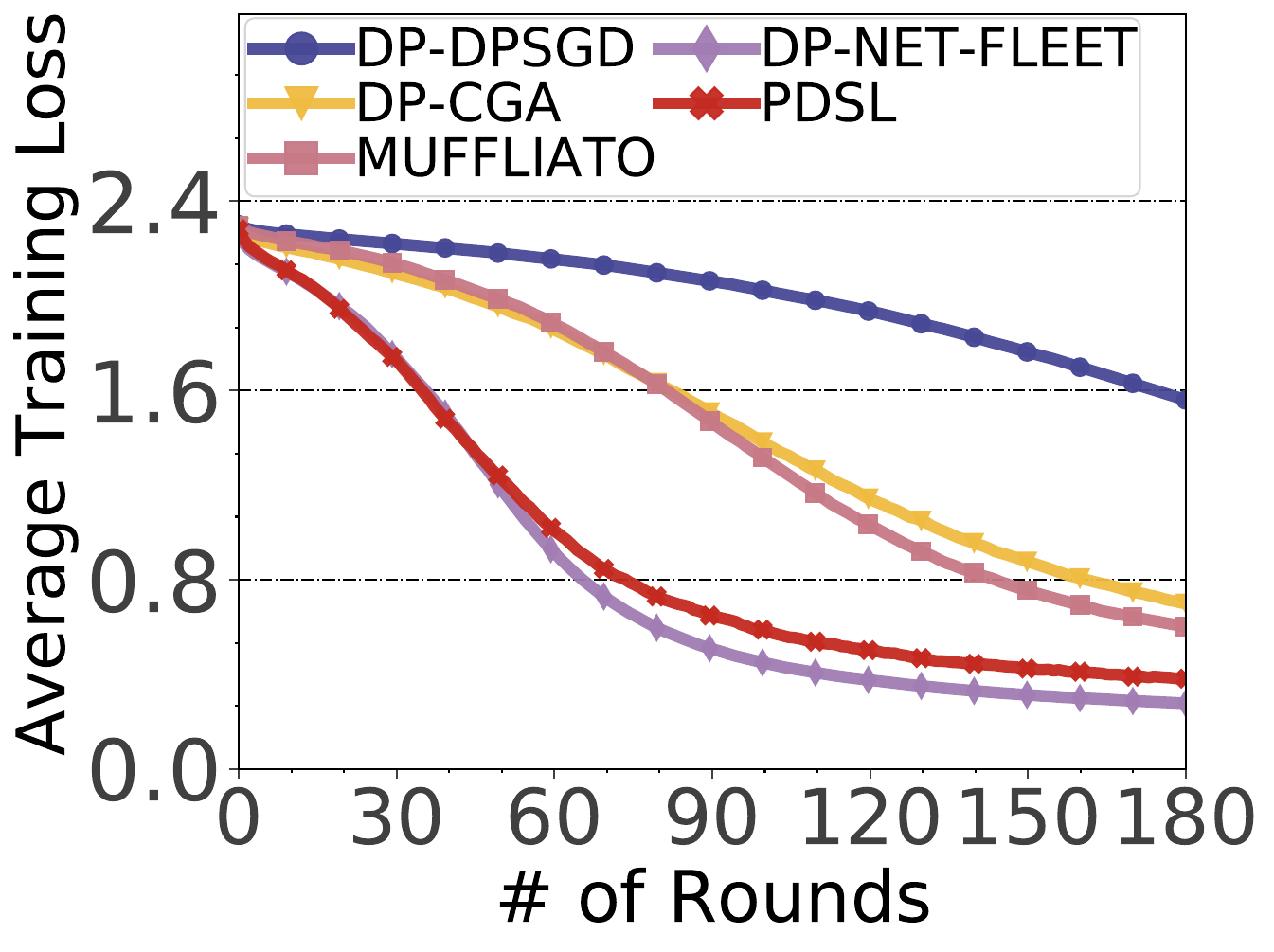}}
            \parbox{.3\textwidth}{\center\includegraphics[width=.25\textwidth]{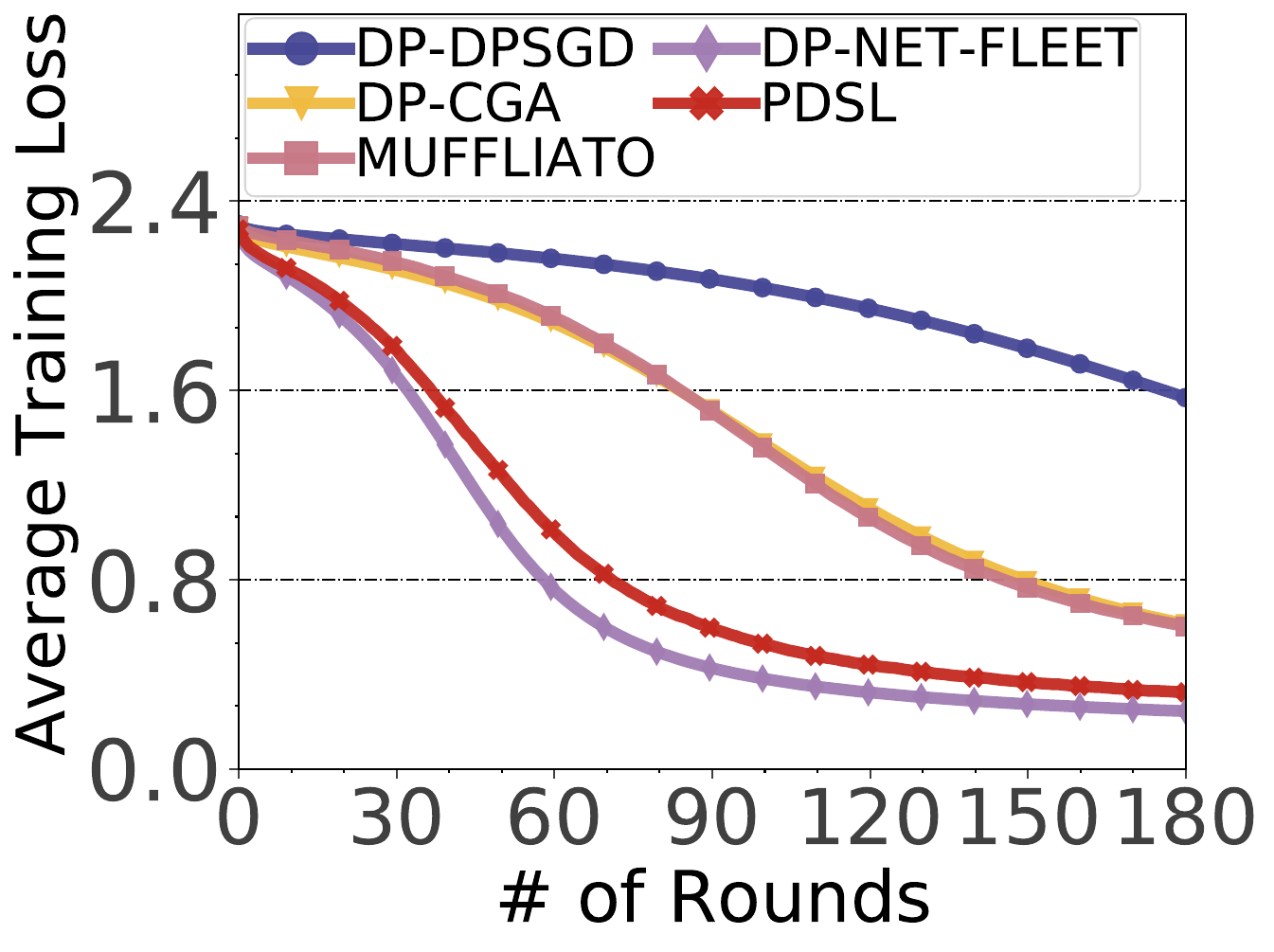}}
            \parbox{.3\textwidth}{\center\scriptsize(c1) $\epsilon=0.08$ with $M=20$}
            \parbox{.3\textwidth}{\center\scriptsize(c2) $\epsilon=0.1$ with $M=20$}
            \parbox{.3\textwidth}{\center\scriptsize(c3) $\epsilon=0.3$ with $M=20$}
          \caption{Comparison results on MNIST dataset over ring graphs.}
          \label{fig:mnist-loss-ring}
          \end{center}
       \end{figure*}

       Fig.~\ref{fig:mnist-loss-bipartite} and Fig.~\ref{fig:mnist-loss-ring} show the average losses obtained on bipartite and ring graphs, respectively. Even on the sparser communication graphs, our PDSL algorithm has smaller average loss than the other reference algorithms and converges within about $100 \sim 110$ rounds across different values of $\epsilon$. In contrast, the DP-CGA, MUFFLIATO and DP-DPSGD algorithms require much more communication rounds to achieve convergence. Moreover, in comparison to PDSL, the DP-NET-FLEET algorithm demonstrates comparable convergence performance on the ring graph. However, as will be shown later, the test accuracy of DP-NET-FLEET is much smaller.

       Table~\ref{tb:mnist-acc} presents the test accuracy of PDSL along with other reference algorithms across different values of $\epsilon$ on MNIST dataset. The results reveal that PDSL significantly outperforms the other reference algorithms. Take the performance improvement on the fully connected communication graph as an illustration. When $M=10, \epsilon = 0.08$, the test accuracy of PDSL is $0.938$; it is an improvement of $8.5\%$ over DP-DPSGD, $5.7\%$ over DP-CGA, $10\%$ over MUFFLIATO, and $2.6\%$ over DP-NET-FLEET, respectively. Moreover, as the number of agents increases, the test accuracy of PDSL remains relatively stable, in contrast to the significant decline observed in the other reference algorithms. Specifically, when $M = 20$, $\epsilon = 0.08$, the test accuracy of PDSL is $0.932$, demonstrating an improvement of around $3\%\sim26\%$ over the reference algorithms. Furthermore, as privacy budget $\epsilon$ is increased (e.g., $\epsilon$ = 0.1, 0.3), the test accuracy of PDSL increases significantly. Table~\ref{tb:mnist-acc} also reveals that the test accuracy of PDSL is better than the other reference algorithms on the different communication graphs. Especially when $M=20, \epsilon = 0.3$, the test accuracy of PDSL is $1.38$ times higher than DP-DPSGD, $1.12$ times higher than DP-CGA, $1.13$ times higher than MUFFLIATO, and $1.06$ times higher than DP-NET-FLEET on bipartite graphs; On the ring graph, the test accuracy of PDSL is around $1.02 \sim 1.36$ times higher than the those of the reference algorithms.
       \begin{table*}
          \caption{Test accuracy on MNIST dataset.}
          \label{tb:mnist-acc}
          \centering
          \setlength{\tabcolsep}{1.2mm}
          \renewcommand{\arraystretch}{1.1}
          \begin{tabular}{c|c|c|c|c|c|c|c|c|c|c} 
            \hline
            \multirow{2}{*}{Privacy budget $\epsilon$}  & \multirow{2}{*}{Methods} 
            & \multicolumn{3}{c|}{Fully-connected} & \multicolumn{3}{c|}{Bipartite} & \multicolumn{3}{c}{Ring}  \\
            \cline{3-11}
            &  & $M=10$    & $M=15$     & $M=20$     
               & $M=10$   & $M=15$    & $M=20$   
               & $M=10$  & $M=15$  & $M=20$   \\
            \hline
            \multirow{5}{*}{$\epsilon=0.08$}  
            & DP-DPSGD   & 0.853  & 0.768  & 0.668  & 0.822  & 0.739  & 0.655  & 0.817 & 0.731  & 0.604   \\
            & DP-CGA  & 0.881  & 0.843  & 0.817  & 0.865  & 0.854  & 0.779  & 0.857  & 0.799  & 0.715     \\
            & MUFFLIATO  & 0.838   & 0.802  & 0.788  & 0.833  & 0.815  & 0.782  & 0.825  & 0.805   & 0.762   \\
            & DP-NET-FLEET  & 0.912  & 0.907  & 0.902  & 0.892  & 0.891  & 0.872  & 0.872  & 0.868  & 0.842  \\
            & \textbf{PDSL}  & \textbf{0.938}   
                    & \textbf{0.934}  
                    & \textbf{0.932}  
                    & \textbf{0.925}  
                    & \textbf{0.923}  
                    & \textbf{0.917}  
                    & \textbf{0.884}  
                    & \textbf{0.875}  
                    & \textbf{0.849}     \\
            \hline
            \multirow{5}{*}{$\epsilon=0.1$} 
             & DP-DPSGD  & 0.865  & 0.794   & 0.705  & 0.838  & 0.761  & 0.675  & 0.833  & 0.753  & 0.642   \\
            & DP-CGA  & 0.902  & 0.869  & 0.851  & 0.887  & 0.869  & 0.813  & 0.881  & 0.829  & 0.769   \\
            & MUFFLIATO   & 0.887   & 0.879  & 0.846  & 0.885  & 0.872  & 0.841  & 0.887  & 0.868   & 0.835   \\
            & DP-NET-FLEET  & 0.925  & 0.921  & 0.914  & 0.906  & 0.903  & 0.887  & 0.893  & 0.896  & 0.858   \\
            & \textbf{PDSL}  & \textbf{0.957}  
                    & \textbf{0.952}   
                    & \textbf{0.955}  
                    & \textbf{0.946}  
                    & \textbf{0.949}  
                    & \textbf{0.943}  
                    & \textbf{0.913}    
                    & \textbf{0.907}  
                    & \textbf{0.871}   \\
            \hline
            \multirow{5}{*}{$\epsilon=0.3$} 
             & DP-DPSGD  & 0.875  & 0.817  & 0.723  & 0.851  & 0.781  & 0.693  & 0.847  & 0.775   & 0.653\\
            & DP-CGA  & 0.915  & 0.903  & 0.866  & 0.904  & 0.882  & 0.852  & 0.898  & 0.865  & 0.793  \\
            & MUFFLIATO  & 0.906  & 0.889   & 0.857  & 0.899  & 0.878   & 0.848  & 0.892  & 0.874   & 0.841  \\
            & DP-NET-FLEET  & 0.938  & 0.929  & 0.922  & 0.927  & 0.917  & 0.903  & 0.912  & 0.905  & 0.875   \\
            & \textbf{PDSL}  & \textbf{0.965} 
                    & \textbf{0.963}   
                    & \textbf{0.962}  
                    & \textbf{0.965}  
                    & \textbf{0.963}  
                    & \textbf{0.961}  
                    & \textbf{0.936}    
                    & \textbf{0.918}  
                    & \textbf{0.892}   \\
            \hline
          \end{tabular}
       \end{table*}

    \subsubsection{Experiment Results on CIFAR-10 Dataset}  \label{ssec:res-cifar10}
\begin{figure*}[htb!]
          \begin{center}
            \parbox{.3\textwidth}{\center\includegraphics[width=.25\textwidth]{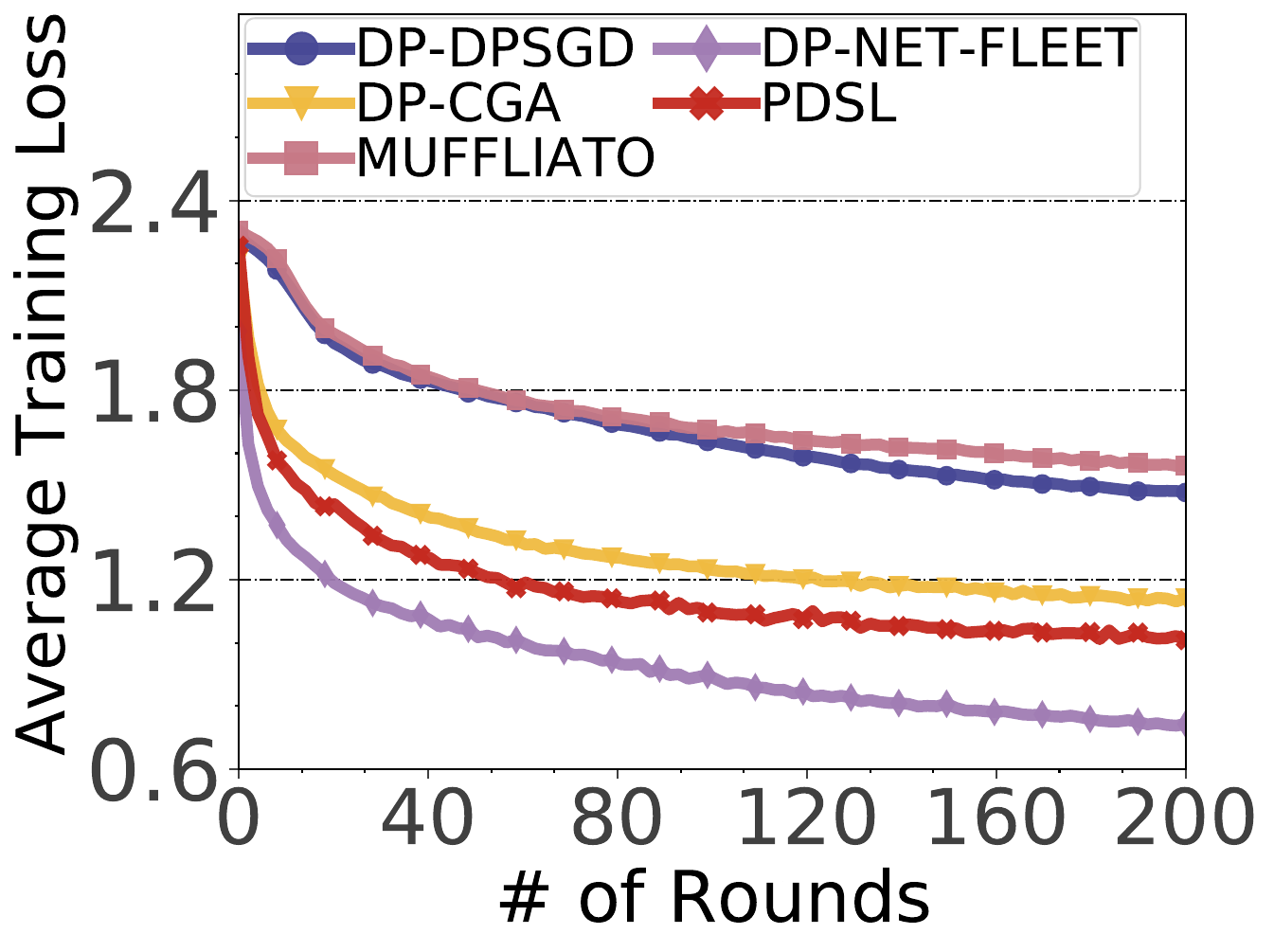}}
            \parbox{.3\textwidth}{\center\includegraphics[width=.25\textwidth]{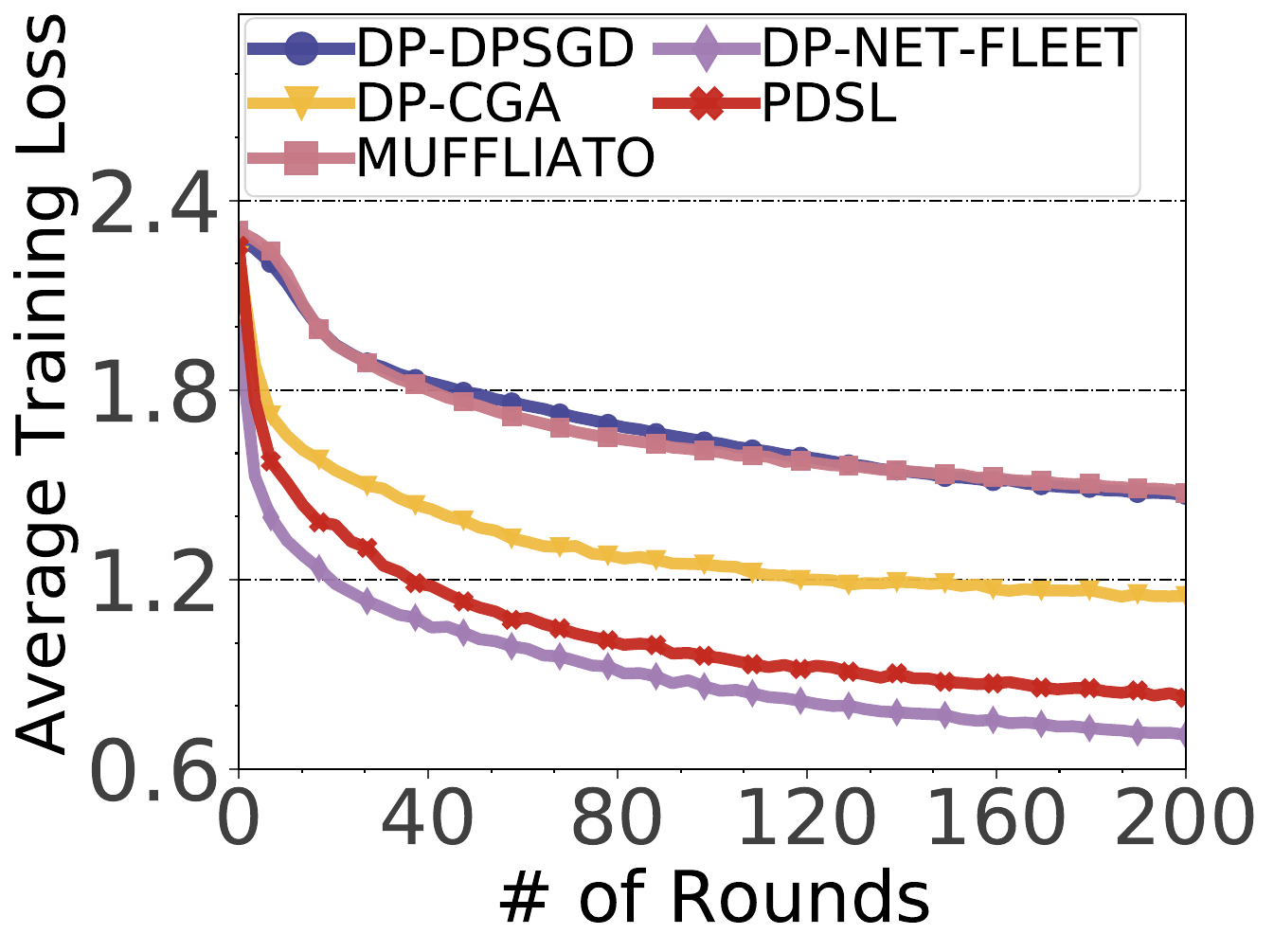}}
            \parbox{.3\textwidth}{\center\includegraphics[width=.25\textwidth]{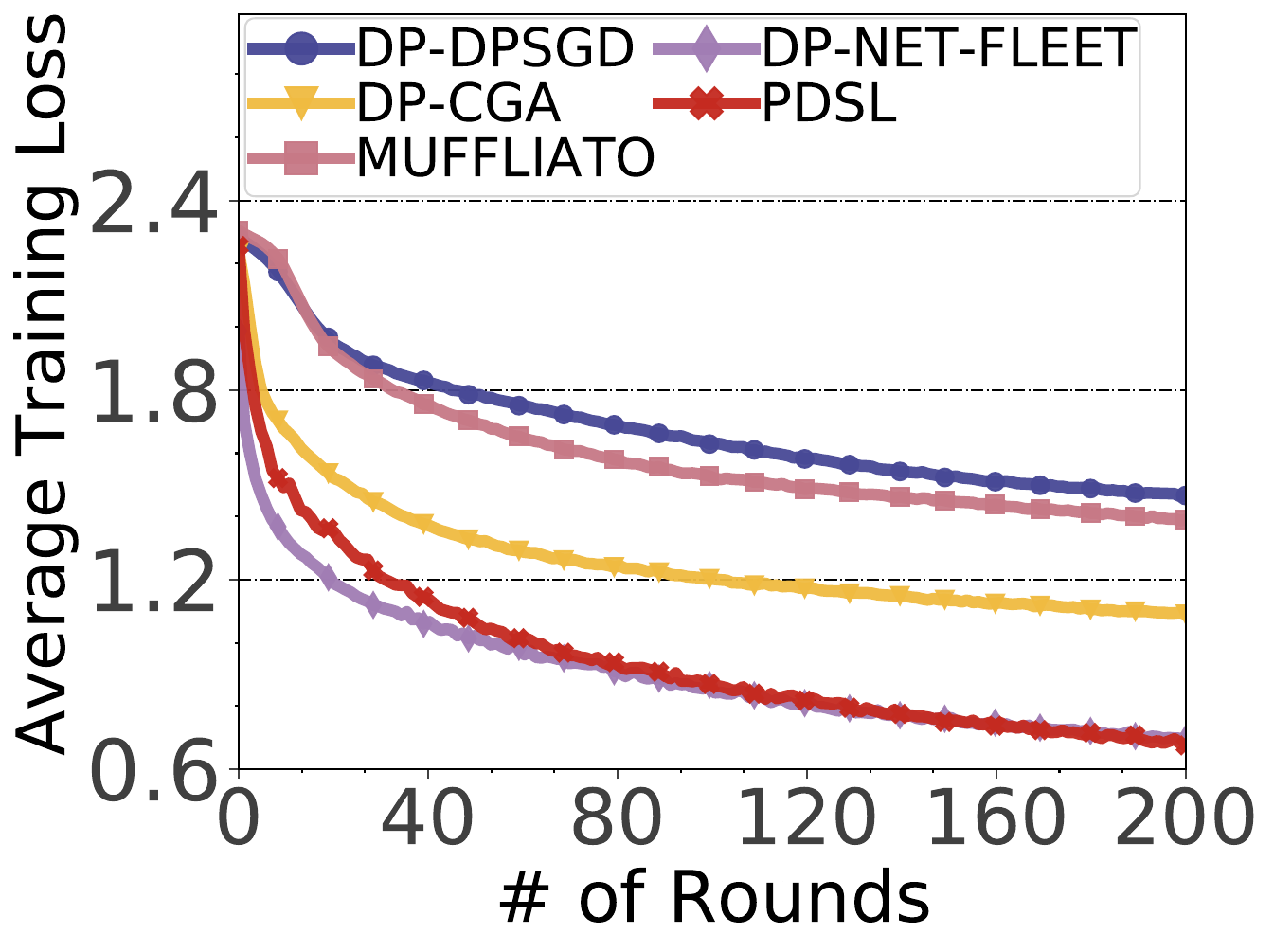}}
            \parbox{.3\textwidth}{\center\scriptsize(a1) $\epsilon=0.5$ with $M=10$}
            \parbox{.3\textwidth}{\center\scriptsize(a2) $\epsilon=0.7$ with $M=10$}
            \parbox{.3\textwidth}{\center\scriptsize(a3) $\epsilon=1$ with $M=10$}
            \parbox{.3\textwidth}{\center\includegraphics[width=.25\textwidth]{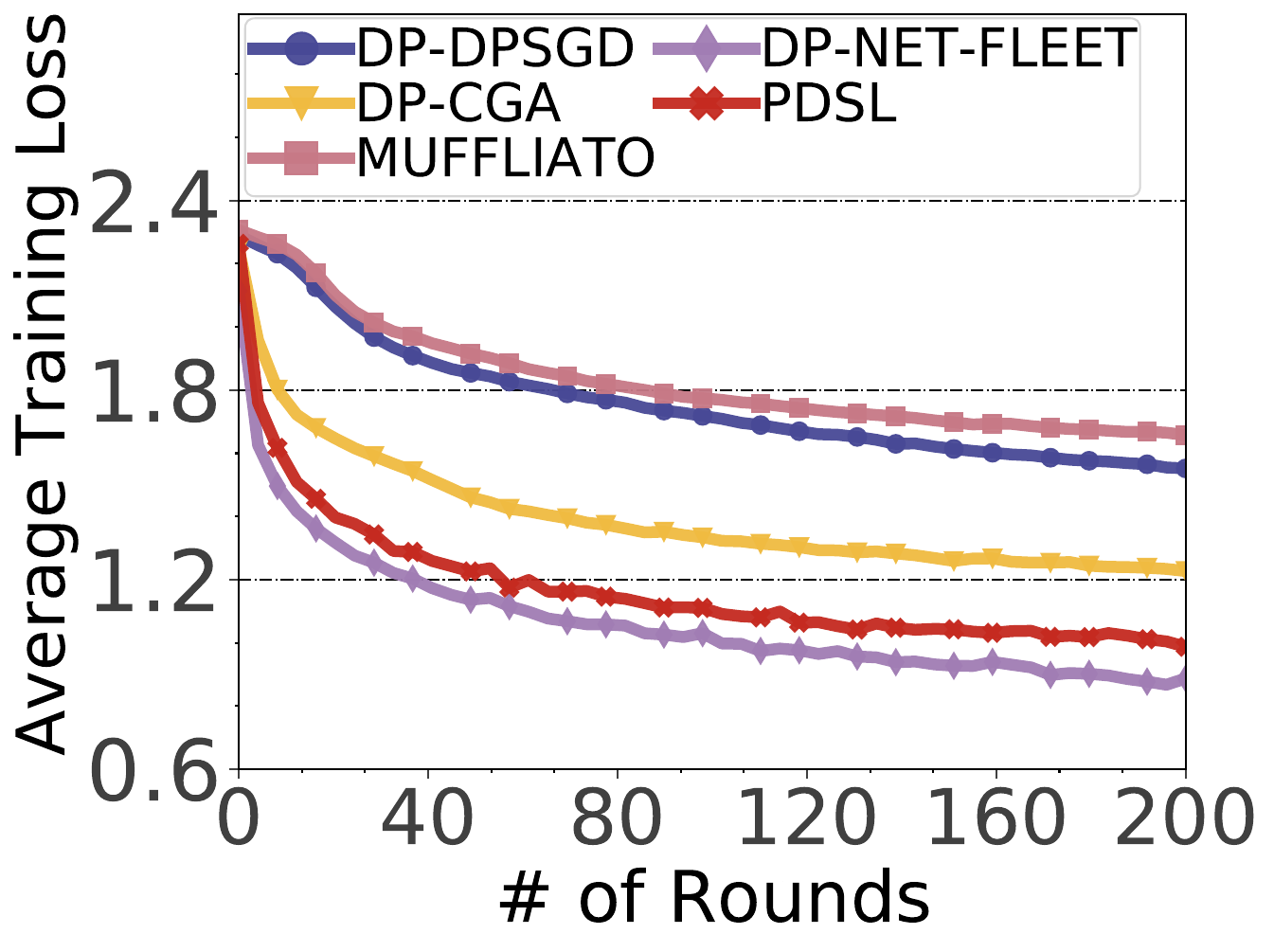}}
            \parbox{.3\textwidth}{\center\includegraphics[width=.25\textwidth]{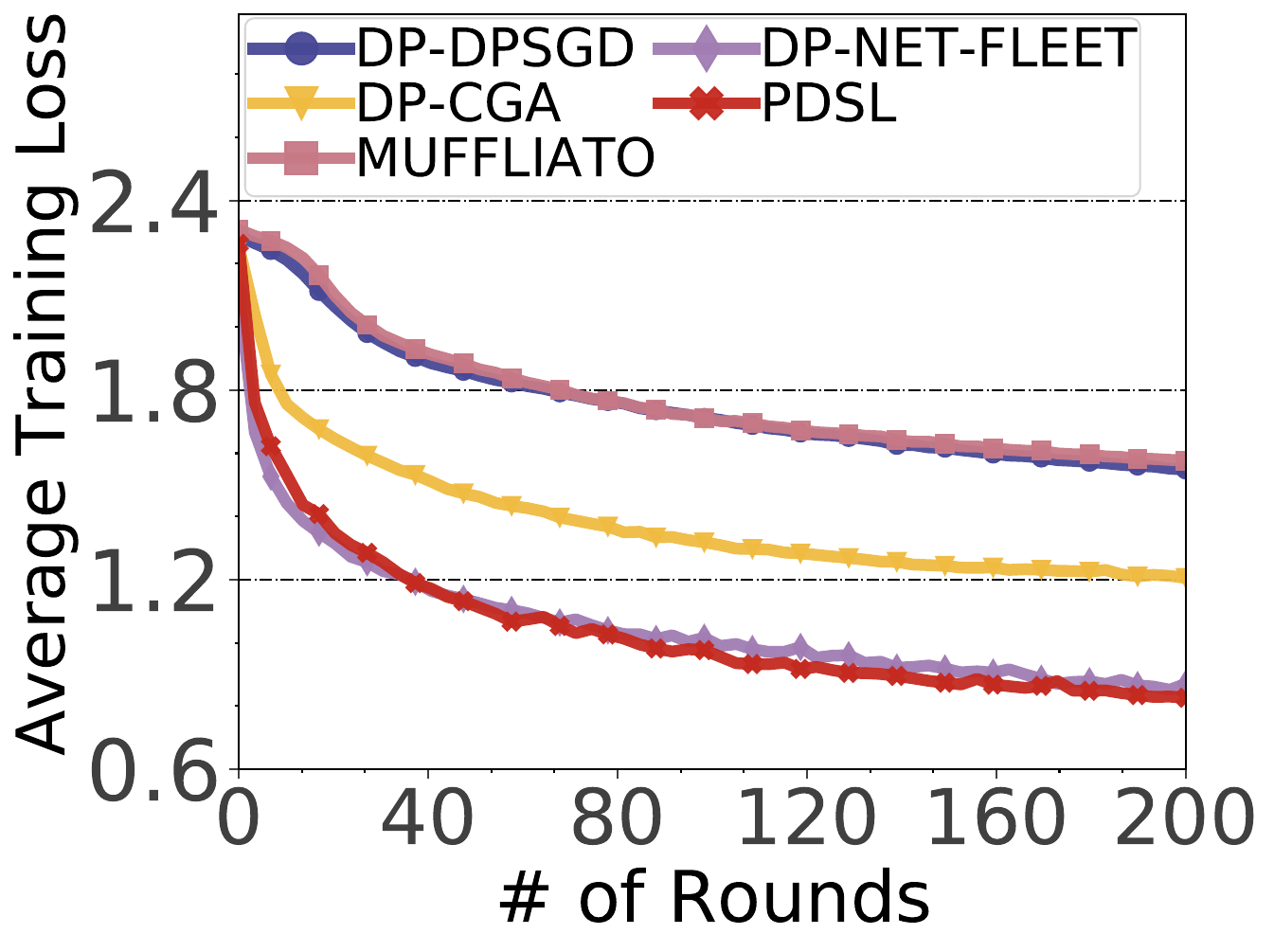}}
            \parbox{.3\textwidth}{\center\includegraphics[width=.25\textwidth]{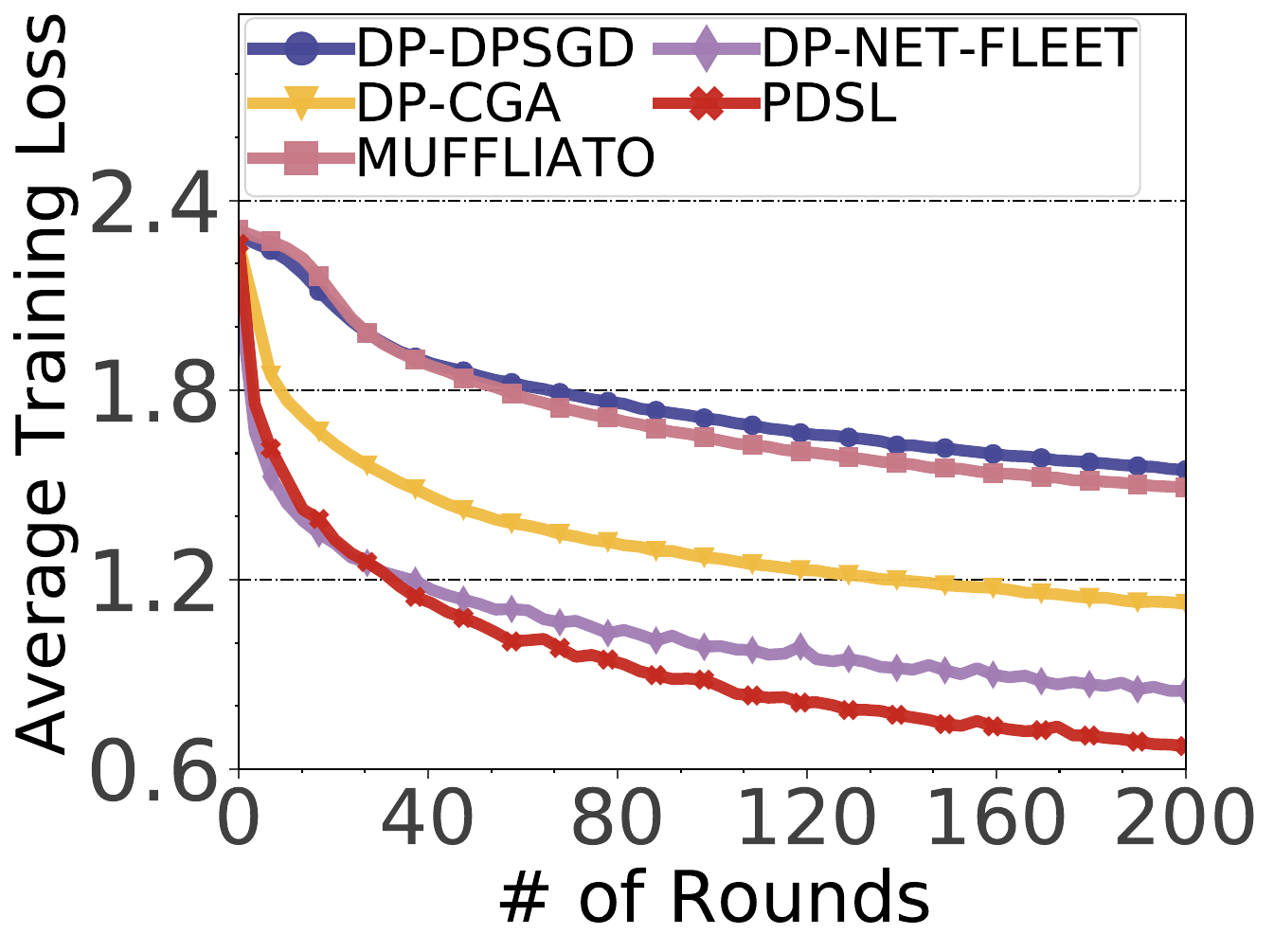}}
            \parbox{.3\textwidth}{\center\scriptsize(b1) $\epsilon=0.5$ with $M=15$}
            \parbox{.3\textwidth}{\center\scriptsize(b2) $\epsilon=0.7$ with $M=15$}
            \parbox{.3\textwidth}{\center\scriptsize(b3) $\epsilon=1$ with $M=15$}
            \parbox{.3\textwidth}{\center\includegraphics[width=.25\textwidth]{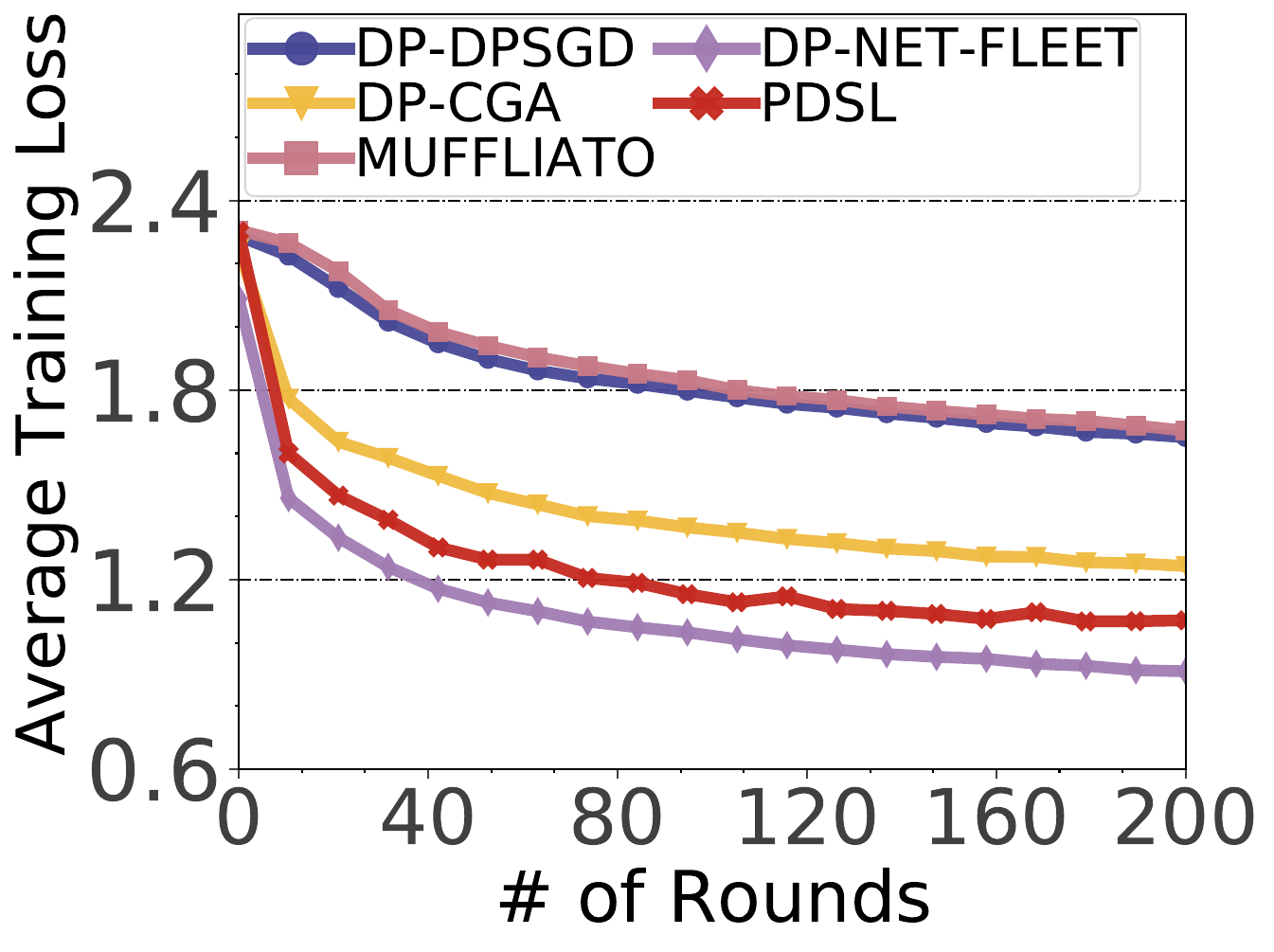}}
            \parbox{.3\textwidth}{\center\includegraphics[width=.25\textwidth]{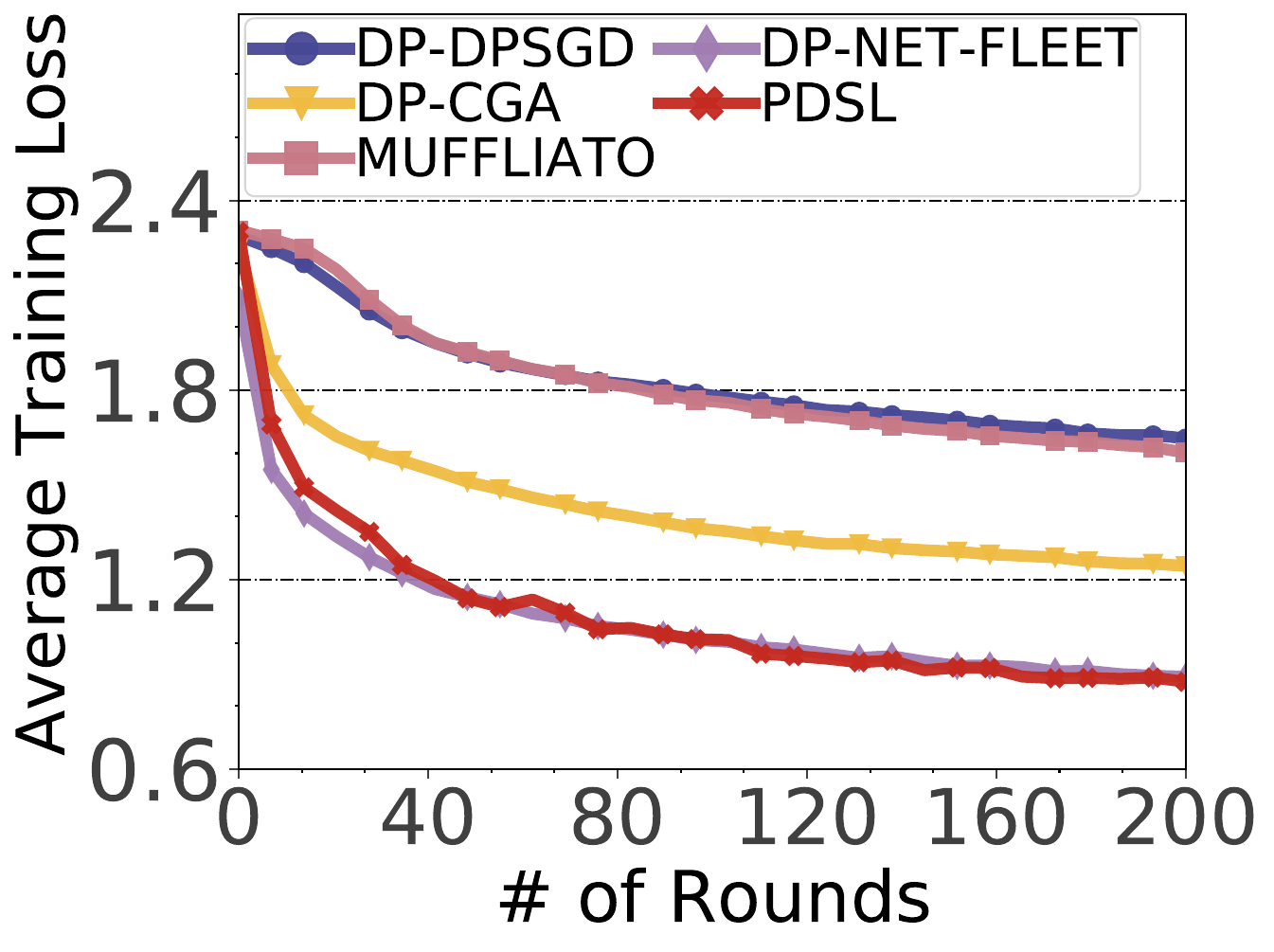}}
            \parbox{.3\textwidth}{\center\includegraphics[width=.25\textwidth]{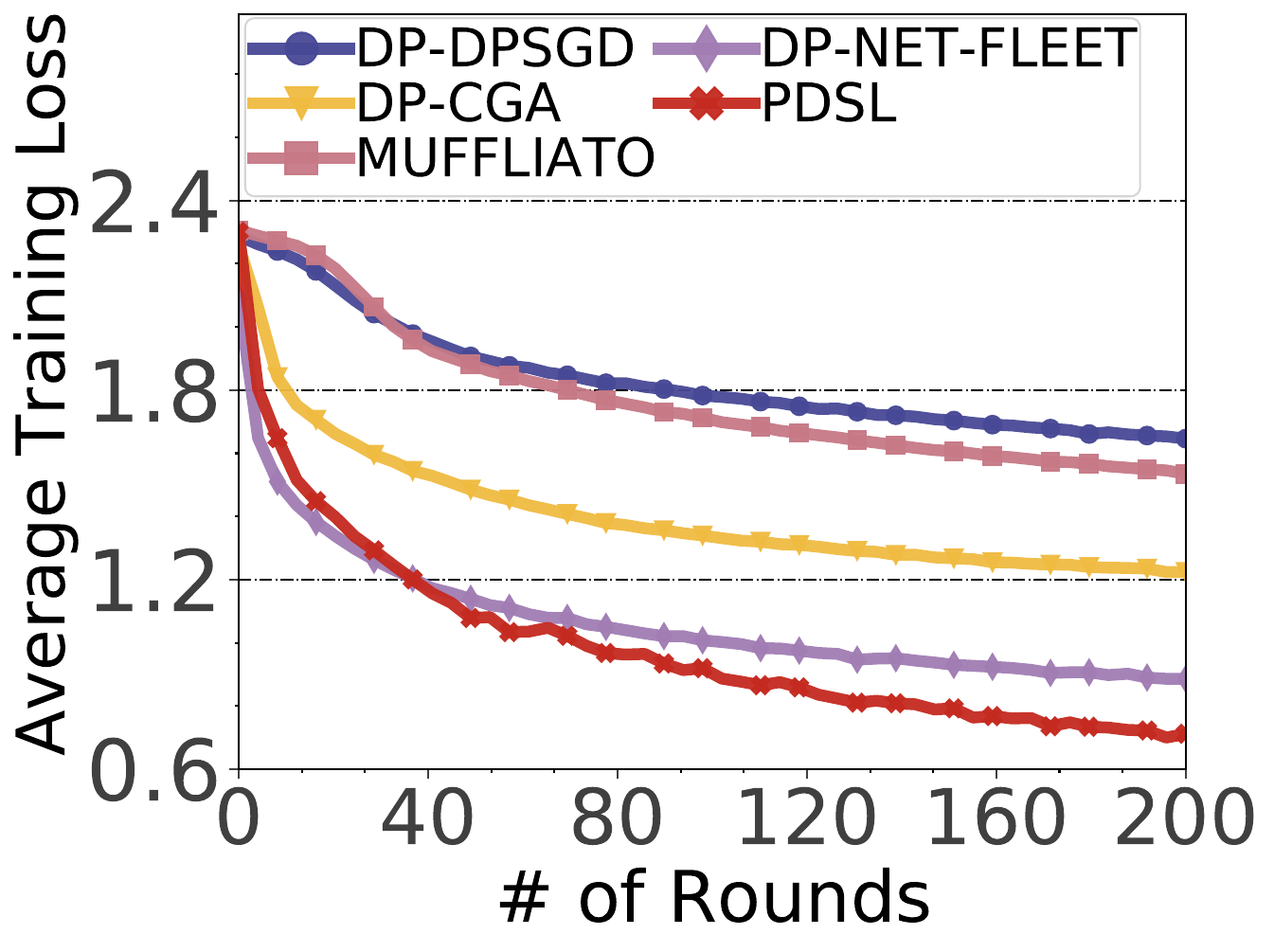}}
            \parbox{.3\textwidth}{\center\scriptsize(c1) $\epsilon=0.5$ with $M=20$}
            \parbox{.3\textwidth}{\center\scriptsize(c2) $\epsilon=0.7$ with $M=20$}
            \parbox{.3\textwidth}{\center\scriptsize(c3) $\epsilon=1$ with $M=20$}
          \caption{Comparison results on CIFAR-10 dataset over fully connected graphs.}
          \label{fig:cifar10-loss-full}
          \end{center}
       \end{figure*}

\begin{figure*}[htb!]
          \begin{center}
            \parbox{.3\textwidth}{\center\includegraphics[width=.25\textwidth]{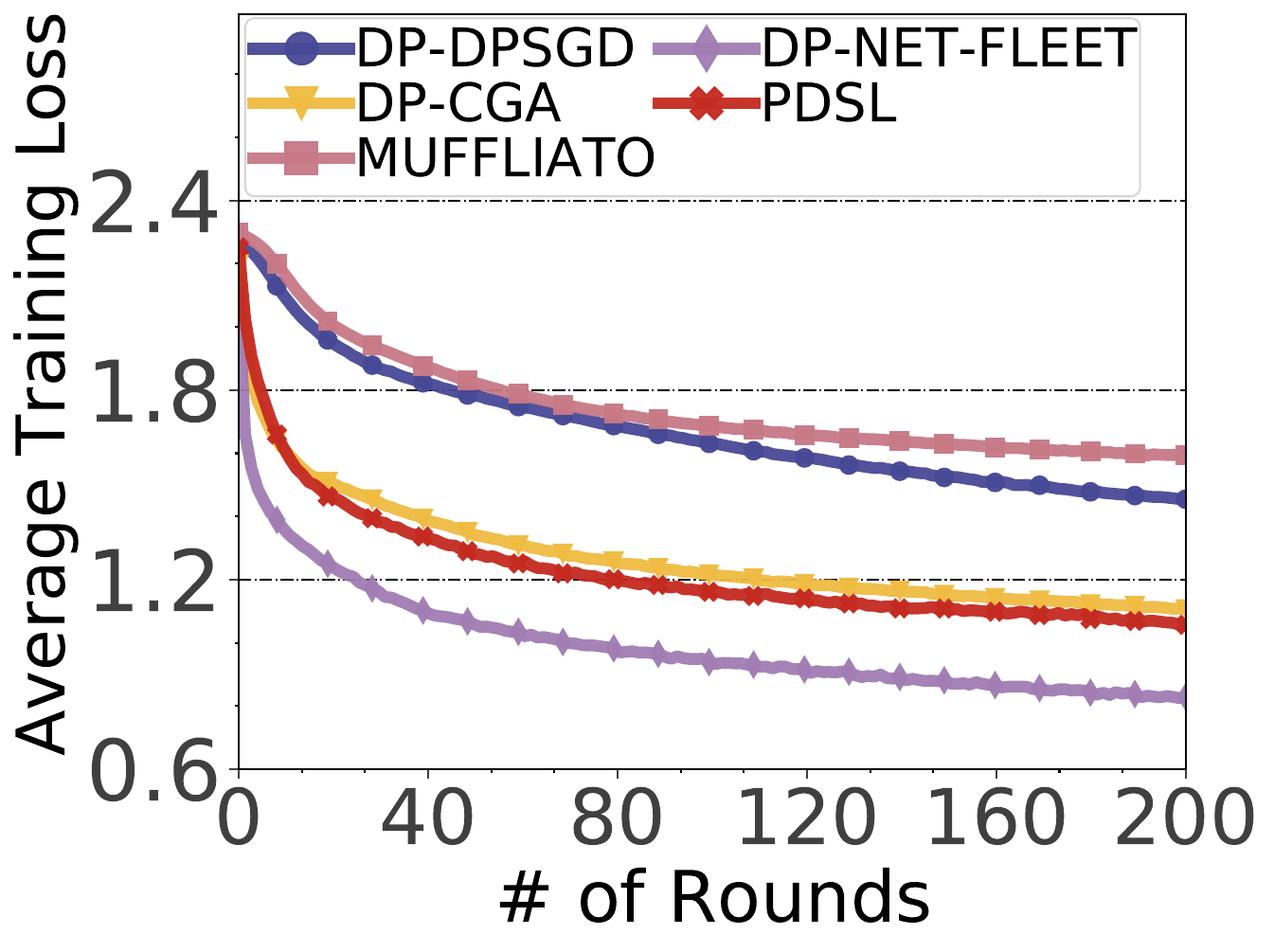}}
            \parbox{.3\textwidth}{\center\includegraphics[width=.25\textwidth]{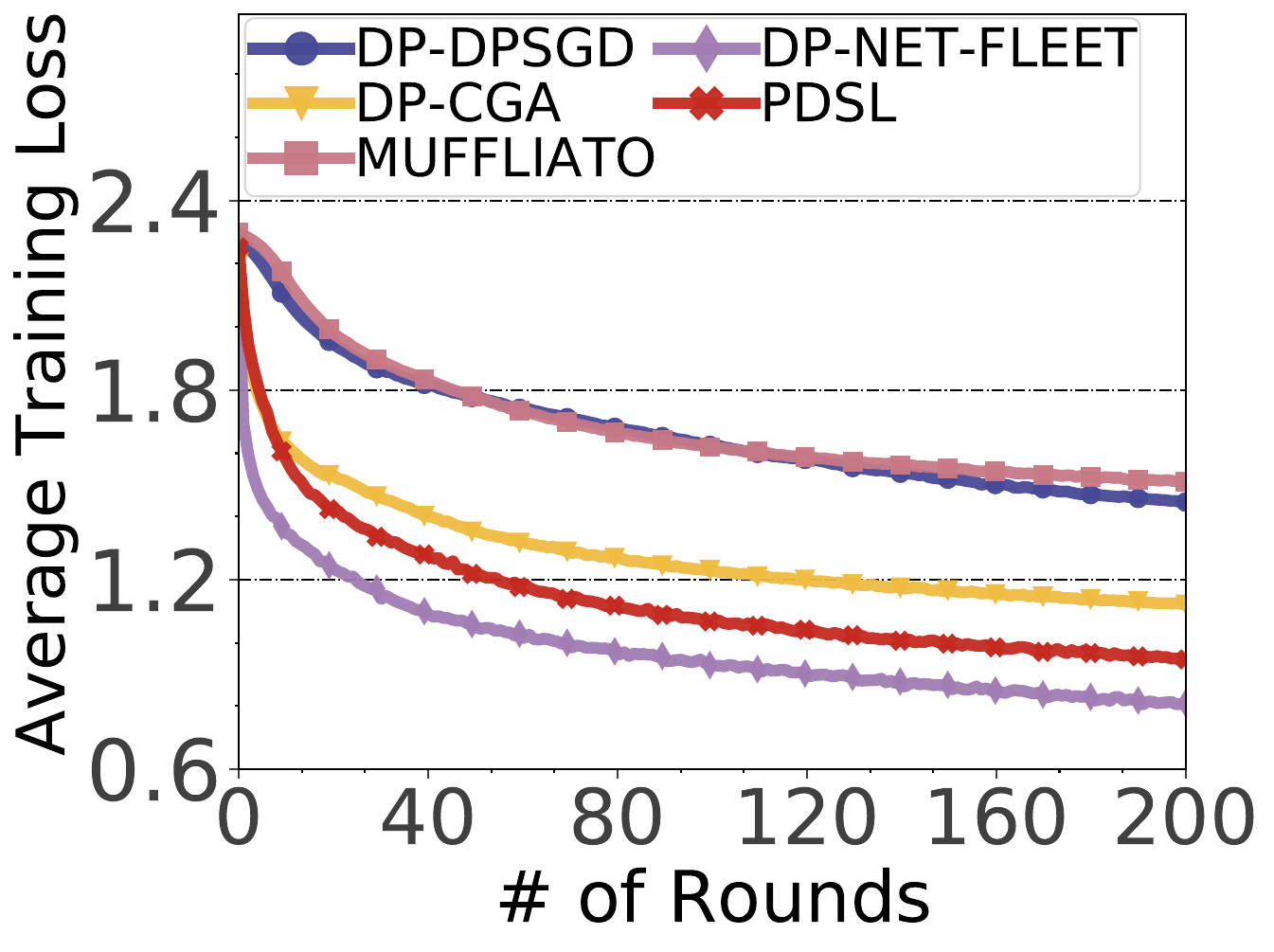}}
            \parbox{.3\textwidth}{\center\includegraphics[width=.25\textwidth]{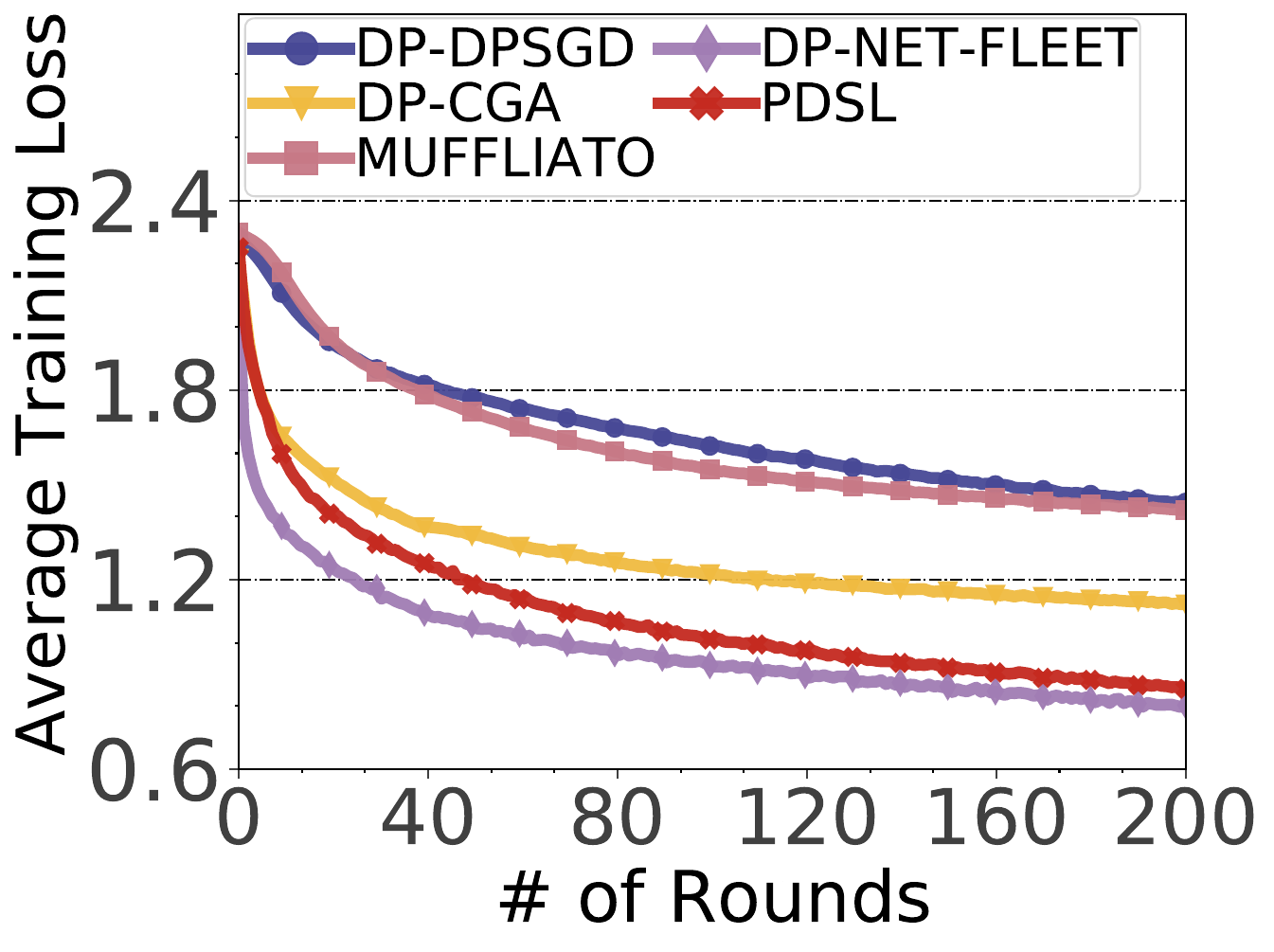}}
            \parbox{.3\textwidth}{\center\scriptsize(a1) $\epsilon=0.5$ with $M=10$}
            \parbox{.3\textwidth}{\center\scriptsize(a2) $\epsilon=0.7$ with $M=10$}
            \parbox{.3\textwidth}{\center\scriptsize(a3) $\epsilon=1$ with $M=10$}
            \parbox{.3\textwidth}{\center\includegraphics[width=.25\textwidth]{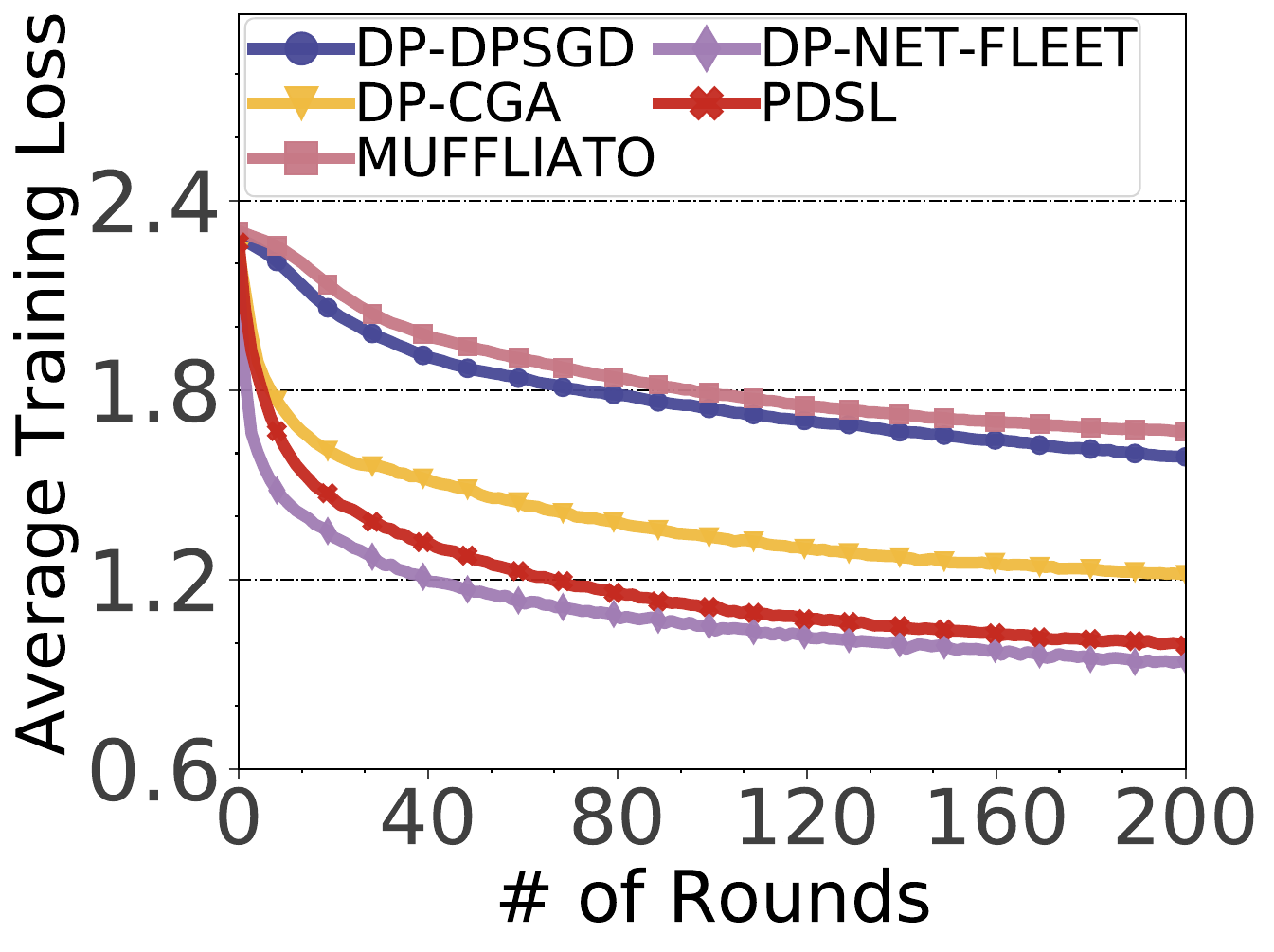}}
            \parbox{.3\textwidth}{\center\includegraphics[width=.25\textwidth]{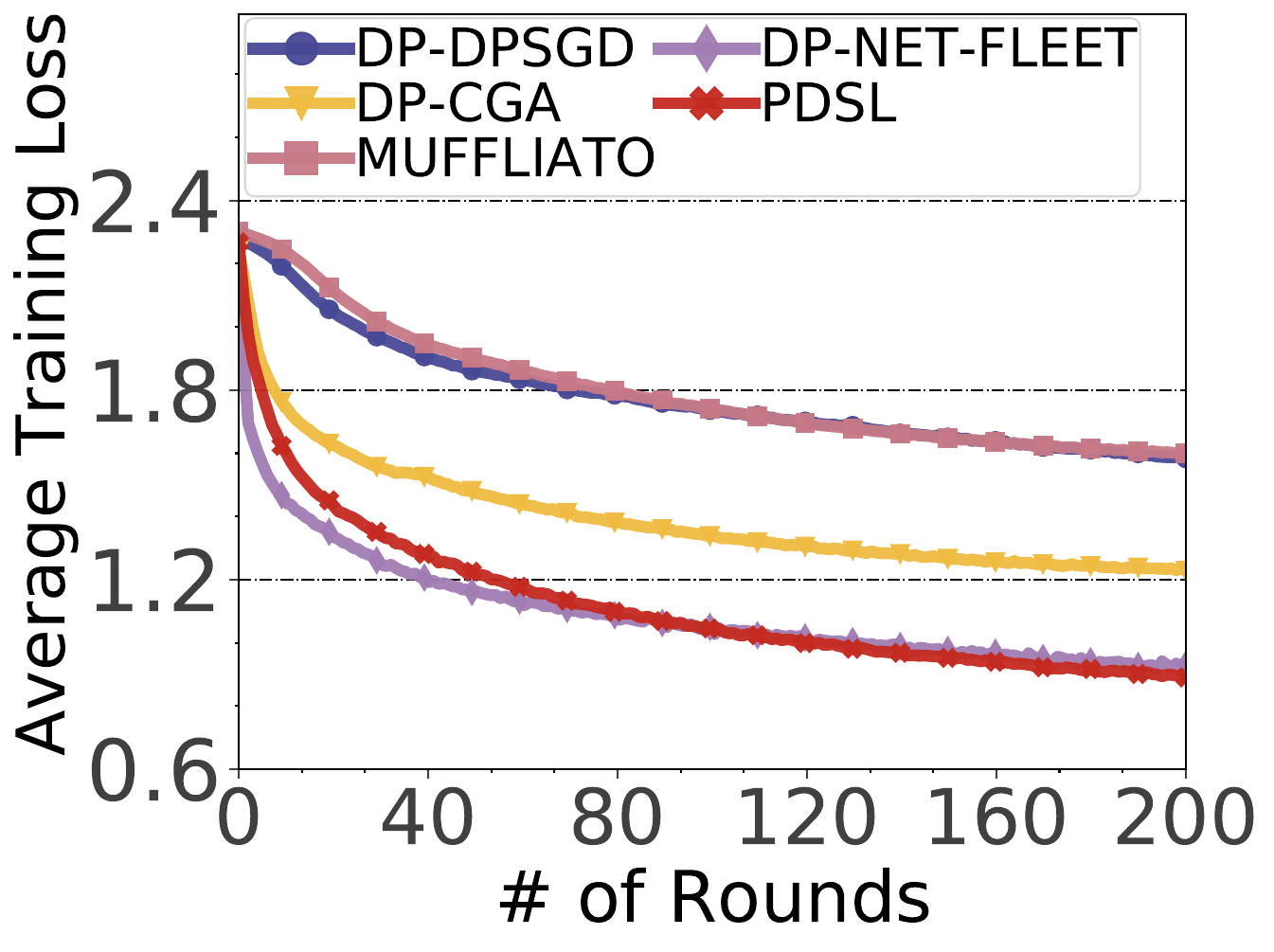}}
            \parbox{.3\textwidth}{\center\includegraphics[width=.25\textwidth]{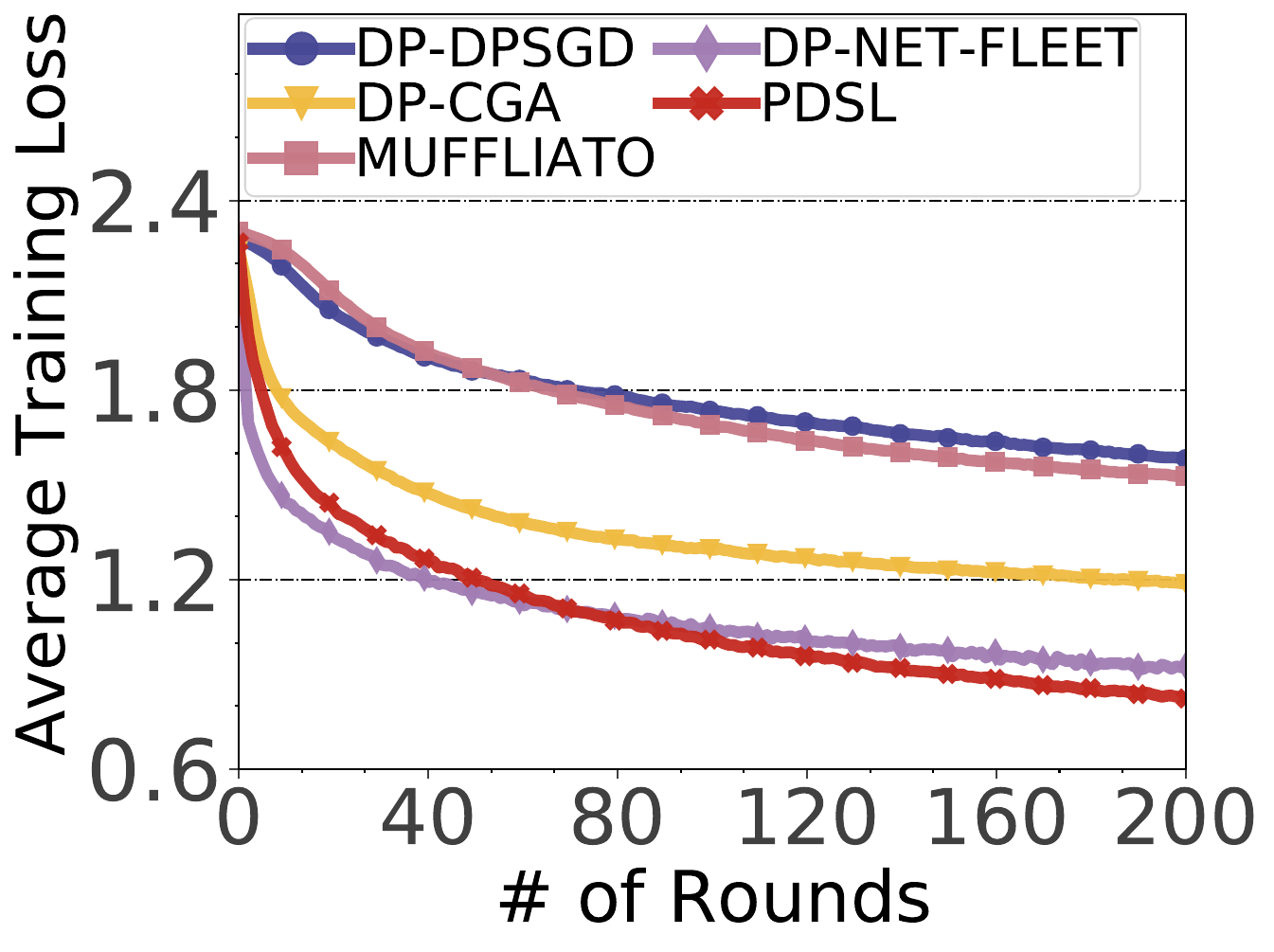}}
            \parbox{.3\textwidth}{\center\scriptsize(b1) $\epsilon=0.5$ with $M=15$}
            \parbox{.3\textwidth}{\center\scriptsize(b2) $\epsilon=0.7$ with $M=15$}
            \parbox{.3\textwidth}{\center\scriptsize(b3) $\epsilon=1$ with $M=15$}
            \parbox{.3\textwidth}{\center\includegraphics[width=.25\textwidth]{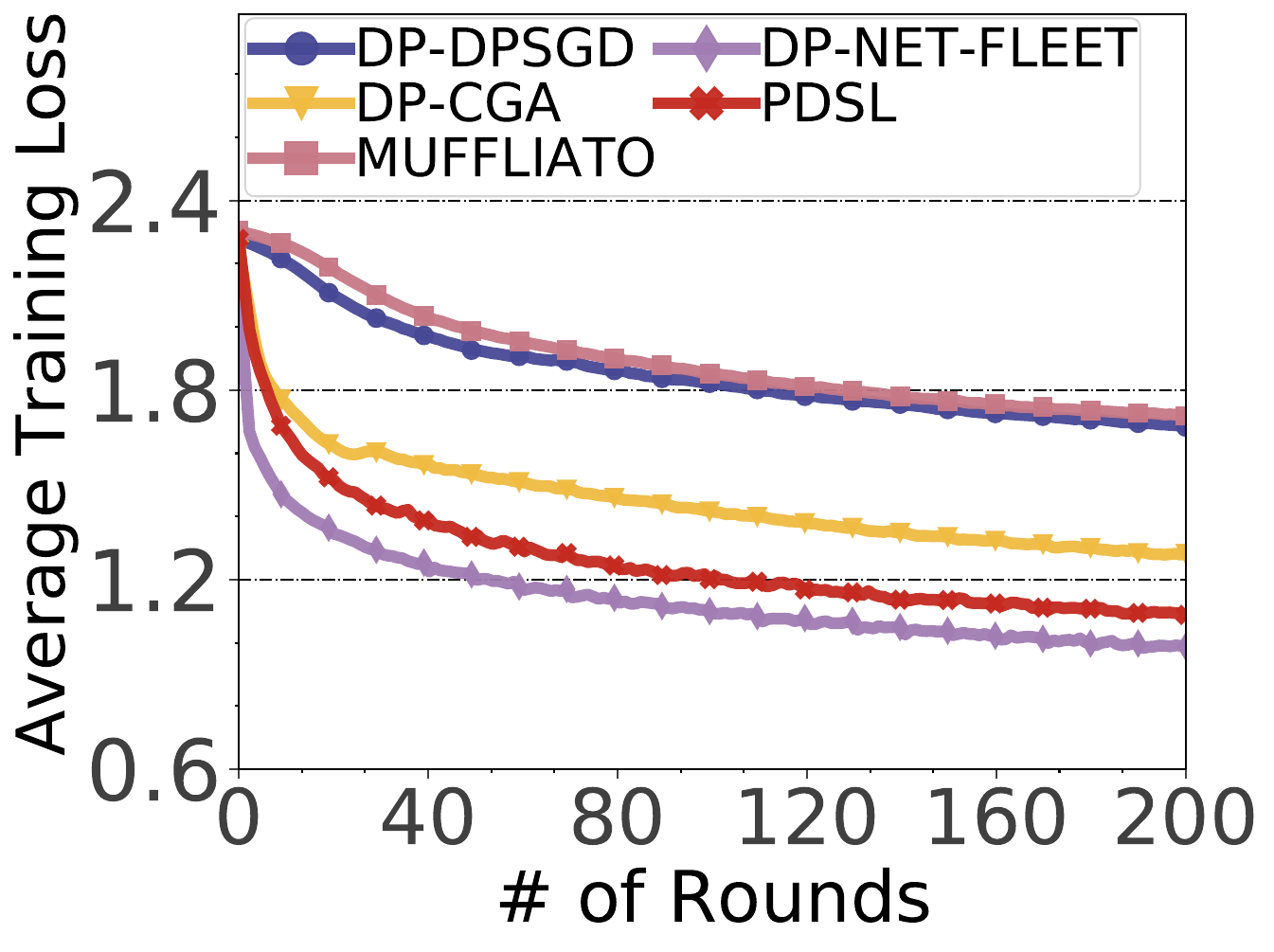}}
            \parbox{.3\textwidth}{\center\includegraphics[width=.25\textwidth]{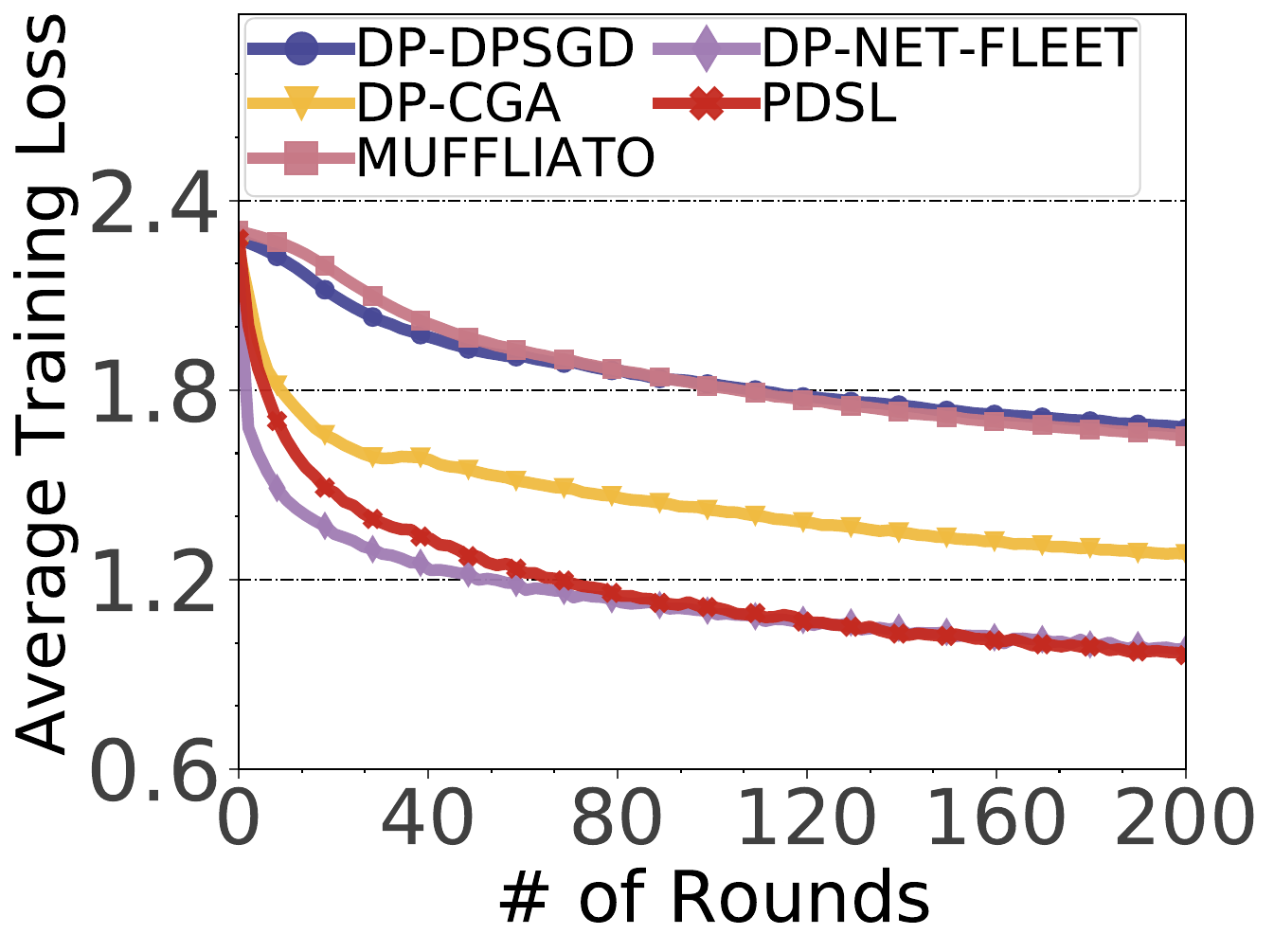}}
            \parbox{.3\textwidth}{\center\includegraphics[width=.25\textwidth]{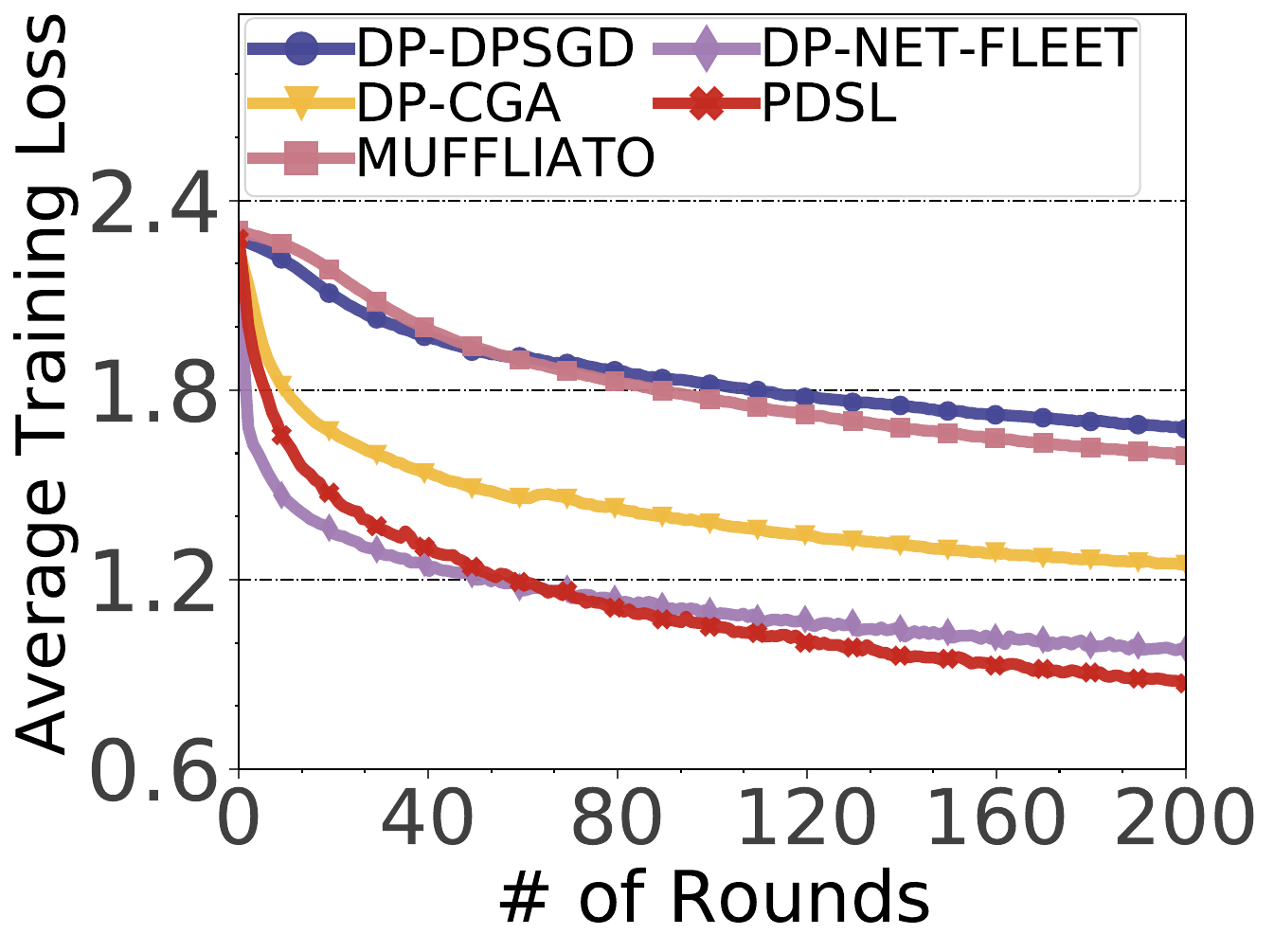}}
            \parbox{.3\textwidth}{\center\scriptsize(c1) $\epsilon=0.5$ with $M=20$}
            \parbox{.3\textwidth}{\center\scriptsize(c2) $\epsilon=0.7$ with $M=20$}
            \parbox{.3\textwidth}{\center\scriptsize(c3) $\epsilon=1$ with $M=20$}
          \caption{Comparison results on CIFAR-10 dataset over bipartite graphs.}
          \label{fig:cifar10-loss-bipartite}
          \end{center}
       \end{figure*}

\begin{figure*}[htb!]
          \begin{center}
            \parbox{.3\textwidth}{\center\includegraphics[width=.25\textwidth]{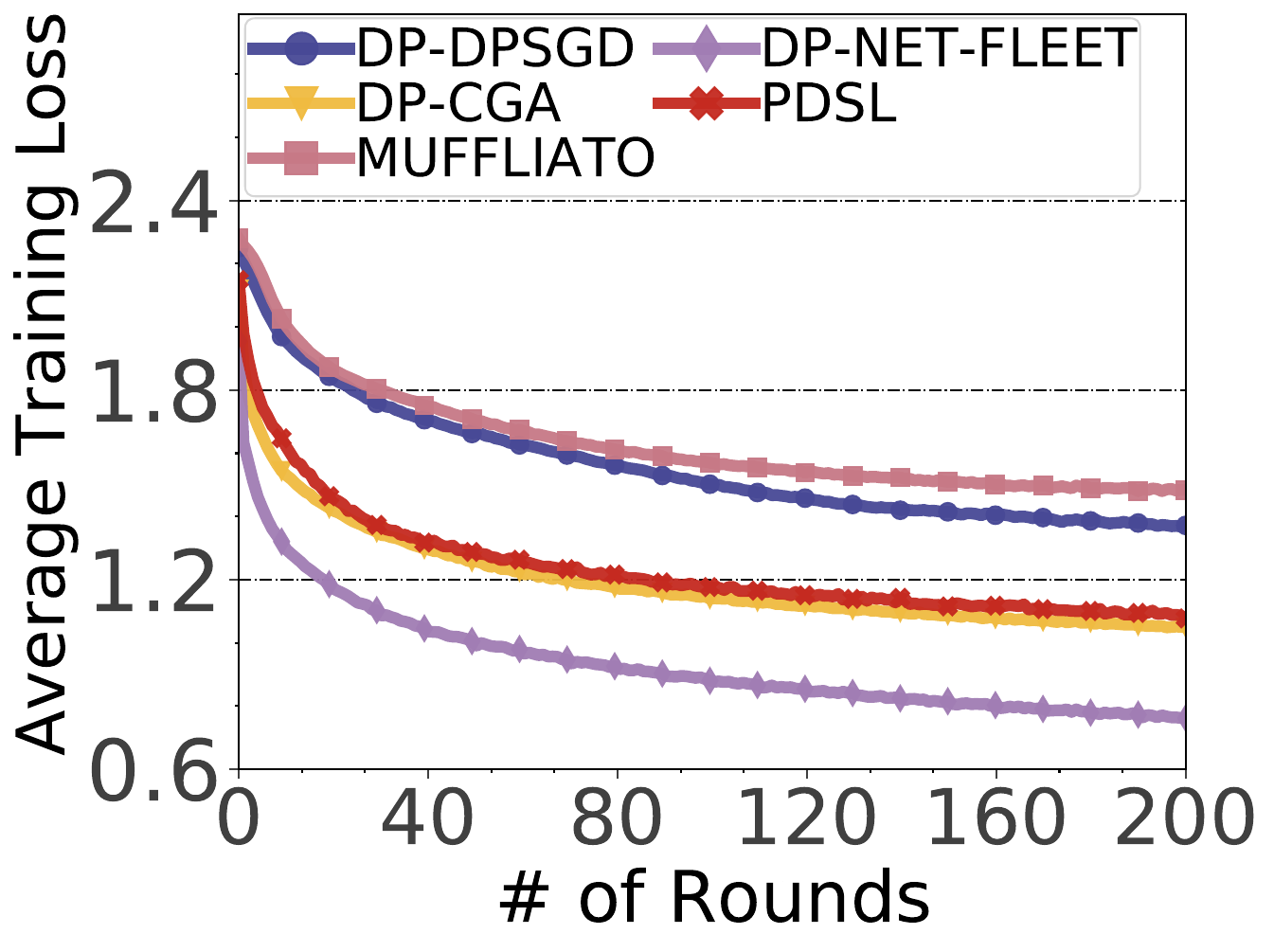}}
            \parbox{.3\textwidth}{\center\includegraphics[width=.25\textwidth]{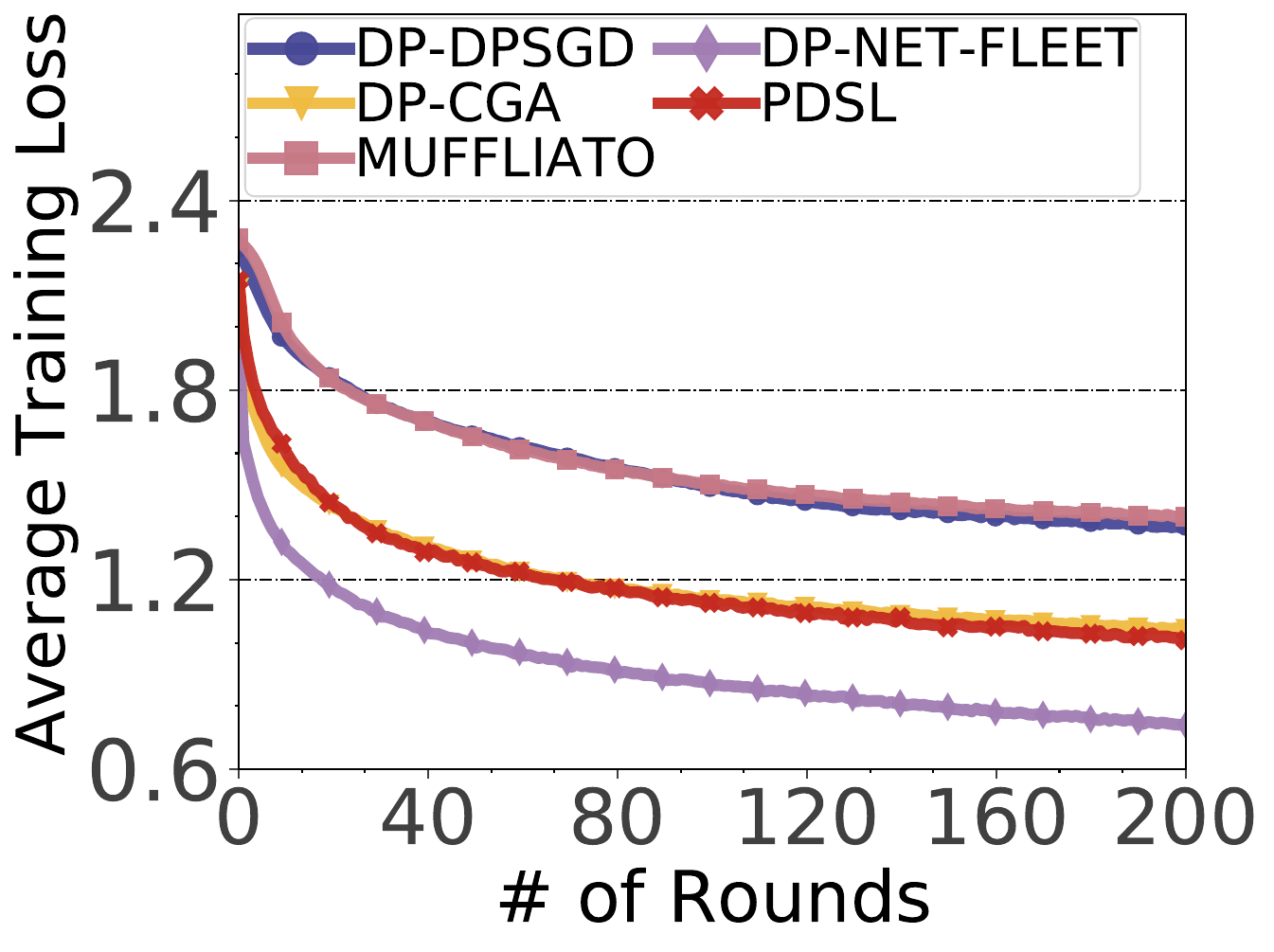}}
            \parbox{.3\textwidth}{\center\includegraphics[width=.25\textwidth]{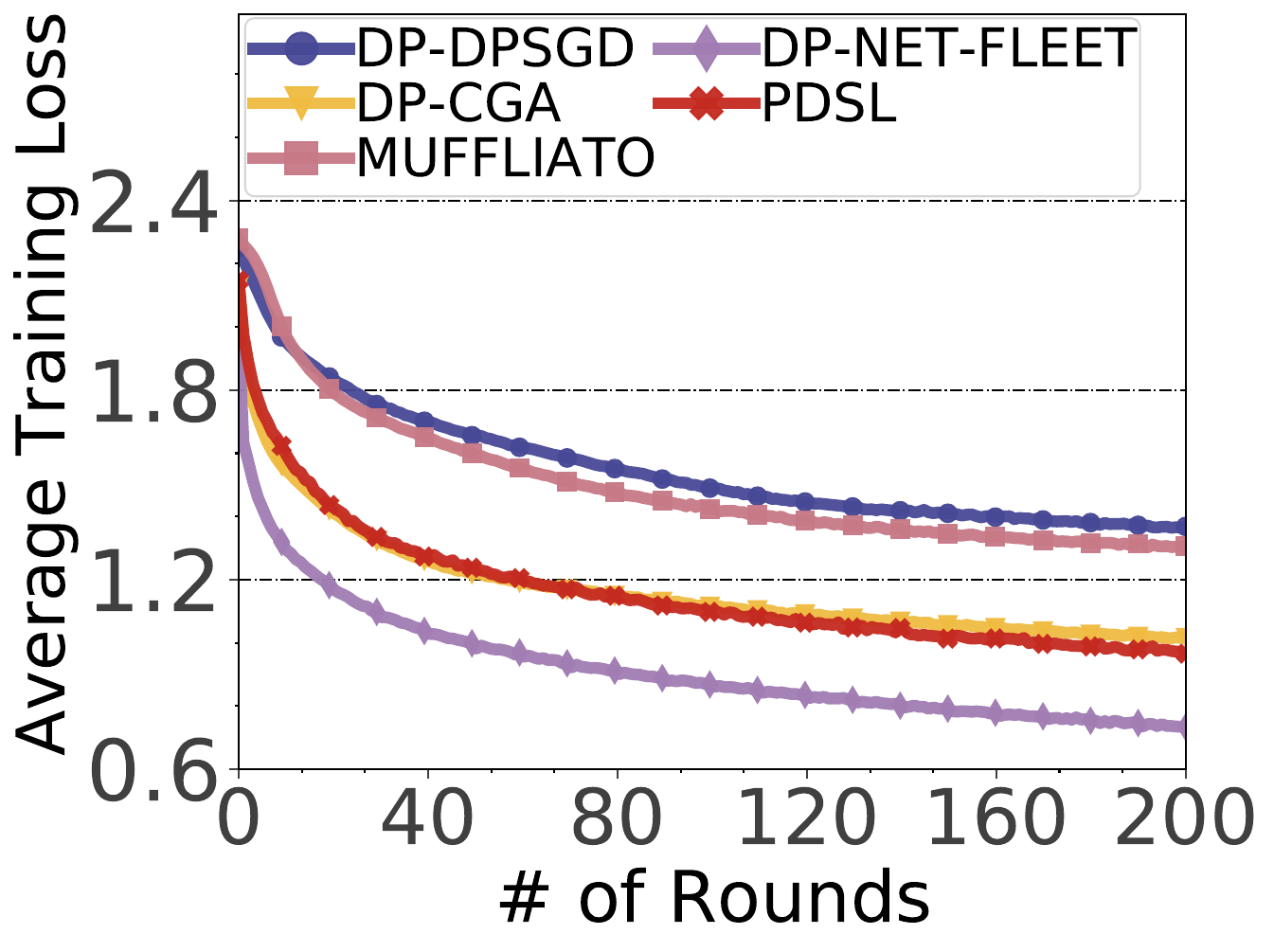}}
            \parbox{.3\textwidth}{\center\scriptsize(a1) $\epsilon=0.5$ with $M=10$}
            \parbox{.3\textwidth}{\center\scriptsize(a2) $\epsilon=0.7$ with $M=10$}
            \parbox{.3\textwidth}{\center\scriptsize(a3) $\epsilon=1$ with $M=10$}
            \parbox{.3\textwidth}{\center\includegraphics[width=.25\textwidth]{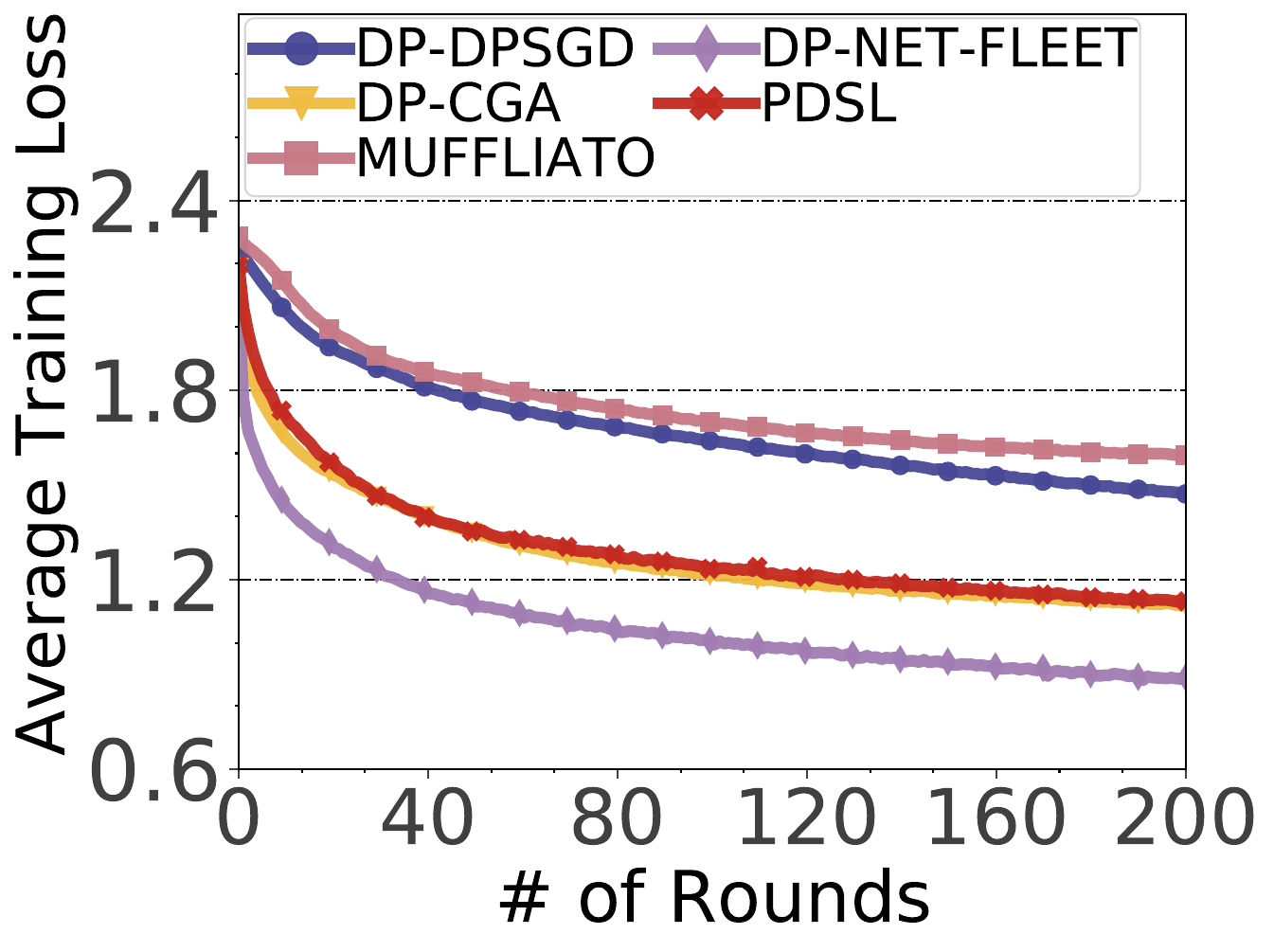}}
            \parbox{.3\textwidth}{\center\includegraphics[width=.25\textwidth]{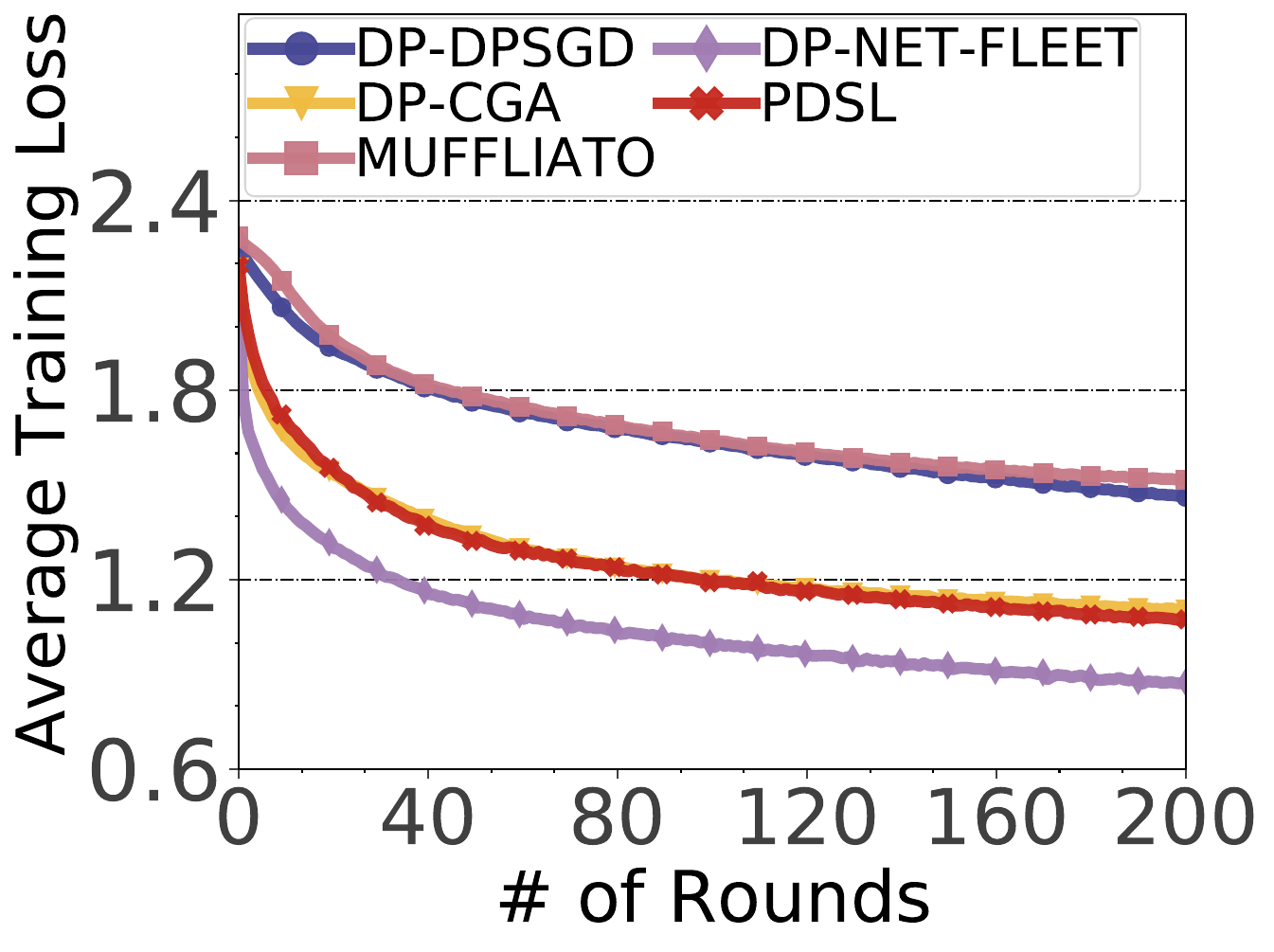}}
            \parbox{.3\textwidth}{\center\includegraphics[width=.25\textwidth]{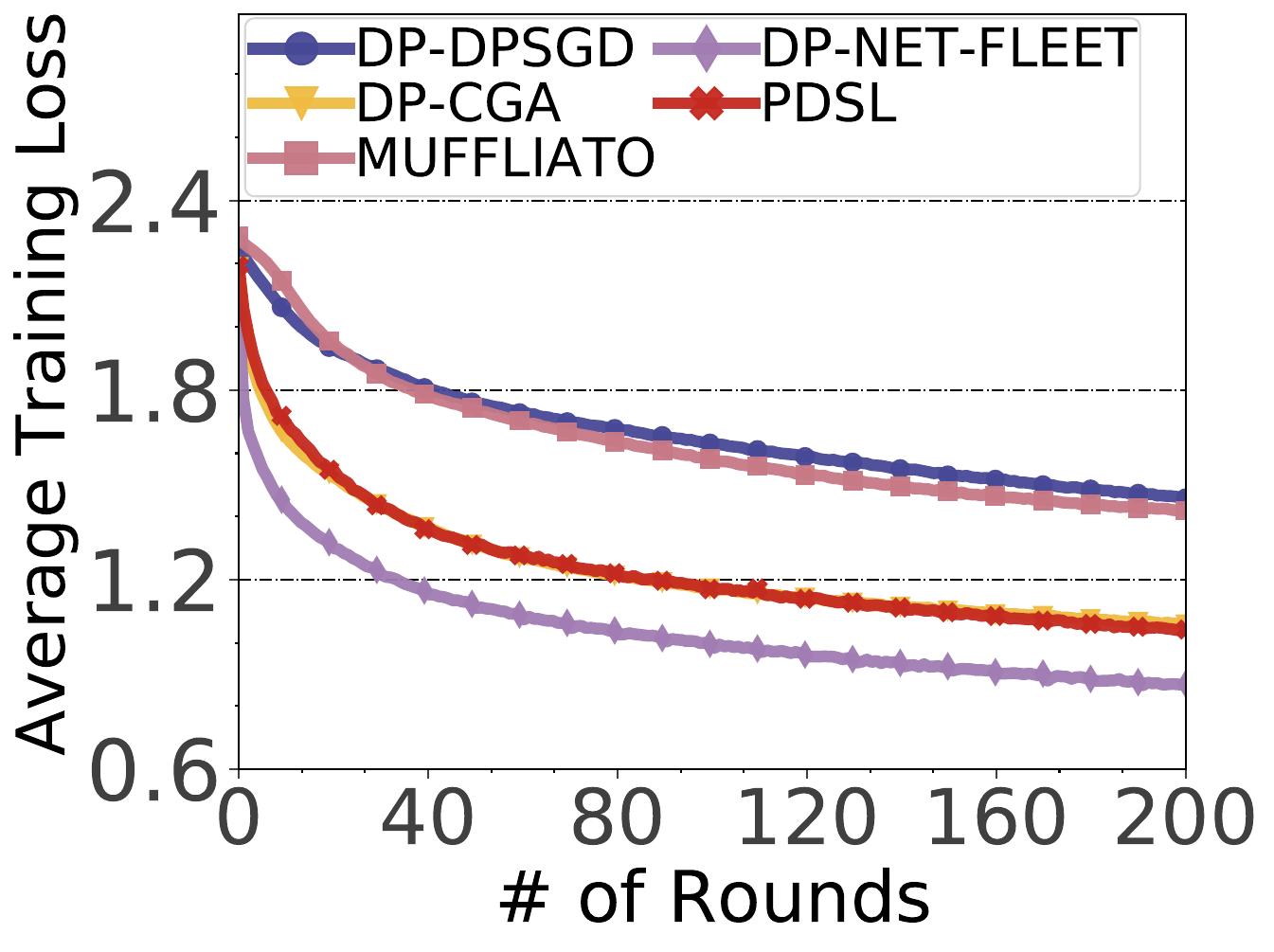}}
            \parbox{.3\textwidth}{\center\scriptsize(b1) $\epsilon=0.5$ with $M=15$}
            \parbox{.3\textwidth}{\center\scriptsize(b2) $\epsilon=0.7$ with $M=15$}
            \parbox{.3\textwidth}{\center\scriptsize(b3) $\epsilon=1$ with $M=15$}
            \parbox{.3\textwidth}{\center\includegraphics[width=.25\textwidth]{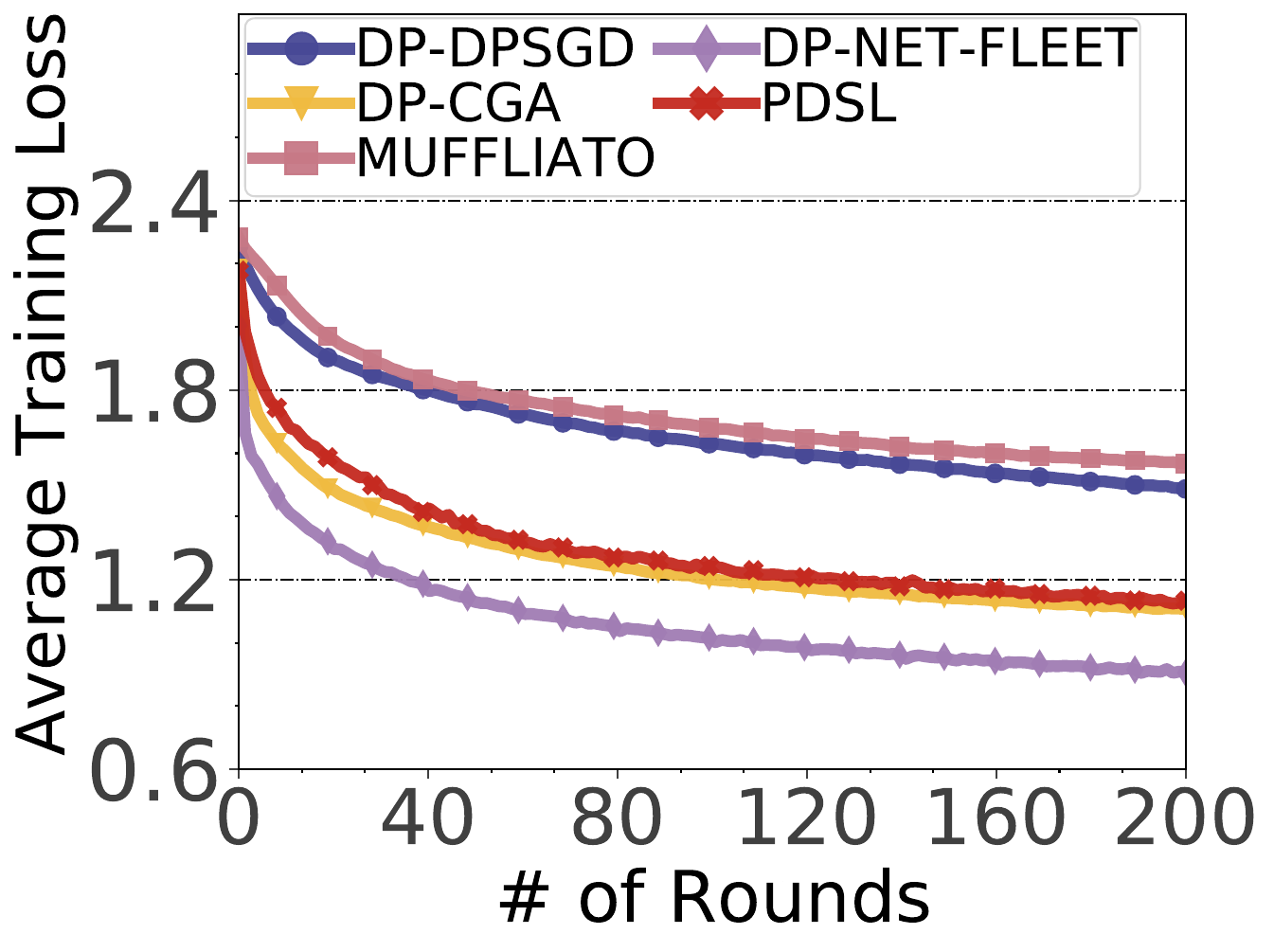}}
            \parbox{.3\textwidth}{\center\includegraphics[width=.25\textwidth]{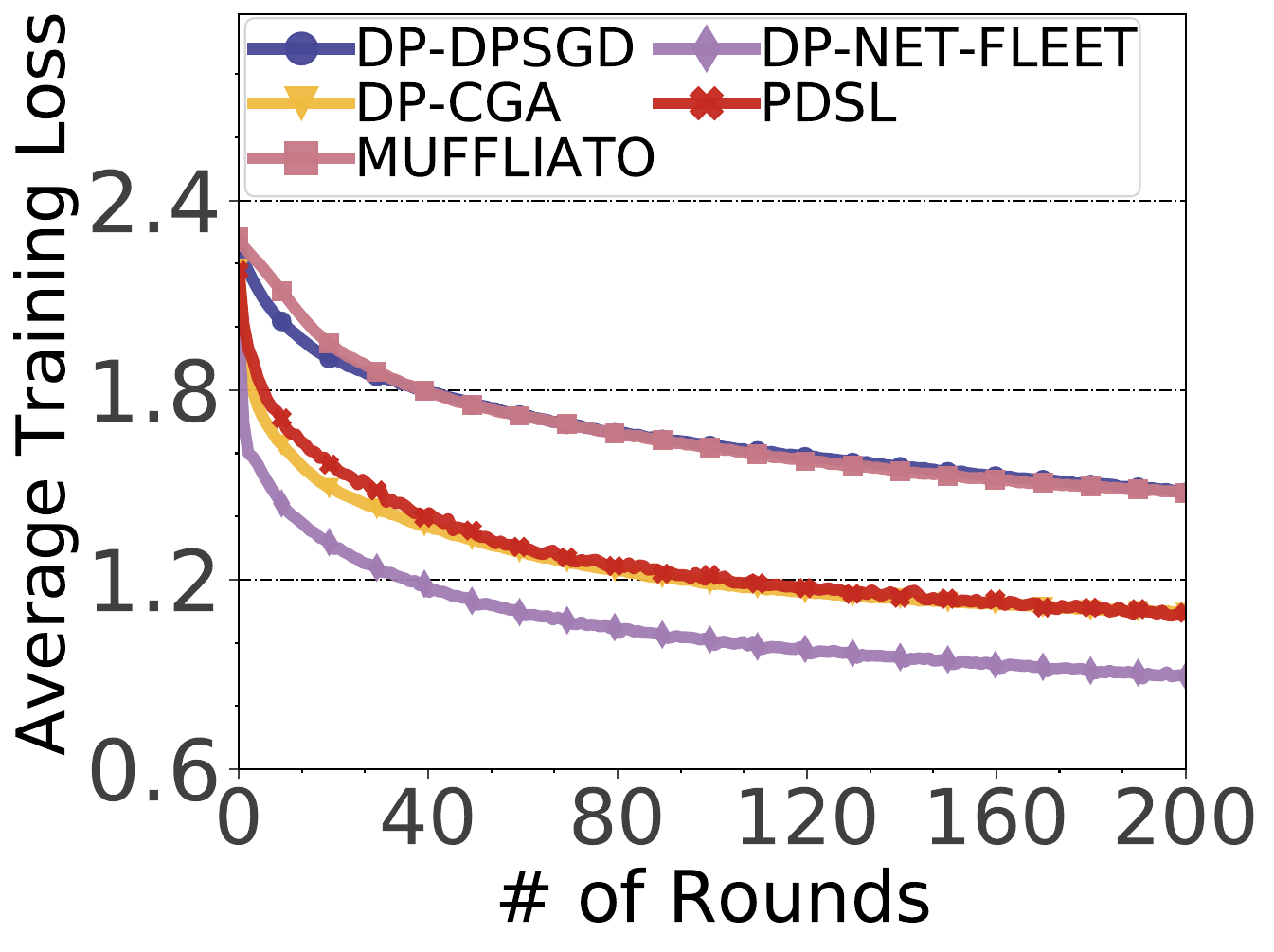}}
            \parbox{.3\textwidth}{\center\includegraphics[width=.25\textwidth]{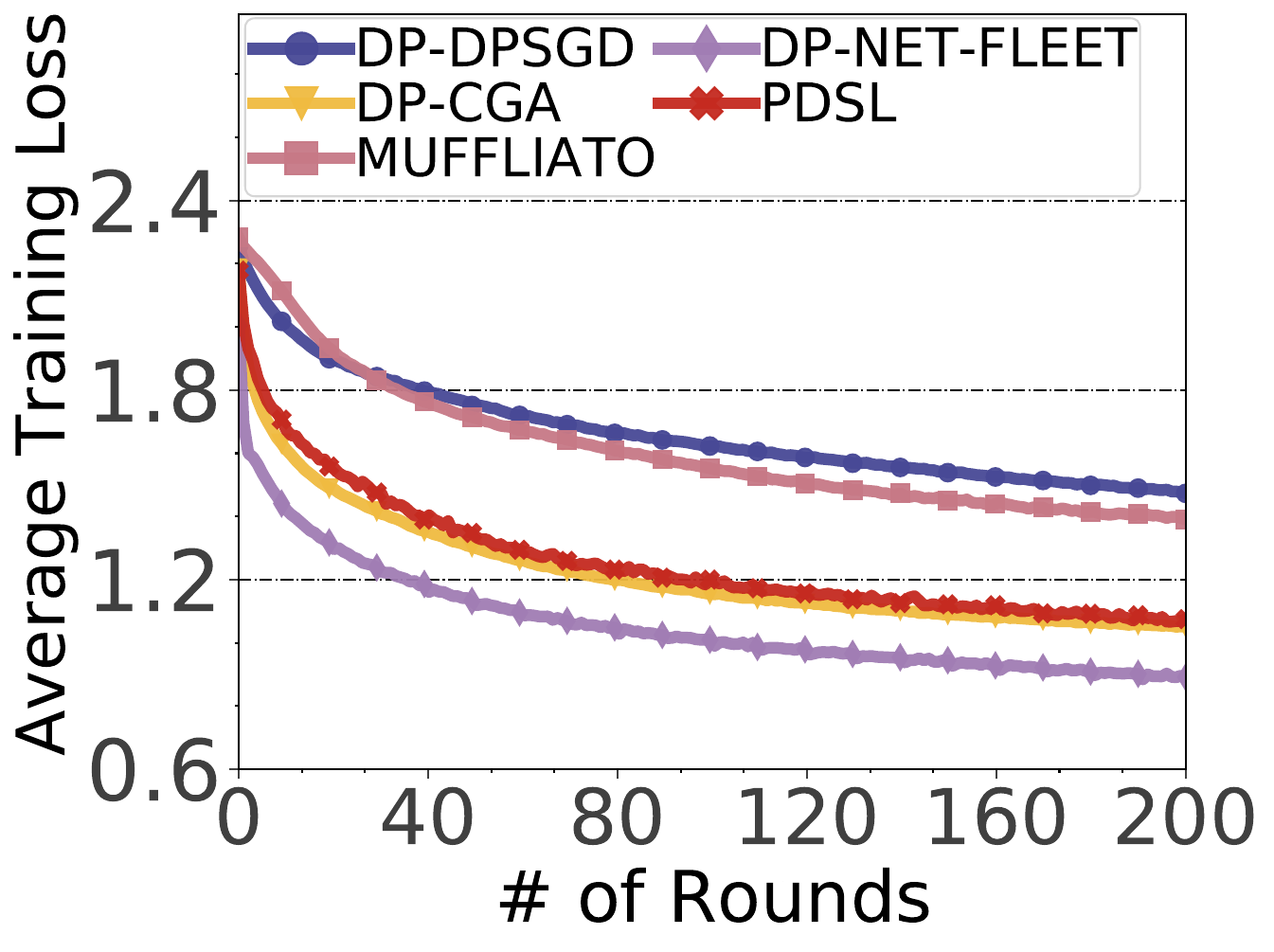}}
            \parbox{.3\textwidth}{\center\scriptsize(c1) $\epsilon=0.5$ with $M=20$}
            \parbox{.3\textwidth}{\center\scriptsize(c2) $\epsilon=0.7$ with $M=20$}
            \parbox{.3\textwidth}{\center\scriptsize(c3) $\epsilon=1$ with $M=20$}
          \caption{Comparison results on CIFAR-10 dataset over ring graphs.}
          \label{fig:cifar10-loss-ring}
          \end{center}
       \end{figure*}

       The results of average losses achieved by our PDSL algorithm, as well as the other four reference algorithms across different privacy budget over the three different communication topologies are presented in Fig.~\ref{fig:cifar10-loss-full}, Fig.~\ref{fig:cifar10-loss-bipartite} and Fig.~\ref{fig:cifar10-loss-ring}, respectively. The results indicate that, our PDSL algorithm significantly outperforms DP-DPSGD, DP-CGA and MUFFLIATO algorithms with different privacy budget on both fully connected graphs and bipartite  graphs. For example, when $\epsilon=1, M=10$, the average loss of PDSL after $200$ communication rounds is around $0.65$ on the fully connected graph; it is similar to the average loss of DP-NET-FLEET, and $2.2$ times smaller than DP-DPSGD, $2.1$ times smaller than MUFFLIATO, and $1.67$ times smaller than DP-CGA. Furthermore, we can see that PDSL still maintains better average loss than other reference algorithms as the number of agents increases. Specifically, when $\epsilon=1$ and $M$ is increased to $20$, the average loss of PDSL is $2.34$ times smaller than DP-DPSGD, $1.74$ times smaller than DP-CGA, $2.2$ times smaller than MUFFLIATO, while $1.23$ times smaller than DP-NET-FLEET. Moreover, compared to PDSL, DP-NET-FLEET exhibits equivalent or even superior convergence properties, as evidenced in Fig.~\ref{fig:cifar10-loss-ring}. However, the test accuracy for DP-NET-FLEET is lower than that achieved by our PDSL algorithm, a detail that will be further elaborated upon subsequently.

       We also compare the test accuracy of the different algorithms with respect to the number of agents and the different values of $\epsilon$ over the three different communication topologies. The results are presented in Table~\ref{tb:cifar10-acc}, from which we can observe that PDSL performs better in test accuracy compared to the other reference algorithms. For instance, when $\epsilon = 1, M = 10$, the test accuracy of our PDSL algorithm on the fully connected graph is $0.758$; it is an improvement of $27.3\%$ over DP-DPSGD, $16.2\%$ over DP-CGA, $23\%$ over MUFFLIATO, and $20.7\%$ over DP-NET-FLEET. Additionally, as the privacy budget $\epsilon$ becomes smaller or the number of agents increases, PDSL still maintains superior performance over the other reference algorithms in terms of test accuracy. Specifically, when $\epsilon = 1$ and $M=20$, our PDSL algorithm has an improvement of around $21\% \sim 30\%$ over the four reference algorithms. When $\epsilon = 0.5, M = 10$, the test accuracy of PDSL reaches $0.691$; that is an improvement of $15\% \sim 31\%$ over the reference algorithms. Moreover, our PDSL algorithm still has better test accuracy with sparser communication graphs. For example, when $\epsilon = 1, M = 20$, the test accuracy of PDSL algorithm is around $0.712$ on bipartite graphs; it is $1.68$ times higher than DP-DPSGD, $1.55$ times higher than DP-CGA, $1.52$ times higher than MUFFLIATO, and $1.71$ times higher than DP-NET-FLEET. If ring graphs are adopted, the test accuracy of PDSL algorithm is $1.49$ times higher than DP-DPSGD, $1.21$ times higher than DP-CGA, $1.23$ times higher than MUFFLIATO, and $1.16$ times higher than DP-NET-FLEET.
       \begin{table*}
          \caption{Test accuracy on CIFAR-10 dataset.}
          \label{tb:cifar10-acc}
          \centering
          \setlength{\tabcolsep}{1.2mm}
          \renewcommand{\arraystretch}{1.1}
          \begin{tabular}{c|c|c|c|c|c|c|c|c|c|c} 
            \hline
            \multirow{2}{*}{Privacy budget $\epsilon$}  & \multirow{2}{*}{Methods} 
            & \multicolumn{3}{c|}{Fully-connected} & \multicolumn{3}{c|}{Bipartite} & \multicolumn{3}{c}{Ring}  \\
            \cline{3-11}
            &  & $M=10$    & $M=15$     & $M=20$     
               & $M=10$   & $M=15$    & $M=20$   
               & $M=10$  & $M=15$  & $M=20$   \\
            \hline
            \multirow{5}{*}{$\epsilon=0.5$}  
            & DP-DPSGD   & 0.453  & 0.451  & 0.423  & 0.441  & 0.443  & 0.412  & 0.374 & 0.388  & 0.326   \\
            & DP-CGA  & 0.375  & 0.369  & 0.325  & 0.395  & 0.415  & 0.383  & 0.385  & 0.391  & 0.356     \\
            & MUFFLIATO  & 0.484   & 0.459  & 0.454  & 0.481  & 0.462  & 0.445  & 0.436  & 0.449   & 0.417   \\
            & DP-NET-FLEET  & 0.541  & 0.524  & 0.502  & 0.455  & 0.458  & 0.408  & 0.483  & 0.471  & 0.439  \\
            & \textbf{PDSL}  & \textbf{0.691}   
                    & \textbf{0.688}  
                    & \textbf{0.686}  
                    & \textbf{0.668}  
                    & \textbf{0.673}  
                    & \textbf{0.654}  
                    & \textbf{0.554}  
                    & \textbf{0.559}  
                    & \textbf{0.517}     \\
            \hline
            \multirow{5}{*}{$\epsilon=0.7$} 
             & DP-DPSGD  & 0.464  & 0.459   & 0.433  & 0.453  & 0.449  & 0.413  & 0.385  & 0.385  & 0.335   \\
            & DP-CGA  & 0.427  & 0.412  & 0.398  & 0.421  & 0.434  & 0.406  & 0.408  & 0.424  & 0.386   \\
            & MUFFLIATO   & 0.505   & 0.481  & 0.466  & 0.503  & 0.476  & 0.457  & 0.447  & 0.458   & 0.422   \\
            & DP-NET-FLEET  & 0.548  & 0.531  & 0.508  & 0.465  & 0.462  & 0.411  & 0.496  & 0.477  & 0.445   \\
            & \textbf{PDSL}  & \textbf{0.744}  
                    & \textbf{0.738}   
                    & \textbf{0.734}  
                    & \textbf{0.718}  
                    & \textbf{0.715}  
                    & \textbf{0.691}  
                    & \textbf{0.569}    
                    & \textbf{0.573}  
                    & \textbf{0.524}   \\
            \hline
            \multirow{5}{*}{$\epsilon=1$} 
             & DP-DPSGD  & 0.485  & 0.463  & 0.447  & 0.461  & 0.455  & 0.422  & 0.403  & 0.383   & 0.353\\
            & DP-CGA  & 0.596  & 0.602  & 0.537  & 0.475  & 0.484  & 0.459  & 0.455  & 0.467  & 0.433  \\
            & MUFFLIATO  & 0.528  & 0.498  & 0.476  & 0.525  & 0.492   & 0.467  & 0.457  & 0.463  & 0.427  \\
            & DP-NET-FLEET  & 0.551  & 0.539  & 0.512  & 0.473  & 0.466  & 0.416  & 0.501  & 0.483  & 0.452   \\
            & \textbf{PDSL}  & \textbf{0.758} 
                    & \textbf{0.753}   
                    & \textbf{0.746}  
                    & \textbf{0.725}  
                    & \textbf{0.731}  
                    & \textbf{0.712}  
                    & \textbf{0.573}    
                    & \textbf{0.582}  
                    & \textbf{0.528}   \\
            \hline
          \end{tabular}
       \end{table*}

 \section{Conclusion} \label{sec:con}
   In this paper, we propose PDSL, a novel privacy-preserved decentralized stochastic learning algorithm with heterogeneous data distribution. In particular, on one hand, we leverage the notion of DP to perturb the cross-gradient information; one the other hand, we utilize the concept of Shapley value to measure the contributions of the perturbed cross-gradient information at each agent for calculating the weighted aggregation of the perturbed gradients. We conduct solid theoretical analysis to reveal the privacy of each agent can be preserved effectively during the decentralized learning process, and the convergence of our algorithm can be ensured with a linear speedup. The efficacy of our algorithm is also verified by our extensive experiments on two real-world datasets.

\bibliographystyle{IEEEtran}
\bibliography{IEEEabrv,ref}

\appendices

\section{Proof of \textbf{Theorem}~\ref{thm:privacy}}
 In \textbf{Algorithm}~1, each agent aggregates the perturbed stochastic gradients from its neighbors (see Lines~13$\sim$18). In any round $t$, the gradient obtained by agent $i$ is
 \begin{align} \label{eq:finalgrad}
    \bar{g}^{[t]}_i &= \sum_{j \in \mathcal{M}_i} \frac{\hat{\varphi}^{[t]}_{i,j}}{\omega_{i,j} \sum_{k \in \mathcal{M}_i} \hat{\varphi}^{[t]}_{i,k}} \left( \frac{g^{[t]}_{j,i}}{\max \left\{1, C^{-1} \left\| g^{[t]}_{j,i} \right\| \right\}} + n^{[t]}_{j,i} \right)  \nonumber\\
    &= \sum_{j \in \mathcal{M}_i} \frac{\hat{\varphi}^{[t]}_{i,j}}{\omega_{i,j} \sum_{k \in \mathcal{M}_i} \hat{\varphi}^{[t]}_{i,k}} \cdot \frac{g^{[t]}_{j,i}}{\max \left\{1, C^{-1} \left\| g^{[t]}_{j,i} \right\| \right\}} + \sum_{j \in \mathcal{M}_i} \frac{\hat{\varphi}^{[t]}_{i,j}}{\omega_{i,j} \sum_{k \in \mathcal{M}_i} \hat{\varphi}^{[t]}_{i,k}} n^{[t]}_{j,i}  \nonumber\\
    & \geq \sum_{j \in \mathcal{M}_i} \frac{\hat{\varphi}^{[t]}_{i,j}}{\omega_{i,j} \sum_{k \in \mathcal{M}_i} \hat{\varphi}^{[t]}_{i,k}} \cdot \frac{g^{[t]}_{j,i}}{\max \left\{1, C^{-1} \left\| g^{[t]}_{j,i} \right\| \right\}} + \sum_{j \in \mathcal{M}_i} \frac{\hat{\varphi}_{\min}}{\omega_{i,j}} n^{[t]}_{j,i}   
 \end{align}
 where $\hat{\varphi}_{\min} = \min_{j \in \mathcal{M}_i, t \in T} \frac{\hat{\varphi}^{[t]}_{i,j}}{\sum_{k \in \mathcal{M}_i} \hat{\varphi}^{[t]}_{i,k}}$.
 For the second term in $(\ref{eq:finalgrad})$, we have
 \begin{align} \label{eq:noise-expectation}
    \mathbb{E} \left[ \sum_{j \in \mathcal{M}_i} \frac{\hat{\varphi}_{\min}}{\omega_{i,j}} n^{[t]}_{j,i} \right] = \sum_{j \in \mathcal{M}_i} \frac{\hat{\varphi}_{\min}}{\omega_{i,j}} \mathbb{E} \left[ n^{[t]}_{j,i} \right] = 0   
 \end{align}
 and
 \begin{align}  \label{eq:noise-variance}
    \text{Var}\left( \sum_{j \in \mathcal{M}_i} \frac{\hat{\varphi}_{\min}}{\omega_{i,j}} n^{[t]}_{j,i} \right) = \left(\hat{\varphi}_{\min} \right)^2 \sum_{j \in \mathcal{M}_i} \frac{1}{\omega^2_{i,j}} \text{Var} \left( n^{[t]}_{j,i} \right) = \sum_{j \in \mathcal{M}_i} \frac{\sigma^2 d \left(\hat{\varphi}_{\min} \right)^2 }{\omega^2_{i,j}}  
 \end{align}
 Therefore, $\sum_{j \in \mathcal{M}_i} \frac{\hat{\varphi}_{\min}}{\omega_{i,j}} n^{[t]}_{j,i} \sim \mathcal{N}\left(0, \sum_{j \in \mathcal{M}_i} \frac{ \left(\hat{\varphi}_{\min} \right)^2 }{\omega^2_{i,j}} \sigma^2 \mathbf{I}_d \right)$.

 We then bound the $L2$-sensitivity of $q^{[t]} = \sum_{j \in \mathcal{M}_i} \frac{\hat{\varphi}^{[t]}_{i,j}}{\omega_{i,j} \sum_{k \in \mathcal{M}_i} \hat{\varphi}^{[t]}_{i,k}} \cdot \frac{g^{[t]}_{j,i}}{\max \left\{1, C^{-1} \left\| g^{[t]}_{j,i} \right\| \right\}}$. For agent $i$, consider any two adjacent datasets $\mathcal{D}_i$ and $\mathcal{D}'_i$. We let the datasets of its neighbors be fixed and thus have
 \begin{align} \label{eq:up-l2-sensitivity-00}
    \triangle_{2} q^{[t]} =& \max_{\mathcal{D}_i, \mathcal{D}'_i} \left\| q^{[t]}(\mathcal{D}_i) - q^{[t]}(\mathcal{D}'_i) \right\|  \nonumber\\
    =& \max_{\mathcal{D}_i, \mathcal{D}'_i} \left\| \begin{aligned}
    & \frac{\hat{\varphi}^{[t]}_{i,i}}{\omega_{i,i} \sum_{k \in \mathcal{M}_i} \hat{\varphi}^{[t]}_{i,k}} \cdot \frac{g^{[t]}_{i,i} (\mathcal{D}_i)}{\max \left\{1, C^{-1} \left\| g^{[t]}_{i,i} (\mathcal{D}_i) \right\| \right\}}  \\
    & + \sum_{j \in \mathcal{M}_i, j\neq i} \frac{\hat{\varphi}^{[t]}_{i,j}}{\omega_{i,j} \sum_{k \in \mathcal{M}_i} \hat{\varphi}^{[t]}_{i,k}} \cdot \frac{g^{[t]}_{j,i} (\mathcal{D}_j)}{\max \left\{1, C^{-1} \left\| g^{[t]}_{j,i} (\mathcal{D}_j) \right\| \right\}} \\
    & - \frac{(\hat{\varphi}^{[t]}_{i,i})^-}{\omega_{i,i} \sum_{k \in \mathcal{M}_i} (\hat{\varphi}^{[t]}_{i,k})^-} \cdot \frac{g^{[t]}_{i,i} (\mathcal{D}'_i)}{\max \left\{1, C^{-1} \left\| g^{[t]}_{i,i} (\mathcal{D}'_i) \right\| \right\}}  \\
    & - \sum_{j \in \mathcal{M}_i, j \neq i} \frac{(\hat{\varphi}^{[t]}_{i,j})^-}{\omega_{i,j} \sum_{k \in \mathcal{M}_i} (\hat{\varphi}^{[t]}_{i,k})^-} \cdot \frac{g^{[t]}_{j,i} (\mathcal{D}_j)}{\max \left\{1, C^{-1} \left\| g^{[t]}_{j,i} (\mathcal{D}_j) \right\| \right\}} \end{aligned}  \right\|  \nonumber\\
    =& \max_{\mathcal{D}_i, \mathcal{D}'_i} \left\| \begin{aligned} &  \frac{\hat{\varphi}^{[t]}_{i,i}}{\omega_{i,i} \sum_{k \in \mathcal{M}_i} \hat{\varphi}^{[t]}_{i,k}} \left( \frac{g^{[t]}_{i,i} (\mathcal{D}_i)}{\max \left\{1, C^{-1} \left\| g^{[t]}_{i,i} (\mathcal{D}_i) \right\| \right\}} - \frac{g^{[t]}_{i,i} (\mathcal{D}'_i)}{\max \left\{1, C^{-1} \left\| g^{[t]}_{i,i} (\mathcal{D}'_i) \right\| \right\}} \right)  \\ 
    & + \left( \frac{\hat{\varphi}^{[t]}_{i,i}}{\omega_{i,i} \sum_{k \in \mathcal{M}_i} \hat{\varphi}^{[t]}_{i,k}} - \frac{(\hat{\varphi}^{[t]}_{i,i})^-}{\omega_{i,i} \sum_{k \in \mathcal{M}_i} (\hat{\varphi}^{[t]}_{i,k})^-} \right) \frac{g^{[t]}_{i,i} (\mathcal{D}'_i)}{\max \left\{1, C^{-1} \left\| g^{[t]}_{i,i} (\mathcal{D}'_i) \right\| \right\}}  \\
    & + \sum_{j \in \mathcal{M}_i, j\neq i} \left(\frac{\hat{\varphi}^{[t]}_{i,j}}{\omega_{i,j} \sum_{k \in \mathcal{M}_i}  \hat{\varphi}^{[t]}_{i,k}} - \frac{(\hat{\varphi}^{[t]}_{i,j})^-}{\omega_{i,j} \sum_{k \in \mathcal{M}_i} (\hat{\varphi}^{[t]}_{i,k})^-} \right) \frac{g^{[t]}_{j,i} (\mathcal{D}_j)}{\max \left\{1, C^{-1} \left\| g^{[t]}_{j,i} (\mathcal{D}_j) \right\| \right\}} \end{aligned} \right\|  \nonumber\\
    \leq & \max_{\mathcal{D}_i, \mathcal{D}'_i} \left\|   \frac{\hat{\varphi}^{[t]}_{i,i}}{\omega_{i,i} \sum_{k \in \mathcal{M}_i} \hat{\varphi}^{[t]}_{i,k}} \left( \frac{g^{[t]}_{i,i} (\mathcal{D}_i)}{\max \left\{1, C^{-1} \left\| g^{[t]}_{i,i} (\mathcal{D}_i) \right\| \right\}} - \frac{g^{[t]}_{i,i} (\mathcal{D}'_i)}{\max \left\{1, C^{-1} \left\| g^{[t]}_{i,i} (\mathcal{D}'_i) \right\| \right\}} \right) \right\|  \nonumber\\
    & + \max_{\mathcal{D}_i, \mathcal{D}'_i} \left\|  \left( \frac{\hat{\varphi}^{[t]}_{i,i}}{\omega_{i,i} \sum_{k \in \mathcal{M}_i} \hat{\varphi}^{[t]}_{i,k}} - \frac{(\hat{\varphi}^{[t]}_{i,i})^-}{\omega_{i,i} \sum_{k \in \mathcal{M}_i} (\hat{\varphi}^{[t]}_{i,k})^-} \right) \frac{g^{[t]}_{i,i} (\mathcal{D}'_i)}{\max \left\{1, C^{-1} \left\| g^{[t]}_{i,i} (\mathcal{D}'_i) \right\| \right\}} \right\|  \nonumber\\
    & + \max_{\mathcal{D}_i, \mathcal{D}'_i} \left\|  \sum_{j \in \mathcal{M}_i, j\neq i} \left(\frac{\hat{\varphi}^{[t]}_{i,j}}{\omega_{i,j} \sum_{k \in \mathcal{M}_i}  \hat{\varphi}^{[t]}_{i,k}} - \frac{(\hat{\varphi}^{[t]}_{i,j})^-}{\omega_{i,j} \sum_{k \in \mathcal{M}_i} (\hat{\varphi}^{[t]}_{i,k})^-} \right) \frac{g^{[t]}_{j,i} (\mathcal{D}_j)}{\max \left\{1, C^{-1} \left\| g^{[t]}_{j,i} (\mathcal{D}_j) \right\| \right\}} \right\|  \nonumber\\
    \leq & \frac{2C}{ \omega_{i,i}} + \sum_{j \in \mathcal{M}_i} \frac{2C}{\omega_{i,j}}  
 \end{align}
 where $g^{[t]}_{i,i}\left(\mathcal{D}_i\right)$ and $g^{[t]}_{i,i}\left(\mathcal{D}'_i\right)$ represent the gradients corresponding to the adjacent datasets $\mathcal{D}_i$ and $\mathcal{D}'_i$, respectively. $g^{[t]}_{j,i}\left(\mathcal{D}_j\right)$ is the gradient of agent $j$'s local dataset $\mathcal{D}_j$ with respect to $i$'s local model.
 Similarly, for any other agent $j \in \mathcal{M}_i$ such that $j \neq i$, consider any two adjacent datasets $\mathcal{D}_j$ and $\mathcal{D}'_j$, when fixing the datasets of $\forall k \in \mathcal{M}_i \setminus j$, we also have
 \begin{align} \label{eq:up-l2-sensitivity-01}
    \triangle_{2} q^{[t]} = \max_{\mathcal{D}_j, \mathcal{D}'_j} \left\| q^{[t]}(\mathcal{D}_j) - q^{[t]}(\mathcal{D}'_j) \right\| \leq  \frac{2C}{ \omega_{i,j}} + \sum_{j \in \mathcal{M}_i} \frac{2C}{\omega_{i,j}}  
 \end{align}
 therefore, based on (\ref{eq:up-l2-sensitivity-00}) and (\ref{eq:up-l2-sensitivity-01}), we have
 \begin{align} \label{eq:up-l2-sensitivity-02}
    \triangle_{2} q^{[t]} \leq  \frac{2C}{ \omega_{\min}} + \sum_{j \in \mathcal{M}_i} \frac{2C}{\omega_{i,j}} 
 \end{align}
 where $\omega_{\min} = \min_{i \in \mathcal{M}, j \in \mathcal{M}_i} \omega_{i,j}$.
 Based on \textbf{Theorem}~3.22 in [32], the Gaussian mechanism (i.e., (4)) meets $\sigma \geq \frac{ \sqrt{2 \ln(1.25/\delta)} \triangle_{2} q }{\epsilon}$. Therefore, for agent $i$, we have
 \begin{align*}
    \sigma \geq \frac{ 2C \left( \frac{1}{ \omega_{\min}} + \sum_{j \in \mathcal{M}_i} \frac{1}{\omega_{i,j}} \right) \sqrt{2 \ln(1.25/\delta)} }{\hat{\varphi}_{\min} \epsilon  \sqrt{\sum_{j \in \mathcal{M}_i} \omega^{-2}_{i,j}}} 
 \end{align*}
 based on which, we have
 \begin{align*}
    \sigma \geq \max_{i \in \mathcal{M}} \left\{ \frac{ 2C \left( \frac{1}{ \omega_{\min}} + \sum_{j \in \mathcal{M}_i} \frac{1}{\omega_{i,j}} \right) \sqrt{2 \ln(1.25/\delta)} }{\hat{\varphi}_{\min} \epsilon  \sqrt{\sum_{j \in \mathcal{M}_i} \omega^{-2}_{i,j}}} \right\}
 \end{align*}

\section{Proof of \textbf{Theorem}~\ref{thm:convergence-main} }
 We show the detailed proof of \textbf{Theorem}~\ref{thm:convergence-main}. Recall that $\bar{g}^{[t]}_i$ is the average of the perturbed gradients $\Big\{ \hat{g}^{[t]}_{j,i} \Big\}_{j\in\mathcal{M}_i}$ weighted by the Shapley values (see (\ref{eq:avggrad})),  \textbf{Lemma}~\ref{lem:bdavggrad} bounds the (expected) weighted average perturbed gradient.

 \begin{lemma} \label{lem:bdavggrad}
    When \textbf{Assumptions}~\ref{ass:L-smooth}$\sim$\ref{ass:ds_mat} hold, for any $t \geq 1$, we have
    \begin{align} \label{eq:avggrad01}
        \mathbb{E} \left[ \left\| \frac{1}{M}\sum^{M}_{i=1} \bar{g}^{[t]}_i \right\|^2 \right] \leq \frac{2C^2}{\omega^4_{\min}} + \frac{2\sigma^2 d}{\omega^4_{\min}} 
    \end{align}
 \end{lemma} 
 \begin{proof}
    \begin{align*}
        \mathbb{E} \left[ \left\| \frac{1}{M}\sum^{M}_{i=1} \bar{g}^{[t]}_i \right\|^2 \right]  
        =& \mathbb{E} \left[ \left\| \frac{1}{M} \sum^M_{i=1} \sum_{j \in \mathcal{M}_i} \pi^{[t]}_{i,j} \left( \frac{g^{[t]}_{j,i}}{\max \left\{1, C^{-1} \left\| g^{[t]}_{j,i} \right\| \right\}} + n^{[t]}_{j,i} \right) \right\|^2 \right]  \\
        \stackrel{\scriptsize{\circled{1}}} \leq & 2 \mathbb{E} \left[ \left\| \frac{1}{M} \sum^M_{i=1} \sum_{j \in \mathcal{M}_i} \frac{\hat{\varphi}^{[t]}_{i,j}}{\omega_{i,j} \sum_{k\in\mathcal{M}_i} \hat{\varphi}^{[t]}_{i,k}} \left( \frac{g^{[t]}_{j,i}}{\max \left\{1, C^{-1} \left\| g^{[t]}_{j,i} \right\| \right\}} \right) \right\|^2 
        \right]   \\
        & + 2 \mathbb{E} \left[ \left\| \frac{1}{M} \sum^M_{i=1}  \sum_{j \in \mathcal{M}_i} \frac{\hat{\varphi}^{[t]}_{i,j}}{\omega_{i,j} \sum_{k\in\mathcal{M}_i} \hat{\varphi}^{[t]}_{i,k}} n^{[t]}_{j,i} \right\|^2 \right] \\
        \stackrel{\scriptsize{\circled{2}}} \leq & \frac{2}{M} \sum^M_{i=1} \sum_{j \in \mathcal{M}_i} \omega_{i,j} \mathbb{E} \left[ \left\| \frac{\hat{\varphi}^{[t]}_{i,j}}{\omega^2_{i,j} \sum_{k\in\mathcal{M}_i} \hat{\varphi}^{[t]}_{i,k}} \right\|^2 
        \left\| \frac{g^{[t]}_{j,i}}{\max \left\{1, C^{-1} \left\| g^{[t]}_{j,i} \right\| \right\}} \right\|^2 
        \right]   \\
        & + \frac{2}{M} \sum^M_{i=1} \sum_{j \in \mathcal{M}_i} \omega_{i,j} \mathbb{E} \left[ \left\| \frac{\hat{\varphi}^{[t]}_{i,j}}{\omega^2_{i,j} \sum_{k\in\mathcal{M}_i} \hat{\varphi}^{[t]}_{i,k}} \right\|^2 \left\| n^{[t]}_{j,i} \right\|^2 \right]    \\
        \stackrel{\scriptsize{\circled{3}}} \leq & \frac{2}{M} \sum^M_{i=1} \sum_{j \in \mathcal{M}_i} \omega_{i,j} \frac{1}{\omega^4_{i,j}} \mathbb{E} \left[ \left\| \frac{g^{[t]}_{j,i}}{\max \left\{1, C^{-1} \left\| g^{[t]}_{j,i} \right\| \right\}} \right\|^2 
        \right]  \\
        & + \frac{2}{M} \sum^M_{i=1} \sum_{j \in \mathcal{M}_i} \omega_{i,j} \frac{1}{\omega^4_{i,j}} \mathbb{E} \left[ \left\| n^{[t]}_{j,i} \right\|^2 \right]  \\
        \stackrel{\scriptsize{\circled{4}}} \leq & \frac{2}{M} \sum^M_{i=1} \sum_{j \in \mathcal{M}_i} \omega_{i,j} \frac{C^2}{\omega^4_{i,j}} + \frac{2}{M} \sum^M_{i=1} \sum_{j \in \mathcal{M}_i} \omega_{i,j} \frac{\sigma^2 d}{\omega^4_{i,j}}  \\
        \stackrel{\scriptsize{\circled{5}}} \leq & \frac{2C^2}{\omega^4_{\min}} + \frac{2\sigma^2 d}{\omega^4_{\min}} 
    \end{align*}
    where $\scriptsize{\circled{1}}$ follows from the inequality $\|a+b\|^2 \leq 2\|a\|^2 + 2\|b\|^2$; $\scriptsize{\circled{2}}$ follows from the convexity of $\|\cdot\|^2$ and \textit{Jensen's Inequality}; $\scriptsize{\circled{3}}$ according to $\frac{\hat{\varphi}^{[t]}_{i,j}} {\sum_{k\in\mathcal{M}_i} \hat{\varphi}^{[t]}_{i,k}} \leq 1$; $\scriptsize{\circled{4}}$ due to the characteristic of clipping process that $\left\| \frac{g^{[t]}_{j,i}}{\max \left\{1, C^{-1} \left\| g^{[t]}_{j,i} \right\| \right\}} \right\|^2 \leq C^2$, and the property of Gaussian noises that $\mathbb{E} \left[ \left\| n^{[t]}_{j,i} \right\|^2 \right] = \sigma^2 d$; and $\scriptsize{\circled{5}}$ follows from $ \omega_{\min} = \min_{i \in \mathcal{M}, j \in \mathcal{M}_i}  \omega_{i,j}$.
 \end{proof}

 Based on \textbf{Lemma}~\ref{lem:bdavggrad}, \textbf{Lemma}~\ref{lem:diffhatgrad} presents that $\mathbb{E} \left[ \left\| \frac{1}{M} \sum^M_{i=1} \left( \bar{g}^{[t]}_i - g^{[t]}_{i,i} \right) \right\|^2 \right]$ is bounded by $\mathbb{E} \left[ \left\| \frac{1}{M} \sum^M_{i=1} \nabla f_i \left( x^{[t-1]}_i \right) \right\|^2 \right]$.

 \begin{lemma} \label{lem:diffhatgrad}
    When \textbf{Assumptions}~\ref{ass:L-smooth}$\sim$\ref{ass:ds_mat} hold, for any $t \geq 1$, we have 
    \begin{align} \label{eq:diffhatgrad}
        & \mathbb{E} \left[ \left\| \frac{1}{M} \sum^M_{i=1} \left( \bar{g}^{[t]}_i - g^{[t]}_{i,i} \right) \right\|^2 \right]  \nonumber\\
        \leq & \frac{4C^2}{\omega^4_{\min}} + \frac{4 \sigma^2 d}{\omega^4_{\min}} + \frac{2 \zeta^2}{M} + 2 \mathbb{E} \left[ \left\| \frac{1}{M} \sum^M_{i=1} \nabla f_i \left( x^{[t-1]}_i \right) \right\|^2
        \right]
    \end{align}
 \end{lemma}
 \begin{proof}
    Recall that $g^{[t]}_{i,i} = \nabla F_i \left( x^{[t-1]}_i; \xi_{i,t} \right)$ is mutually independent unbiased stochastic gradient at $x^{[t-1]}_i$ such that $\mathbb{E} \left[ g^{[t]}_{i,i} \right] = \nabla f_i \left( x^{[t-1]}_i \right)$ for $\forall i \in \mathcal{M}$. We have
    \begin{align*}
        \mathbb{E} \left[ \left\| \frac{1}{M} \sum^M_{i=1} \left( \bar{g}^{[t]}_i - g^{[t]}_{i,i} \right) \right\|^2 \right]  
        \stackrel{\scriptsize{\circled{1}}} \leq & 2 \mathbb{E} \left[ \left\| \frac{1}{M} \sum^M_{i=1} \bar{g}^{[t]}_i \right\|^2 \right] + 2 \mathbb{E} \left[ \left\| \frac{1}{M} \sum^M_{i=1} g^{[t]}_{i,i} \right\|^2 \right]  \\
        \stackrel{\scriptsize{\circled{2}}} \leq &  \frac{4C^2}{\omega^4_{\min}} + \frac{4 \sigma^2 d}{\omega^4_{\min}} 
        + 2 \mathbb{E} \left[ \left\| \frac{1}{M} \sum^M_{i=1} g^{[t]}_{i,i} \right\|^2 \right]  \\
        \stackrel{\scriptsize{\circled{3}}} \leq & \frac{4C^2}{\omega^4_{\min}} + \frac{4 \sigma^2 d}{\omega^4_{\min}} + + 2 \mathbb{E} \left[ \left\| \frac{1}{M} \sum^M_{i=1} \nabla f_i \left( x^{[t-1]}_i \right) \right\|^2 \right]  \\ 
        & + 2 \mathbb{E} \left[ \left\| \frac{1}{M} \sum^M_{i=1} \left( \nabla F_i \left( x^{[t-1]}_i; \xi_{i,t} \right) - \nabla f_i \left( x^{[t-1]}_i \right) \right)  \right\|^2 \right]  \\
        \stackrel{\scriptsize{\circled{4}}} \leq & \frac{4C^2}{\omega^4_{\min}} + \frac{4 \sigma^2 d}{\omega^4_{\min}} + 2 \mathbb{E} \left[ \left\| \frac{1}{M} \sum^M_{i=1} \nabla f_i \left( x^{[t-1]}_i \right) \right\|^2 \right]  \\
        & + \frac{2}{M^2} \sum^M_{i=1} \mathbb{E} \left[ \left\| \nabla F_i \left( x^{[t-1]}_i; \xi_{i,t} \right) - \nabla f_i \left( x^{[t-1]}_i \right)  \right\|^2 \right]  \\
        \stackrel{\scriptsize{\circled{5}}} \leq & \frac{4C^2}{\omega^4_{\min}} + \frac{4 \sigma^2 d}{\omega^4_{\min}} + \frac{2 \zeta^2}{M} + 2 \mathbb{E} \left[ \left\| \frac{1}{M} \sum^M_{i=1} \nabla f_i \left( x^{[t-1]}_i \right) \right\|^2 \right]
    \end{align*}
    where $\scriptsize{\circled{1}}$ follows from the basic inequality $\|a + b\|^2 \leq 2\|a\|^2 + 2\|b\|^2 $ for any vectors $a$ and $b$; $\scriptsize{\circled{2}}$ follows from \textbf{Lemma}~\ref{lem:bdavggrad};
    $\scriptsize{\circled{3}}$ based on the facts that $\mathbb{E} \left[ g^{[t]}_{i,i} \right] = \mathbb{E} \left[ \nabla F_i \left( x^{[t-1]}_i; \xi_{i,t} \right)  \right] =
    \nabla f_i \left( x^{[t-1]}_i \right)$ and $\mathbb{E} \left[ \|a\|^2 \right] = \mathbb{E} \left[ \|a-\mathbb{E}[a]\|^2 \right] + \mathbb{E}\left[ \|\mathbb{E}[a]\|^2 \right]$ holds for any random vector $a$; $\scriptsize{\circled{4}}$ follows from the facts that $z_i := \nabla F_i \left( x^{[t-1]}_i; \xi_{i,t} \right) - \nabla f_i \left( x^{[t-1]}_i \right)$'s are independent random vectors (with respect to $i$) with zero means and $\mathbb{E} \left[ \left\| \sum^M_{i=1} z_i \right\|^2 \right] = \sum^M_{i=1} \mathbb{E} \left[ \left\| z_i \right\|^2 \right]$; $\scriptsize{\circled{5}}$ follows from \textbf{Assumption}~\ref{ass:bd-varia}.
 \end{proof}

 To prove $\textbf{Theorem}~\ref{thm:convergence-main}$, we first present an auxiliary sequence $\left\{ \bar{S}^{[t]} \right\}_{t \geq 0}$ such that
    \begin{align} \label{eq:auxseq}
        \bar{S}^{[t]} = 
        \begin{cases}
          \bar{x}^{[t]} & \text{with}~t = 0  \\
          \frac{1}{1-\alpha} \bar{x}^{[t]} - \frac{\alpha}{1-\alpha} \bar{x}^{[t-1]} & \text{with}~t \geq 1
        \end{cases}
    \end{align}
 where $\bar{x}^{[t]} = \frac{1}{M} \sum^M_{i=1} x^{[t]}_i$. In the following, the initial value will be set to $0$. Based on \textbf{Algorithm}~\ref{alg:pdsl}, we have the following useful $\textbf{Lemmas}$.

 \begin{lemma} \label{lem:seqdiff}
    According to \textbf{Algorithm}~\ref{alg:pdsl}, for any $t \geq 1$, we have 
    \begin{equation}  \label{eq:seqdiff}
        \bar{S}^{[t]} - \bar{S}^{[t-1]} = \frac{-\gamma}{M(1-\alpha)} \sum^M_{i=1} \bar{g}_i^{[t]}
    \end{equation}
 \end{lemma}
 \begin{proof}
    As shown in Line~\ref{ln:uplocmoment02} and Line~\ref{ln:uplocmoment01} in \textbf{Algorithm}~\ref{alg:pdsl}, we have
    \begin{align} \label{eq:seqdiff00}
        \bar{u}^{[t]} & = \frac{1}{M} \sum^{M}_{i=1} u^{[t]}_i
        = \frac{1}{M} \sum^{M}_{i=1} \sum_{j \in \mathcal{M}_i} \omega_{i,j} \hat{u}^{[t]}_j 
        = \frac{1}{M} \sum^{M}_{i=1} \hat{u}^{[t]}_i \left(\sum^{M}_{j=1} w_{j,i} \right)  \nonumber\\
        & = \frac{1}{M} \sum^{M}_{i=1} \hat{u}^{[t]}_i  
        = \frac{1}{M} \sum^{M}_{i=1} \left( \alpha u^{[t-1]}_i + \bar{g}^{[t]}_i \right)
        = \alpha \bar{u}^{[t-1]} + \frac{1}{M} \sum^{M}_{i=1} \bar{g}^{[t]}_i
    \end{align}
    Moreover, according to Line~\ref{ln:uplocpara02} and Line~\ref{ln:uplocpara01} in \textbf{Algorithm}~\ref{alg:pdsl}, we have
    \begin{align} \label{eq:seqdiff01}
        \bar{x}^{[t]} & = \frac{1}{M} \sum^{M}_{i=1} x^{[t]}_i
        = \frac{1}{M} \sum^M_{i=1} \sum_{j \in \mathcal{M}_i} \omega_{i,j} \hat{x}^{[t]}_j 
        = \frac{1}{M} \sum^M_{i=1} \hat{x}^{[t]}_i \left(\sum^M_{j=1} w_{j,i} \right) \nonumber\\
        & = \frac{1}{M} \sum^M_{i=1} \hat{x}^{[t]}_i
        = \frac{1}{M} \sum^M_{i=1} \left(x^{[t-1]}_i - \gamma \hat{u}^{t}_i \right)
        = \bar{x}^{[t-1]} - \gamma \bar{u}^{[t]}
    \end{align}
    We then prove this lemma by induction. When $t=1$, we have
    \begin{align*}
        \bar{S}^{[1]} - \bar{S}^{[0]} =& \frac{1}{1-\alpha} \bar{x}^{[1]} - \frac{\alpha}{1-\alpha} \bar{x}^{[0]} - \bar{x}^{[0]}  \nonumber\\
        =& \frac{1}{1-\alpha} \left( \bar{x}^{[1]} - \bar{x}^{[0]} \right)  
        = \frac{-\gamma \bar{u}^{[1]}}{1-\alpha}  \nonumber\\
        =& \frac{-\gamma}{M(1-\alpha)} \sum^M_{i=1} \bar{g}_i^{[1]}  \nonumber
    \end{align*}
    by considering $\gamma \bar{u}^{[1]} = x^{[0]}-x^{[1]}$ due to (\ref{eq:seqdiff01}) and $\bar{u}^{[1]} =  \frac{1}{M} \sum^M_{i=1} \bar{g}_i^{[1]}$ due to (\ref{eq:seqdiff00}). For any $t > 1$, we have
    \begin{align*}
        \bar{S}^{[t]} - \bar{S}^{[t-1]} 
        =& \left( \frac{1}{1-\alpha} \bar{x}^{[t]} - \frac{\alpha}{1-\alpha} \bar{x}^{[t-1]} \right) - \left( \frac{1}{1-\alpha} \bar{x}^{[t-1]} - \frac{\alpha}{1-\alpha} \bar{x}^{[t-2]} \right)  \nonumber\\
        =& \frac{1}{1-\alpha} \left( \bar{x}^{[t]} - \bar{x}^{[t-1]} \right) - \frac{\alpha}{1-\alpha} \left( \bar{x}^{[t-1]} - \bar{x}^{[t-2]} \right)  \nonumber\\
        =& \frac{1}{1-\alpha} \left( -\gamma \bar{u}^{[t]} - \alpha \left(-\gamma \bar{u}^{[t-1]} \right) \right)  \nonumber\\
        =& \frac{-\gamma}{M(1-\alpha)} \sum^M_{i=1} \bar{g}_i^{[t]}  \nonumber
    \end{align*}
    which finally completes the proof.
 \end{proof}

 \begin{lemma} \label{lem:sandx}
    For any $T \geq 1$, we have
    \begin{equation}
        \sum^{T}_{t=1} \left\| \bar{S}^{[t]} - \bar{x}^{[t]} \right\|^2 \leq \frac{\gamma^2 \alpha^2}{(1-\alpha)^4} \sum^{T}_{t=1} \left\| \frac{1}{M}\sum^{M}_{i=1} \bar{g}_i^{[t]} \right\|^2 
    \end{equation}
 \end{lemma}
 \begin{proof}
    By applying (\ref{eq:seqdiff00}) in \textbf{Lemma}~\ref{lem:seqdiff} recursively, we have
    \begin{equation} \label{eq:up-bar-u}
        \bar{u}^{[t]} = \sum^{t}_{\tau=1}   \alpha^{t-\tau} \left( \frac{1}{M}\sum^{M}_{i=1} \bar{g}_i^{[\tau]} \right), ~\forall t \geq 1
    \end{equation}
    and thus, 
    \begin{align} \label{eq:sandx01}
        \bar{S}^{[t]} - \bar{x}^{[t]} = \frac{-\gamma \alpha}{1-\alpha} \bar{u}^{[t]} = \frac{-\gamma \alpha}{1-\alpha} \sum^{t}_{\tau=1} \alpha^{t-\tau} \left( \frac{1}{M}\sum^{M}_{i=1} \bar{g}_i^{[\tau]} \right) 
    \end{align}
    Based on (\ref{eq:sandx01}), we have
    \begin{align*}
        \left\| \bar{S}^{[t]} - \bar{x}^{[t]} \right\|^2 
        =& \left\| \frac{ -\gamma \alpha}{1-\alpha} \sum^{t}_{\tau=1} \alpha^{t-\tau} \left( \frac{1}{M}\sum^{M}_{i=1} \bar{g}_i^{[\tau]} \right) \right\|^2  \\
        \stackrel{\scriptsize{\circled{1}}}=& \frac{ \gamma^2 \alpha^2}{(1-\alpha)^2} \phi^2_t \left\| \sum^{t}_{\tau=1} \frac{\alpha^{t-\tau}}{\phi_t} \left( \frac{1}{M}\sum^{M}_{i=1} \bar{g}_i^{[\tau]} \right) \right\|^2  \\
        \stackrel{\scriptsize{\circled{2}}}\leq& \frac{ \gamma^2 \alpha^2}{(1-\alpha)^2} \phi^2_t \sum^{t}_{\tau=1} \frac{\alpha^{t-\tau}}{\phi_t} \left\| \frac{1}{M}\sum^{M}_{i=1} \bar{g}_i^{[\tau]} \right\|^2  \\
        \stackrel{\scriptsize{\circled{3}}}\leq& \frac{\gamma^2 \alpha^2}{(1-\alpha)^3} \sum^{t}_{\tau=1} \alpha^{t-\tau} \left\| \frac{1}{M}\sum^{M}_{i=1} \bar{g}_i^{[\tau]} \right\|^2
    \end{align*}
    where we have $\scriptsize{\circled{1}}$ by letting  $\phi_t = \sum^{t}_{\tau=1} \alpha^{t-\tau} = \frac{1-\alpha^t}{1-\alpha}$; $\scriptsize{\circled{2}}$ follows from the convexity of $\|\cdot\|^2$ and Jensen's Inequality; and $\scriptsize{\circled{3}}$ by considering $ 1-\alpha^t \leq 1$.

    Furthermore, by summing $\left\| \bar{S}^{[t]} - \bar{x}^{[t]} \right\|^2$ over $ t = 1,2, \cdots, T$, we obtain
    \begin{align*}
        \sum^{T}_{t=1} \left\| \bar{S}^{[t]} - \bar{x}^{[t]} \right\|^2
        \leq & \frac{\gamma^2 \alpha^2}{(1-\alpha)^3} \sum^{T}_{t=1} \sum^{t}_{\tau=1} \alpha^{t-\tau} \left\| \frac{1}{M}\sum^{M}_{i=1} \bar{g}_i^{[\tau]} \right\|^2  \nonumber\\
        \leq & \frac{\gamma^2 \alpha^2}{(1-\alpha)^3} \sum^{T}_{t=1} \left( \left\| \frac{1}{M}\sum^{M}_{i=1} \bar{g}_i^{[t]} \right\|^2 \sum^{T}_{\tau=t} \alpha^{\tau-t} \right)  \nonumber\\
        \stackrel{\scriptsize{\circled{1}}}\leq & \frac{\gamma^2 \alpha^2}{(1-\alpha)^4} \sum^{T}_{t=1} \left\| \frac{1}{M}\sum^{M}_{i=1} \bar{g}_i^{[t]} \right\|^2  
    \end{align*}
    where $\scriptsize{\circled{1}}$ follows from $\sum^{T}_{\tau=t} \alpha^{\tau-t} = \frac{1 - \alpha^{T-t+1}}{1-\alpha} \leq \frac{1}{1-\alpha}$.
 \end{proof}

 Before proceeding with our proof, we first show some key notations (see (\ref{eq:notation})) and facts (i.e., \textbf{Fact}~\ref{fact:fact-1} and \textbf{Fact}~\ref{fact:fact-2}) which will be helpful in the subsequent discussion. Similar facts are adopted in \cite{EsfandiariTJBHHS-ICML21, LianZZHZL-NeurIPS17}.
 \begin{align} \label{eq:notation}
    \begin{cases}
        \bar{\mathbf{G}}^{[t]} &\triangleq \left[ \bar{g}_1^{[t]}, \bar{g}_2^{[t]}, \cdots, \bar{g}_M^{[t]} \right]  \vspace{1ex}\\
        \mathbf{U}^{[t]} &\triangleq \left[ u^{[t]}_1, u^{[t]}_2, \cdots,  u^{[t]}_M \right]  \vspace{1ex}\\
        \hat{\mathbf{U}}^{[t]} &\triangleq \left[ \hat{u}^{[t]}_1, \hat{u}^{[t]}_2, \cdots,  \hat{u}^{[t]}_M \right]  \vspace{1ex}\\
        \mathbf{X}^{[t]} &\triangleq \left[ x^{[t]}_1, x^{[t]}_2, \cdots, x^{[t]}_M \right]   \vspace{1ex}\\
        \mathbf{G}^{[t]} &\triangleq \left[ g^{[t]}_{1,1}, g^{[t]}_{2,2}, \cdots, g^{[t]}_{M,M} \right]  \vspace{1ex}\\
        \mathbf{H}^{[t]} &\triangleq \left[ \nabla f_1\left(x^{[t-1]}_1\right), \cdots, \nabla f_M\left(x^{[t-1]}_M\right)  \right]  \vspace{1ex}\\   
        \mathbf{J}^{[t]} &\triangleq \left[ \nabla f_1\left(\bar{x}^{[t-1]}\right), \cdots, \nabla f_M\left(\bar{x}^{[t-1]}\right)  \right] 
    \end{cases}
 \end{align}

 \begin{fact} \label{fact:fact-1}
    Let $\mathbf{Q} = \frac{1}{M} \mathbf{1}\mathbf{1}^{\mathrm T}$ where $\mathbf{1}$ denotes a $M$-dimensional column vector with all entries being $1$. For any doubly stochastic matrix $\mathbf{W}$, we have:
    \begin{itemize}
        \item $\mathbf{Q} \mathbf{W} = \mathbf{W} \mathbf{Q}$;
        \item $(\mathbf{I}- \mathbf{Q}) \mathbf{W} = \mathbf{W} (\mathbf{I}- \mathbf{Q})$;
        \item For any integer $t \geq 1$, $\left\| (\mathbf{I}- \mathbf{Q}) \mathbf{W}^t \right\|_{\mathfrak{s}} \leq (\sqrt{\rho})^t$, where $\|\cdot\|_{\mathfrak{s}}$ is the spectrum norm.
    \end{itemize}
 \end{fact}

 \begin{fact} \label{fact:fact-2}
    Let $\mathbf{B}_1, \mathbf{B}_2, \cdots, \mathbf{B}_m$ be $m$ arbitrary real-valued matrices such that $\|\mathbf{B}_i\|^2_{\mathfrak{F}} = \sum^M_{k=1} \| \mathbf{b}_{i,k} \|^2$, where $\|\cdot\|_{\mathfrak{F}}$ is the Frobenius norm, and $ \mathbf{b}_{i,k}$ is the $k$-th column of $\mathbf{B}_i$. We have 
    \begin{align}
        \left\| \sum^m_{i=1} \mathbf{B}_i \right\|^2_{\mathfrak{F}} \leq \sum^m_{i=1} \sum^m_{j=1} \left\| \mathbf{B}_i \right\|_{\mathfrak{F}} \left\| \mathbf{B}_j \right\|_{\mathfrak{F}}
    \end{align}
 \end{fact}

 Next, in \textbf{Lemma}~\ref{lem:barG-G}, we present the bound of $\mathbb{E} \left[ \left\| \bar{\mathbf{G}}^{[t]} - \mathbf{G}^{[t]} \right\|^2_{\mathfrak{F}} \right]$, based on which, in \textbf{Lemma}~\ref{lem:barx-x}, we demonstrate that $\sum^{T}_{t=1} \frac{1}{M} \sum^M_{i=1} \mathbb{E} \left[ \left\| \bar{x}^{[t]} - x^{[t]}_i \right\|^2 \right]$ is bounded by $\sum^{T}_{t=1} \mathbb{E} \left[ \left\| \frac{1}{M} \sum^M_{i=1} \nabla f_i \left( x^{[t-1]}_i \right) \right\|^2 \right]$.

 \begin{lemma} \label{lem:barG-G} 
    When all \textbf{Assumptions}~\ref{ass:L-smooth}$\sim$\ref{ass:ds_mat} hold, for any $t \geq 1$, we have
    \begin{align} \label{eq:barG-G}
        & \mathbb{E} \left[ \left\| \bar{\mathbf{G}}^{[t]} - \mathbf{G}^{[t]} \right\|^2_{\mathfrak{F}} \right]  \nonumber\\
        \leq & \frac{4M C^2}{\omega^4_{\min}} + \frac{4M \sigma^2 d}{\omega^4_{\min}} + 2M \zeta^2 + 8M \kappa^2  \nonumber\\
        & + 16L^2 \sum^M_{i=1} \mathbb{E} \left[ \left\| x^{[t-1]}_i - \bar{x}^{[t-1]} \right\|^2 \right]   \nonumber\\
        & + 8M \mathbb{E} \left[ \left\| \frac{1}{M} \sum^M_{i=1} \nabla f_i \left( x^{[t-1]}_i \right) \right\|^2 \right]
    \end{align}
 \end{lemma}
 \begin{proof}
    According to (\ref{eq:notation}) and \textbf{Fact}~\ref{fact:fact-2}, we have
    \begin{align*}
        \mathbb{E} \left[ \left\| \bar{\mathbf{G}}^{[t]} - \mathbf{G}^{[t]} \right\|^2_{\mathfrak{F}} \right]  
        = & \sum^M_{i=1} \mathbb{E} \left[ \left\| \bar{g}^{[t]}_i - g^{[t]}_{i,i} \right\|^2 \right]  \\
        \stackrel{\scriptsize{\circled{1}}} \leq & 2 \sum^M_{i=1} \mathbb{E} \left[ \left\| \bar{g}^{[t]}_i \right\|^2 \right] + 2 \sum^M_{i=1} \mathbb{E} \left[ \left\| g^{[t]}_{i,i} \right\|^2 \right]   \\
        \leq & 2 \sum^M_{i=1} \mathbb{E} \left[ \left\| \sum_{j \in \mathcal{M}_i} \pi^{[t]}_{i,j} \left( \frac{g^{[t]}_{j,i}}{\max \left\{1, C^{-1} \left\| g^{[t]}_{j,i} \right\| \right\}} + n^{[t]}_{j,i} \right) \right\|^2 \right]  
        + 2 \sum^M_{i=1} \mathbb{E} \left[ \left\| g^{[t]}_{i,i} \right\|^2 \right]   \\
        \stackrel{\scriptsize{\circled{2}}} \leq & 4 \sum^M_{i=1} \mathbb{E} \left[ \left\| \sum_{j \in \mathcal{M}_i} \frac{\hat{\varphi}^{[t]}_{i,j}}{\omega_{i,j} \sum_{k\in\mathcal{M}_i} \hat{\varphi}^{[t]}_{i,k}} \left( \frac{g^{[t]}_{j,i}}{\max \left\{1, C^{-1} \left\| g^{[t]}_{j,i} \right\| \right\}} \right) \right\|^2 \right]   \\
        & + 4 \sum^M_{i=1} \mathbb{E} \left[ \left\| \sum_{j \in \mathcal{M}_i} \frac{\hat{\varphi}^{[t]}_{i,j}}{\omega_{i,j} \sum_{k\in\mathcal{M}_i} \hat{\varphi}^{[t]}_{i,k}} n^{[t]}_{j,i}  \right\|^2 \right] 
        + 2 \sum^M_{i=1} \mathbb{E} \left[ \left\| g^{[t]}_{i,i} \right\|^2 \right]  \\
        \stackrel{\scriptsize{\circled{3}}} \leq & 4 \sum^M_{i=1} \sum_{j \in \mathcal{M}_i} \omega_{i,j} \mathbb{E} \left[ \left\| \frac{\hat{\varphi}^{[t]}_{i,j}}{\omega^2_{i,j} \sum_{k\in\mathcal{M}_i} \hat{\varphi}^{[t]}_{i,k}} \right\|^2 \left\| \frac{g^{[t]}_{j,i}}{\max \left\{1, C^{-1} \left\| g^{[t]}_{j,i} \right\| \right\}} \right\|^2  \right]  \\
        & + 4 \sum^M_{i=1} \sum_{j \in \mathcal{M}_i} \omega_{i,j} \mathbb{E} \left[ \left\| \frac{\hat{\varphi}^{[t]}_{i,j}}{\omega^2_{i,j} \sum_{k\in\mathcal{M}_i} \hat{\varphi}^{[t]}_{i,k}} \right\|^2 \left\| n^{[t]}_{j,i} \right\|^2 \right]  
        + 2 \sum^M_{i=1} \mathbb{E} \left[ \left\| g^{[t]}_{i,i} \right\|^2 \right]  \\
        \stackrel{\scriptsize{\circled{4}}} \leq & 4 \sum^M_{i=1} \sum_{j \in \mathcal{M}_i} \omega_{i,j} \frac{C^2}{\omega^4_{i,j}} + 4 \sum^M_{i=1} \sum_{j \in \mathcal{M}_i} \omega_{i,j} \frac{\sigma^2 d}{\omega^4_{i,j}} + 2 \sum^M_{i=1} \mathbb{E} \left[ \left\| g^{[t]}_{i,i} \right\|^2 \right]   \\
        \stackrel{\scriptsize{\circled{5}}} \leq & 4 \sum^M_{i=1} \sum_{j \in \mathcal{M}_i} \omega_{i,j} \frac{C^2}{\omega^4_{i,j}} + 4 \sum^M_{i=1} \sum_{j \in \mathcal{M}_i} \omega_{i,j} \frac{\sigma^2 d}{\omega^4_{i,j}}  \\
        & + 2 \sum^M_{i=1} \mathbb{E} \left[ \left\|  \nabla F_i \left( x^{[t-1]}_i; \xi_{i,t} \right)- \nabla f_i \left(x^{[t-1]}_i \right)  \right\|^2 \right]   
        + 2 \sum^{M}_{i=1} \mathbb{E} \left[ \left\| \nabla f_i \left( x^{[t-1]}_i \right) \right\|^2 \right]  \\
        \leq & 4 \sum^M_{i=1} \sum_{j \in \mathcal{M}_i} \omega_{i,j} \frac{C^2}{\omega^4_{i,j}} + 4 \sum^M_{i=1} \sum_{j \in \mathcal{M}_i} \omega_{i,j} \frac{\sigma^2 d}{\omega^4_{i,j}}  \\
        & + 2 \sum^M_{i=1} \mathbb{E} \left[ \left\|  \nabla F_i \left( x^{[t-1]}_i; \xi_{i,t} \right)- \nabla f_i \left(x^{[t-1]}_i \right)  \right\|^2 \right]   \\
        & + 2\sum^M_{i=1} \mathbb{E} \left[ \left\| \begin{aligned} & \nabla f_i \left( x^{[t-1]}_i \right) - \nabla f_i \left( \bar{x}^{[t-1]} \right) + \nabla f_i \left( \bar{x}^{[t-1]} \right) \\ & - \frac{1}{M} \sum^M_{j=1} \nabla f_j \left( \bar{x}^{[t-1]} \right) + \frac{1}{M} \sum^M_{j=1} \nabla f_j \left( \bar{x}^{[t-1]} \right)  \\ & - \frac{1}{M} \sum^M_{j=1} \nabla f_j \left( x^{[t-1]}_j \right) + \frac{1}{M} \sum^M_{j=1} \nabla f_j \left( x^{[t-1]}_j \right) \end{aligned} \right\|^2 \right]    \\
        \stackrel{\scriptsize{\circled{6}}} \leq & 4 \sum^M_{i=1} \sum_{j \in \mathcal{M}_i} \omega_{i,j} \frac{C^2}{\omega^4_{i,j}} + 4 \sum^M_{i=1} \sum_{j \in \mathcal{M}_i} \omega_{i,j} \frac{\sigma^2 d}{\omega^4_{i,j}}  \\
        & + 2 \sum^M_{i=1} \mathbb{E} \left[ \left\|  \nabla F_i \left( x^{[t-1]}_i; \xi_{i,t} \right)- \nabla f_i \left(x^{[t-1]}_i \right)  \right\|^2 \right]   \\
        & + 8 \sum^M_{i=1} \mathbb{E} \left[ \left\| \nabla f_i \left( x^{[t-1]}_i \right) - \nabla f_i \left( \bar{x}^{[t-1]} \right) \right\|^2 \right]  \\
        & + 8 \sum^M_{i=1} \mathbb{E} \left[ \left\| \nabla f_i \left( \bar{x}^{[t-1]} \right) - \frac{1}{M} \sum^M_{j=1} \nabla f_j \left( \bar{x}^{[t-1]} \right) \right\|^2 \right]  \\
        & + 8 \sum^M_{i=1} \mathbb{E} \left[ \left\| \frac{1}{M} \sum^M_{j=1} \nabla f_j \left( \bar{x}^{[t-1]} \right) - \frac{1}{M} \sum^M_{j=1} \nabla f_j \left( x^{[t-1]}_j \right) \right\|^2 \right]  \\
        & + 8 \sum^M_{i=1} \mathbb{E} \left[ \left\| \frac{1}{M} \sum^M_{j=1} \nabla f_j \left( x^{[t-1]}_j \right) \right\|^2 \right]  \\
        \stackrel{\scriptsize{\circled{7}}} \leq & 4 \sum^M_{i=1} \sum_{j \in \mathcal{M}_i} \omega_{i,j} \frac{C^2}{\omega^4_{i,j}} + 4 \sum^M_{i=1} \sum_{j \in \mathcal{M}_i} \omega_{i,j} \frac{\sigma^2 d}{\omega^4_{i,j}} + 2M \zeta^2 \\
        & + 8L^2 \sum^M_{i=1} \mathbb{E} \left[ \left\| x^{[t-1]}_i - \bar{x}^{[t-1]} \right\|^2 \right] + 8M \kappa^2  \\
        & + 8 \sum^M_{i=1} \frac{1}{M} \sum^M_{j=1} L^2 \mathbb{E} \left[ \left\| \bar{x}^{[t-1]} - x^{[t-1]}_j \right\|^2 \right]  \\
        & + 8M \mathbb{E} \left[ \left\| \frac{1}{M} \sum^M_{j=1} \nabla f_j \left( x^{[t-1]}_j \right) \right\|^2 \right]  \\
        \stackrel{\scriptsize{\circled{8}}} \leq & \frac{4M C^2}{\omega^4_{\min}} + \frac{4M \sigma^2 d}{\omega^4_{\min}} + 2M \zeta^2 + 8M \kappa^2 \\
        & + 16L^2 \sum^M_{i=1} \mathbb{E} \left[ \left\| x^{[t-1]}_i - \bar{x}^{[t-1]} \right\|^2 \right]  \\
        & + 8M \mathbb{E} \left[ \left\| \frac{1}{M} \sum^M_{i=1} \nabla f_i \left( x^{[t-1]}_i \right) \right\|^2 \right]
    \end{align*}
    where $\scriptsize{\circled{1}}$ and $\scriptsize{\circled{2}}$ follow from $\|a+b\|^2 \leq 2\|a\|^2 + 2\|b\|^2$; $\scriptsize{\circled{3}}$ follows from the convexity of $\|\cdot\|^2$ and \textit{Jensen's Inequality}; $\scriptsize{\circled{4}}$ follows from $\frac{\hat{\varphi}^{[t]}_{i,j}} {\sum_{k\in\mathcal{M}_i} \hat{\varphi}^{[t]}_{i,k}} \leq 1$, and the characteristic of clipping process that $\left\| \frac{g^{[t]}_{j,i}}{\max \left\{1, C^{-1} \left\| g^{[t]}_{j,i} \right\| \right\}} \right\|^2 \leq C^2$, and the property of Gaussian noises that $\mathbb{E} \left[ \left\| n^{[t]}_{j,i} \right\|^2 \right] = \sigma^2 d$; $\scriptsize{\circled{5}}$ follows from the facts that $\mathbb{E} \left[ g^{[t]}_{i,i} \right] = \mathbb{E} \left[ \nabla F_i \left( x^{[t-1]}_i; \xi_{i,t} \right) \right] = \nabla f_i \left( x^{[t-1]}_i \right)$ and $\mathbb{E} \left[ \|a\|^2 \right] = \mathbb{E} \left[ \|a-\mathbb{E}[a]\|^2 \right] + \mathbb{E}\left[ \|\mathbb{E}[a]\|^2 \right]$ holds for any random vector $a$; $\scriptsize{\circled{6}}$ follows from $\|a + b + c + d \|^2 \leq 4\|a\|^2 + 4\|b\|^2 + 4\|c\|^2 + 4\|d\|^2 $ for any vectors $a,b,c,d$; $\scriptsize{\circled{7}}$ follows from \textbf{Assumption}~\ref{ass:bd-varia}, and $\left\| \nabla f_i \left( x^{[t-1]}_i \right) - \nabla f_i \left( \bar{x}^{[t-1]} \right) \right\|^2 \leq L^2 \left\| x^{[t-1]}_i - \bar{x}^{[t-1]} \right\|^2 $ by the smoothness of $f_i$ in \textbf{Assumption}~\ref{ass:L-smooth}, and $ \left\| \frac{1}{M} \sum^M_{j=1} \nabla f_j(\bar{x}^{[t-1]}) - \frac{1}{M} \sum^M_{j=1} \nabla f_j({x}^{[t-1]}_j) \right\|^2 \leq \frac{1}{M} \sum^M_{j=1} \left\| \nabla f_j(\bar{x}^{[t-1]}) - \nabla f_j({x}^{[t-1]}_j) \right\|^2 \leq  L^2 \frac{1}{M}\sum^M_{j=1} \left\| \bar{x}^{[t-1]} - {x}^{[t-1]}_j \right\|^2 $ where the first inequality holds due to the convexity of $\|\cdot\|^2$ and \textit{Jensen's Inequality}; $\scriptsize{\circled{8}}$ follows from $ \omega_{\min} = \min_{i \in \mathcal{M}, j \in \mathcal{M}_i}  \omega_{i,j}$.
 \end{proof}

 \begin{lemma} \label{lem:barx-x} 
    Let \textbf{Assumptions}~\ref{ass:L-smooth}$\sim$\ref{ass:ds_mat} hold and  $\left\{ \bar{x}^{[t]} \right\}_{t \geq 1}$ be the sequence obtained by the iterations of \textbf{Algorithm}~\ref{alg:pdsl}. If momentum coefficient $\alpha \in [0,1)$ and learning rate 
    \begin{equation}
        \gamma \leq \frac{(1-\alpha)(1-\sqrt{\rho})}{2\sqrt{26} L}
    \end{equation}
    then for any $T \geq 1$, we have
    \begin{align} \label{eq:barx-x}
        & \sum^{T}_{t=1} \frac{1}{M} \sum^M_{i=1} \mathbb{E} \left[ \left\| \bar{x}^{[t]} - x^{[t]}_i \right\|^2 \right]  \nonumber\\
        \leq & T \frac{ \frac{16\gamma^2 }{\omega^4_{\min}} \left( C^2 + \sigma^2 d \right)}{(1-\alpha)^2 (1-\sqrt{\rho})^2} + T \frac{4 \gamma^2  \left( 7\zeta^2 + 13\kappa^2 \right)}{(1-\alpha)^2 (1-\sqrt{\rho})^2}   \nonumber\\ 
        & + \frac{52\gamma^2}{(1-\alpha)^2 (1-\sqrt{\rho})^2} \sum^{T}_{t=1} \mathbb{E} \left[ \left\| \frac{1}{M} \sum^M_{i=1} \nabla f_i \left( x^{[t-1]}_i \right) \right\|^2 \right]  
    \end{align}
 \end{lemma}
 \begin{proof}
    Recall the definition of $\mathbf{U}^{[t]}$ and $\hat{\mathbf{U}}^{[t]}$, according to Line~\ref{ln:uplocmoment02} and Line~\ref{ln:uplocmoment01} in \textbf{Algorithm}~\ref{alg:pdsl}, for all $t \geq 1$, we have
    \begin{align} \label{eq:seqdiff00mat}
        \mathbf{U}^{[t]} & = \hat{\mathbf{U}}^{[t]} \mathbf{W}
        = \left( \alpha \mathbf{U}^{[t-1]} + \bar{\mathbf{G}}^{[t]} \right) \mathbf{W} 
    \end{align}
    By applying (\ref{eq:seqdiff00mat}) recursively, we obtain
    \begin{align}
        \mathbf{U}^{[t]} & = \alpha^t \mathbf{U}^{[0]} \mathbf{W}^t + \sum^{t}_{\tau=1} \bar{\mathbf{G}}^{[\tau]} \alpha^{t-\tau} \mathbf{W}^{t-\tau+1} 
        \stackrel{\mathbf{U}^{[0]} = 0} = \sum^{t}_{\tau=1} \bar{\mathbf{G}}^{[\tau]} \alpha^{t-\tau} \mathbf{W}^{t-\tau+1}
    \end{align}
    Recalling the definition of $\mathbf{X}^{[t]}$, according to Line~\ref{ln:uplocpara02} and Line~\ref{ln:uplocpara01} in \textbf{Algorithm}~\ref{alg:pdsl}, for all $t \geq 1$, we have
    \begin{align} \label{eq:seqdiff01mat}
        \mathbf{X}^{[t]} & = \left( \mathbf{X}^{[t-1]} - \gamma \hat{\mathbf{U}}^{[t]} \right) \mathbf{W}
        = \mathbf{X}^{[t-1]} \mathbf{W} - \gamma \hat{\mathbf{U}}^{[t]} \mathbf{W} 
        = \mathbf{X}^{[t-1]} \mathbf{W} - \gamma \mathbf{U}^{[t]}
    \end{align}
    By applying (\ref{eq:seqdiff01mat}) recursively, we can obtain 
    \begin{align}
        \mathbf{X}^{[t]} = \mathbf{X}^{[0]}  \mathbf{W}^{t} - \gamma \sum^t_{\tau=1} \mathbf{U}^{[\tau]} \mathbf{W}^{t-\tau}  
    \end{align}
    Multiplying both sides by $(\mathbf{I} - \mathbf{Q})$ with $\mathbf{Q}$ defined in \textbf{Fact}~\ref{fact:fact-1}, we have
    \begin{align}
        \mathbf{X}^{[t]} (\mathbf{I} - \mathbf{Q}) &= \mathbf{X}^{[0]} \mathbf{W}^t (\mathbf{I} - \mathbf{Q}) - \gamma \sum^t_{\tau=1} \mathbf{U}^{[\tau]} \mathbf{W}^{t-\tau} (\mathbf{I} - \mathbf{Q})  \nonumber\\
        & \stackrel{\scriptsize{\circled{1}}} = \mathbf{X}^{[0]} (\mathbf{I} - \mathbf{Q}) \mathbf{W}^t - \gamma \sum^t_{\tau=1} \mathbf{U}^{[\tau]} (\mathbf{I} - \mathbf{Q}) \mathbf{W}^{t-\tau}
    \end{align}
    where $\scriptsize{\circled{1}}$ follows from $\mathbf{W} (\mathbf{I} - \mathbf{Q}) = (\mathbf{I} - \mathbf{Q}) \mathbf{W} $ in \textbf{Fact}~\ref{fact:fact-1}.

    By considering $\mathbf{X}^{[0]} (\mathbf{I} - \mathbf{Q}) = 0$ due to all columns of $\mathbf{X}^{[0]}$ are identical, we have
    \begin{align} \label{eq:XI-Q-00}
        & \mathbb{E} \left[ \left\| \mathbf{X}^{[t]} (\mathbf{I} - \mathbf{Q}) \right\|^2_{\mathfrak{F}} \right] \nonumber\\
        = & \mathbb{E} \left[ \left\| -\gamma \sum^t_{\tau=1} \mathbf{U}^{[\tau]} (\mathbf{I} - \mathbf{Q}) \mathbf{W}^{t-\tau} \right\|^2_{\mathfrak{F}} \right] \nonumber\\
        = & \mathbb{E} \left[ \left\| -\gamma \sum^t_{\tau=1} \sum^{\tau}_{\tau'=1} \bar{\mathbf{G}}^{[\tau']} \alpha^{\tau-\tau'} \mathbf{W}^{\tau-\tau'+1} (\mathbf{I} - \mathbf{Q}) \mathbf{W}^{t-\tau} \right\|^2_{\mathfrak{F}} \right] \nonumber\\
        = & \mathbb{E} \left[ \left\| -\gamma \sum^t_{\tau=1} \sum^{\tau}_{\tau'=1} \bar{\mathbf{G}}^{[\tau']} \alpha^{\tau-\tau'} \mathbf{W}^{t-\tau'+1} (\mathbf{I} - \mathbf{Q}) \right\|^2_{\mathfrak{F}} \right] \nonumber\\
        = & \mathbb{E} \left[ \left\| -\gamma \sum^{t}_{\tau=1} \bar{\mathbf{G}}^{[\tau]} ( \sum^{t}_{\tau'= \tau} \alpha^{\tau'-\tau} \mathbf{W}^{t-\tau+1} ) (\mathbf{I} - \mathbf{Q}) \right\|^2_{\mathfrak{F}} \right]  \nonumber\\
        = & \mathbb{E} \left[ \left\| -\gamma \sum^{t}_{\tau=1} \bar{\mathbf{G}}^{[\tau]} ( \sum^{t}_{\tau'= \tau} \alpha^{\tau'-\tau} ) \mathbf{W}^{t-\tau+1} (\mathbf{I} - \mathbf{Q}) \right\|^2_{\mathfrak{F}} \right]  \nonumber\\
        = & \mathbb{E} \left[ \left\| -\gamma \sum^{t}_{\tau=1} \frac{1-\alpha^{t-\tau}}{1-\alpha} \bar{\mathbf{G}}^{[\tau]} (\mathbf{I} - \mathbf{Q}) \mathbf{W}^{t-\tau+1} \right\|^2_{\mathfrak{F}} \right]  \nonumber\\
        = & \gamma^2 \mathbb{E} \left[ \left\| \sum^{t}_{\tau=1} \frac{1-\alpha^{t-\tau}}{1-\alpha} \left( ( \bar{\mathbf{G}}^{[\tau]} - \mathbf{G}^{[\tau]}) (\mathbf{I} - \mathbf{Q}) \mathbf{W}^{t-\tau+1} + \mathbf{G}^{[\tau]} (\mathbf{I} - \mathbf{Q}) \mathbf{W}^{t-\tau+1} \right) \right\|^2_{\mathfrak{F}} \right]  \nonumber\\
        \stackrel{\scriptsize{\circled{1}}} \leq & 2\gamma^2 \underbrace{ \mathbb{E} \left[ \left\| \sum^{t}_{\tau=1} \frac{1-\alpha^{t-\tau}}{1-\alpha} \left( \bar{\mathbf{G}}^{[\tau]} - \mathbf{G}^{[\tau]} \right) \left( \mathbf{I} - \mathbf{Q} \right) \mathbf{W}^{t-\tau+1} \right\|^2_{\mathfrak{F}} \right] }_{\mathrm{\MakeTextUppercase{\romannumeral 1}}} \nonumber\\
        & + 2\gamma^2 \underbrace{ \mathbb{E} \left[ \left\| \sum^{t}_{\tau=1} \frac{1-\alpha^{t-\tau}}{1-\alpha} \mathbf{G}^{[\tau]} (\mathbf{I} - \mathbf{Q}) \mathbf{W}^{t-\tau+1} \right\|^2_{\mathfrak{F}} \right] }_{\mathrm{\MakeTextUppercase{\romannumeral 2}}}
    \end{align}
    where we have $\scriptsize{\circled{1}}$ by considering the inequality $\| \mathbf{A} + \mathbf{B} \|^2_{\mathfrak{F}} \leq 2 \| \mathbf{A} \|^2_{\mathfrak{F}} + 2 \| \mathbf{B} \|^2_{\mathfrak{F}}$.

    Next, we first give the upper bound of term $\mathrm{\MakeTextUppercase{\romannumeral 1}}$
    \begin{align} \label{eq:bdtermI}
        & \mathbb{E} \left[ \left\| \sum^{t}_{\tau=1} \frac{1-\alpha^{t-\tau}}{1-\alpha} \left( \bar{\mathbf{G}}^{[\tau]} - \mathbf{G}^{[\tau]} \right) \left( \mathbf{I} - \mathbf{Q} \right) \mathbf{W}^{t-\tau+1} \right\|^2_{\mathfrak{F}} \right]  \nonumber\\
        \stackrel{\scriptsize{\circled{1}}} \leq & \sum^{t}_{\tau=1} \sum^{t}_{\tau'=1} \mathbb{E} \left[ \left\| \frac{1-\alpha^{t-\tau}}{1-\alpha} ( \bar{\mathbf{G}}^{[\tau]} - \mathbf{G}^{[\tau]} ) ( \mathbf{I} - \mathbf{Q} ) \mathbf{W}^{t-\tau+1} \right\|_{\mathfrak{F}}  \left\| \frac{1-\alpha^{t-\tau'}}{1-\alpha}  ( \bar{\mathbf{G}}^{[\tau']} - \mathbf{G}^{[\tau']} ) ( \mathbf{I} - \mathbf{Q} ) \mathbf{W}^{t-\tau'+1} \right\|_{\mathfrak{F}} \right]   \nonumber\\
        \stackrel{\scriptsize{\circled{2}}} \leq & \frac{1}{(1-\alpha)^2} \sum^{t}_{\tau=1} \sum^{t}_{\tau'=1} \mathbb{E} \left[ \left\| \bar{\mathbf{G}}^{[\tau]} - \mathbf{G}^{[\tau]} \right\|_{\mathfrak{F}} \left\| (\mathbf{I} - \mathbf{Q}) \mathbf{W}^{t-\tau+1} \right\|_{\mathfrak{s}} \left\|  \bar{\mathbf{G}}^{[\tau']} - \mathbf{G}^{[\tau']} \right\|_{\mathfrak{F}} \left\| (\mathbf{I} - \mathbf{Q}) \mathbf{W}^{t-\tau'+1} \right\|_{\mathfrak{s}} \right]   \nonumber\\
        \stackrel{\scriptsize{\circled{3}}} \leq & \frac{1}{(1-\alpha)^2} \sum^{t}_{\tau=1} \sum^{t}_{\tau'=1} \rho^{(t+1-\frac{\tau+\tau'}{2})} \mathbb{E} \left[ \left\| \bar{\mathbf{G}}^{[\tau]} - \mathbf{G}^{[\tau]} \right\|_{\mathfrak{F}} \left\|  \bar{\mathbf{G}}^{[\tau']} - \mathbf{G}^{[\tau']} \right\|_{\mathfrak{F}} \right]    \nonumber\\
        \stackrel{\scriptsize{\circled{4}}} \leq & \frac{1}{(1-\alpha)^2} \sum^{t}_{\tau=1} \sum^{t}_{\tau'=1} \rho^{(t+1-\frac{\tau+\tau'}{2})} \left( \frac{1}{2} \mathbb{E} \left[  \left\| \bar{\mathbf{G}}^{[\tau]} - \mathbf{G}^{[\tau]} \right\|^2_{\mathfrak{F}} \right] + \frac{1}{2} \mathbb{E} \left[ \left\|  \bar{\mathbf{G}}^{[\tau']} - \mathbf{G}^{[\tau']} \right\|^2_{\mathfrak{F}} \right] \right)    \nonumber\\
        \stackrel{\scriptsize{\circled{5}}} \leq & \frac{1}{(1-\alpha)^2 (1-\sqrt{\rho})} \sum^{t}_{\tau=1} \rho^{(\frac{t+1-\tau}{2})} \mathbb{E} \left[ \left\| \bar{\mathbf{G}}^{[\tau]} - \mathbf{G}^{[\tau]} \right\|^2_{\mathfrak{F}} \right]     \nonumber\\
        \stackrel{\scriptsize{\circled{6}}} \leq & \frac{1}{(1-\alpha)^2 (1-\sqrt{\rho})} \sum^{t}_{\tau=1} \rho^{(\frac{t+1-\tau}{2})} \left( \begin{aligned} & \frac{4M C^2}{\omega^4_{\min}} + \frac{4M \sigma^2 d}{\omega^4_{\min}} + 2M \zeta^2 + 8M \kappa^2 \\
        & + 16L^2 \sum^M_{i=1} \mathbb{E} \left[ \left\| x^{[\tau-1]}_i - \bar{x}^{[\tau-1]} \right\|^2 \right]   \\
        & + 8M \mathbb{E} \left[ \left\| \frac{1}{M} \sum^M_{i=1} \nabla f_i \left( x^{[\tau-1]}_i \right) \right\|^2 \right] \end{aligned} \right)    \nonumber\\
        \leq & \frac{ \frac{4M}{\omega^4_{\min}} \left( C^2 + \sigma^2 d \right)}{(1-\alpha)^2 (1-\sqrt{\rho})^2} + \frac{2M \left( \zeta^2 + 4 \kappa^2 \right)}{(1-\alpha)^2 (1-\sqrt{\rho})^2}   \nonumber\\
        & + \frac{16L^2}{(1-\alpha)^2 (1-\sqrt{\rho})} \sum^{t}_{\tau=1} \rho^{(\frac{t+1-\tau}{2})} \sum^M_{i=1} \mathbb{E} \left[ \left\|  x^{[\tau-1]}_i - \bar{x}^{[\tau-1]} \right\|^2 \right]     \nonumber\\
        & + \frac{8M}{(1-\alpha)^2 (1-\sqrt{\rho})} \sum^{t}_{\tau=1} \rho^{(\frac{t+1-\tau}{2})} \mathbb{E} \left[ \left\| \frac{1}{M} \sum^M_{i=1} \nabla f_i \left( x^{[\tau-1]}_i \right) \right\|^2 \right]   
    \end{align}
    where $\scriptsize{\circled{1}}$ follows from \textbf{Fact}~\ref{fact:fact-2}; $\scriptsize{\circled{2}}$ follows from the inequality $\left| \frac{1-\alpha^{t-\tau}}{1-\alpha}\right| \leq \frac{1}{1-\alpha}$ and the inequality $\|\mathbf{AB}\|_{\mathfrak{F}} \leq \|\mathbf{B}\|_{\mathfrak{s}} \|\mathbf{A}\|_{\mathfrak{F}}$ if matrix $\mathbf{B}$ is symmetric; $\scriptsize{\circled{3}}$ follows from \textbf{Fact}~\ref{fact:fact-1}; $\scriptsize{\circled{4}}$ follows from the inequality $\|\mathbf{A}\|_{\mathfrak{F}} \|\mathbf{B}\|_{\mathfrak{F}} \leq \frac{1}{2}(\|\mathbf{A}\|^2_{\mathfrak{F}} + \|\mathbf{B}\|^2_{\mathfrak{F}})$; $\scriptsize{\circled{5}}$ follows from the inequality $\sum^{t}_{\tau'=1} \rho^{\left( t+1-\frac{\tau + \tau'}{2} \right)} = \rho^{\frac{t+1-\tau}{2}} (1-\sqrt{\rho})^{-1} $; $\scriptsize{\circled{6}}$ follows from \textbf{Lemma}~\ref{lem:barG-G}.

    We then bound term $\mathrm{\MakeTextUppercase{\romannumeral 2}}$ as follows
    \begin{align} \label{eq:bdtermII-00}
        & \mathbb{E} \left[ \left\| \sum^{t}_{\tau=1} \frac{1-\alpha^{t-\tau}}{1-\alpha} \left( \mathbf{G}^{[\tau]} (\mathbf{I} - \mathbf{Q}) \mathbf{W}^{t-\tau+1} \right) \right\|^2_{\mathfrak{F}} \right] \nonumber\\
        \leq & \sum^{t}_{\tau=1} \sum^{t}_{\tau'=1} \mathbb{E} \left[ \left\| \frac{1-\alpha^{t-\tau}}{1-\alpha} \left( \mathbf{G}^{[\tau]} (\mathbf{I} - \mathbf{Q}) \mathbf{W}^{t-\tau+1} \right) \right\|_{\mathfrak{F}} \left\| \frac{1-\alpha^{t-\tau'}}{1-\alpha} \left( \mathbf{G}^{[\tau']} (\mathbf{I} - \mathbf{Q}) \mathbf{W}^{t-\tau'+1} \right) \right\|_{\mathfrak{F}} \right]    \nonumber\\
        \leq & \frac{1}{(1-\alpha)^2} \sum^{t}_{\tau=1} \sum^{t}_{\tau'=1} \rho^{(t+1-\frac{\tau + \tau'}{2})} \mathbb{E} \left[ \left\| \mathbf{G}^{[\tau]} \right\|_{\mathfrak{F}} \left\| \mathbf{G}^{[\tau']} \right\|_{\mathfrak{F}} \right]    \nonumber\\
        \leq & \frac{1}{(1-\alpha)^2} \sum^{t}_{\tau=1} \sum^{t}_{\tau'=1} \rho^{(t+1-\frac{\tau + \tau'}{2})} \left( \frac{1}{2} \mathbb{E} \left[ \left\| \mathbf{G}^{[\tau]} \right\|^2_{\mathfrak{F}} \right] +     \frac{1}{2} \mathbb{E} \left[ \left\| \mathbf{G}^{[\tau']} \right\|^2_{\mathfrak{F}} \right] \right)    \nonumber\\
        \leq & \frac{1}{(1-\alpha)^2} \sum^{t}_{\tau=1} \sum^{t}_{\tau'=1} \rho^{(t+1-\frac{\tau + \tau'}{2})} \mathbb{E} \left[ \left\| \mathbf{G}^{[\tau]} \right\|^2_{\mathfrak{F}} \right]    \nonumber\\
        \leq & \frac{1}{(1-\alpha)^2 (1-\sqrt{\rho})} \sum^{t}_{\tau=1} \rho^{(\frac{t+1-\tau}{2})} \mathbb{E} \left[ \left\| \mathbf{G}^{[\tau]} \right\|^2_{\mathfrak{F}} \right]
    \end{align}
    It is apparent that the bound of $\mathrm{\MakeTextUppercase{\romannumeral 2}}$ is dependent on $\mathbb{E} \left[ \left\| \mathbf{G}^{[\tau]} \right\|^2_{\mathfrak{F}} \right]$. Therefore, we have
    \begin{align} \label{eq:bdtermII-01}
        \mathbb{E} \left[ \left\| \mathbf{G}^{[\tau]} \right\|^2_{\mathfrak{F}} \right] = & \mathbb{E} \left[ \left\| \mathbf{G}^{[\tau]} - \mathbf{H}^{[\tau]} + \mathbf{H}^{[\tau]} -   \mathbf{J}^{[\tau]} + \mathbf{J}^{[\tau]} - \mathbf{J}^{[\tau]}\mathbf{Q} + \mathbf{J}^{[\tau]}\mathbf{Q} - \mathbf{H}^{[\tau]}\mathbf{Q} + \mathbf{H}^{[\tau]}\mathbf{Q} \right\|^2_{\mathfrak{F}} \right]    \nonumber\\
        \leq & 5 \mathbb{E} \left[ \left\| \mathbf{G}^{[\tau]} - \mathbf{H}^{[\tau]} \right\|^2_{\mathfrak{F}} \right] 
        + 5\mathbb{E} \left[ \left\| \mathbf{H}^{[\tau]} - \mathbf{J}^{[\tau]} \right\|^2_{\mathfrak{F}} \right] + 5\mathbb{E} \left[ \left\| \mathbf{J}^{[\tau]} - \mathbf{J}^{[\tau]}\mathbf{Q} \right\|^2_{\mathfrak{F}} \right]     \nonumber\\
        & + 5\mathbb{E} \left[ \left\| \mathbf{J}^{[\tau]}\mathbf{Q} - \mathbf{H}^{[\tau]}\mathbf{Q}  \right\|^2_{\mathfrak{F}} \right] + 5\mathbb{E} \left[ \left\| \mathbf{H}^{[\tau]}\mathbf{Q}  \right\|^2_{\mathfrak{F}} \right]    \nonumber\\
        \leq & 5 \mathbb{E} \left[\sum^M_{i=1} \left\| \nabla F_i \left( x^{[\tau-1]}_i; \xi_{i,\tau} \right) - \nabla f_i \left( x^{[\tau-1]}_i \right) \right\|^2 \right]   \nonumber\\
        & + 5 \mathbb{E} \left[\sum^M_{i=1} \left\| \nabla f_i \left( x^{[\tau-1]}_i \right) - \nabla f_i \left( \bar{x}^{[\tau-1]} \right) \right\|^2 \right]   \nonumber\\
        & + 5 \mathbb{E} \left[\sum^M_{i=1} \left\| \nabla f_i \left(\bar{x}^{[\tau-1]} \right) - \frac{1}{M}\sum^M_{j=1} \nabla f_j \left( \bar{x}^{[\tau-1]} \right) \right\|^2 \right]  \nonumber\\
        & + 5 \mathbb{E} \left[\sum^M_{i=1} \left\| \frac{1}{M} \sum^M_{j=1} \nabla f_j \left( \bar{x}^{[\tau-1]} \right) - \frac{1}{M} \sum^M_{j=1} \nabla f_j \left( {x}^{[\tau-1]}_j \right) \right\|^2 \right]  \nonumber\\
        & + 5 \mathbb{E} \left[\sum^M_{i=1} \left\| \frac{1}{M} \sum^M_{j=1} \nabla f_j \left( {x}^{[\tau-1]}_j \right) \right\|^2 \right]  \nonumber\\
        \stackrel{\scriptsize{\circled{1}}} \leq & 5M\zeta^2 + 5M\kappa^2 + 10L^2\sum^M_{i=1} \mathbb{E} \left[ \left\|  x^{[\tau-1]}_i - \bar{x}^{[\tau-1]} \right\|^2 \right]   \nonumber\\
        &  + 5M\mathbb{E} \left[ \left\| \frac{1}{M} \sum^M_{i=1} \nabla f_i \left( x^{[\tau-1]}_i \right) \right\|^2 \right] %
    \end{align} 
    where $\scriptsize{\circled{1}}$ follows from \textbf{Assumption}~\ref{ass:bd-varia}, and
    \begin{align*}
        \sum^M_{i=1} \left\| \nabla f_i(x^{[\tau-1]}_i) - \nabla f_i(\bar{x}^{[\tau-1]}) \right\|^2 \leq L^2 \sum^M_{i=1} \left\| x^{[\tau-1]}_i - \bar{x}^{[\tau-1]} \right\|^2
    \end{align*}
    due to the smoothness of each $f_i$ by \textbf{Assumption}~\ref{ass:L-smooth}, and 
    \begin{align*}
        & \sum^M_{i=1} \left\| \frac{1}{M} \sum^M_{j=1} \nabla f_j(\bar{x}^{[\tau-1]}) - \frac{1}{M} \sum^M_{j=1} \nabla f_j({x}^{[\tau-1]}_j) \right\|^2 \\
        \leq & \sum^M_{i=1} \frac{1}{M} \sum^M_{j=1} \left\| \nabla f_j(\bar{x}^{[\tau-1]}) - \nabla f_j({x}^{[\tau-1]}_j) \right\|^2 \leq L^2 \sum^M_{j=1} \left\| \bar{x}^{[\tau-1]} - {x}^{[\tau-1]}_j \right\|^2
    \end{align*}
    where the first inequality holds due to the the convexity of $\|\cdot\|^2$ and \textit{Jensen's Inequality}.
    By substituting (\ref{eq:bdtermII-01}) into (\ref{eq:bdtermII-00}), we continue to derive the bound of $\mathbf{\MakeTextUppercase{\romannumeral 2}}$ as follows,
    \begin{align} \label{eq:bdtermII-02}
        & \mathbb{E} \left[ \left\| \sum^{t}_{\tau=1} \frac{1-\alpha^{t-\tau}}{1-\alpha} \left( \mathbf{G}^{[\tau]} (\mathbf{I} - \mathbf{Q}) \mathbf{W}^{t-\tau+1} \right) \right\|^2_{\mathfrak{F}} \right]    \nonumber\\
        \leq & \frac{1}{(1-\alpha)^2 (1-\sqrt{\rho})} \sum^{t}_{\tau=1} \rho^{(\frac{t+1-\tau}{2})} \left( \begin{aligned}
        & 5M\zeta^2 + 5M\kappa^2 + 10L^2\sum^M_{i=1} \mathbb{E} \left[ \left\| x^{[\tau-1]}_i - \bar{x}^{[\tau-1]} \right\|^2 \right]   \\
        & + 5M\mathbb{E} \left[ \left\| \frac{1}{M} \sum^M_{i=1} \nabla f_i \left( x^{[\tau-1]}_i \right) \right\|^2 \right]
        \end{aligned}  \right)   \nonumber\\
        \leq & \frac{ 5M \left(\zeta^2 + \kappa^2 \right)}{(1-\alpha)^2 (1-\sqrt{\rho})^2}  
        + \frac{10L^2}{(1-\alpha)^2 (1-\sqrt{\rho})} \sum^{t}_{\tau=1} \rho^{(\frac{t+1-\tau}{2})} \sum^M_{i=1} \mathbb{E} \left[ \left\|  x^{[\tau-1]}_i - \bar{x}^{[\tau-1]} \right\|^2 \right]     \nonumber\\
        & + \frac{5M}{(1-\alpha)^2 (1-\sqrt{\rho})} \sum^{t}_{\tau=1} \rho^{(\frac{t+1-\tau}{2})} \mathbb{E} \left[ \left\| \frac{1}{M} \sum^M_{i=1} \nabla f_i \left(x^{[\tau-1]}_i \right) \right\|^2 \right] 
    \end{align}
    By substituting (\ref{eq:bdtermI}) and (\ref{eq:bdtermII-02}) into (\ref{eq:XI-Q-00}), we have
    \begin{align} \label{eq:XI-Q-01}
        \mathbb{E} \left[ \left\| \mathbf{X}^{[t]} (\mathbf{I} - \mathbf{Q}) \right\|^2_{\mathfrak{F}} \right]
        \leq & \frac{ \frac{8\gamma^2 M}{\omega^4_{\min}} \left( C^2 + \sigma^2 d \right)}{(1-\alpha)^2 (1-\sqrt{\rho})^2} + \frac{4\gamma^2 M \left( \zeta^2 + 4 \kappa^2 \right)}{(1-\alpha)^2 (1-\sqrt{\rho})^2}   \nonumber\\
        & + \frac{32\gamma^2 L^2}{(1-\alpha)^2 (1-\sqrt{\rho})} \sum^{t}_{\tau=1} \rho^{(\frac{t+1-\tau}{2})} \sum^M_{i=1} \mathbb{E} \left[ \left\|  x^{[\tau-1]}_i - \bar{x}^{[\tau-1]} \right\|^2 \right]     \nonumber\\
        & + \frac{16\gamma^2 M}{(1-\alpha)^2 (1-\sqrt{\rho})} \sum^{t}_{\tau=1} \rho^{(\frac{t+1-\tau}{2})} \mathbb{E} \left[ \left\| \frac{1}{M} \sum^M_{i=1} \nabla f_i \left( x^{[\tau-1]}_i \right) \right\|^2 \right]     \nonumber\\
        & + \frac{ 10\gamma^2 M \left(\zeta^2 + \kappa^2 \right)}{(1-\alpha)^2 (1-\sqrt{\rho})^2}  
        + \frac{20\gamma^2 L^2}{(1-\alpha)^2 (1-\sqrt{\rho})} \sum^{t}_{\tau=1} \rho^{(\frac{t+1-\tau}{2})} \sum^M_{i=1} \mathbb{E} \left[ \left\|  x^{[\tau-1]}_i - \bar{x}^{[\tau-1]} \right\|^2 \right]     \nonumber\\
        & + \frac{10\gamma^2 M}{(1-\alpha)^2 (1-\sqrt{\rho})} \sum^{t}_{\tau=1} \rho^{(\frac{t+1-\tau}{2})} \mathbb{E} \left[ \left\| \frac{1}{M} \sum^M_{i=1} \nabla f_i \left(x^{[\tau-1]}_i \right) \right\|^2 \right]     \nonumber\\
        \leq & \frac{ \frac{8\gamma^2 M}{\omega^4_{\min}} \left( C^2 + \sigma^2 d \right)}{(1-\alpha)^2 (1-\sqrt{\rho})^2} + \frac{\gamma^2 M \left( 14\zeta^2 + 26\kappa^2 \right)}{(1-\alpha)^2 (1-\sqrt{\rho})^2}   \nonumber\\
        & + \frac{52\gamma^2 L^2}{(1-\alpha)^2 (1-\sqrt{\rho})} \sum^{t}_{\tau=1} \rho^{\left( \frac{t+1-\tau}{2} \right)} \sum^M_{i=1} \mathbb{E} \left[ \left\| x^{[\tau-1]}_i - \bar{x}^{[\tau-1]} \right\|^2 \right] \nonumber\\
        & + \frac{26\gamma^2 M}{(1-\alpha)^2 (1-\sqrt{\rho})} \sum^{t}_{\tau=1} \rho^{\left( \frac{t+1-\tau}{2} \right)} \mathbb{E} \left[ \left\| \frac{1}{M} \sum^M_{i=1} \nabla f_i \left( x^{[\tau-1]}_i \right) \right\|^2 \right]  
    \end{align}
    Summing over $t = 1, 2, \cdots, T$ and noting that $\mathbb{E} \left[ \left\| \mathbf{X}^{[0]} (\mathbf{I} - \mathbf{Q}) \right\|^2_{\mathfrak{F}} \right] = 0$, we have
    \begin{align} \label{eq:XI-Q-02}
        \sum^{T}_{t=1} \mathbb{E} \left[ \left\| \mathbf{X}^{[t]} (\mathbf{I} - \mathbf{Q}) \right\|^2_{\mathfrak{F}} \right]
        \leq & T \frac{ \frac{8\gamma^2 M}{\omega^4_{\min}} \left( C^2 + \sigma^2 d \right)}{(1-\alpha)^2 (1-\sqrt{\rho})^2} + T \frac{\gamma^2 M \left( 14\zeta^2 + 26\kappa^2 \right)}{(1-\alpha)^2 (1-\sqrt{\rho})^2}  \nonumber\\
        & + \frac{52\gamma^2 L^2}{(1-\alpha)^2 (1-\sqrt{\rho})} \sum^{T}_{t=1} \sum^{t}_{\tau=1} \rho^{(\frac{t+1-\tau}{2})} \sum^M_{i=1} \mathbb{E} \left[ \left\|  x^{[\tau-1]}_i - \bar{x}^{[\tau-1]} \right\|^2 \right]      \nonumber\\
        & + \frac{26\gamma^2 M}{(1-\alpha)^2 (1-\sqrt{\rho})} \sum^{T}_{t=1} \sum^{t}_{\tau=1} \rho^{(\frac{t+1-\tau}{2})} \mathbb{E} \left[ \left\| \frac{1}{M} \sum^M_{i=1} \nabla f_i \left( x^{[\tau-1]}_i \right) \right\|^2 \right]   \nonumber\\
        \leq & T \frac{ \frac{8\gamma^2 M}{\omega^4_{\min}} \left( C^2 + \sigma^2 d \right)}{(1-\alpha)^2 (1-\sqrt{\rho})^2} + T \frac{\gamma^2 M \left( 14\zeta^2 + 26\kappa^2 \right)}{(1-\alpha)^2 (1-\sqrt{\rho})^2}  \nonumber\\
        & + \frac{52\gamma^2 L^2}{(1-\alpha)^2 (1-\sqrt{\rho})^2} \sum^{T}_{t=1} \sum^M_{i=1} \mathbb{E} \left[ \left\|  x^{[t]}_i - \bar{x}^{[t]} \right\|^2 \right]     \nonumber\\
        & + \frac{26\gamma^2 M}{(1-\alpha)^2 (1-\sqrt{\rho})^2} \sum^{T}_{t=1} \mathbb{E} \left[ \left\| \frac{1}{M} \sum^M_{i=1} \nabla f_i \left( x^{[t-1]}_i \right) \right\|^2 \right] 
    \end{align}
    Dividing the both sides of (\ref{eq:XI-Q-02}) by $M$, we have
    \begin{align} \label{eq:XI-Q-03}
        \sum^{T}_{t=1} \frac{1}{M} \mathbb{E} \left[ \left\| \mathbf{X}^{[t]} (\mathbf{I} - \mathbf{Q}) \right\|^2_{\mathfrak{F}} \right]
        \leq & T \frac{ \frac{8\gamma^2 }{\omega^4_{\min}} \left( C^2 + \sigma^2 d \right)}{(1-\alpha)^2 (1-\sqrt{\rho})^2} + T \frac{\gamma^2  \left( 14\zeta^2 + 26\kappa^2 \right)}{(1-\alpha)^2 (1-\sqrt{\rho})^2}   \nonumber\\
        & + \frac{52\gamma^2 L^2}{(1-\alpha)^2 (1-\sqrt{\rho})^2} \sum^{T}_{t=1} \frac{1}{M} \mathbb{E} \left[ \left\| \mathbf{X}^{[t]} (\mathbf{I} - \mathbf{Q}) \right\|^2_{\mathfrak{F}} \right]   \nonumber\\
        & + \frac{26\gamma^2}{(1-\alpha)^2 (1-\sqrt{\rho})^2} \sum^{T}_{t=1} \mathbb{E} \left[ \left\| \frac{1}{M} \sum^M_{i=1} \nabla f_i(x^{[t-1]}_i) \right\|^2 \right]  
    \end{align}
    Rearranging terms and dividing the both sides of (\ref{eq:XI-Q-03}) by $1- \frac{52\gamma^2 L^2}{(1-\alpha)^2 (1-\sqrt{\rho})^2}$, we complete the proof as follows
    \begin{align} \label{eq:lem6final}
        & \sum^{T}_{t=1} \frac{1}{M} \sum^M_{i=1} \mathbb{E} \left[ \left\| \bar{x}^{[t]} - x^{[t]}_i \right\|^2 \right]  \nonumber\\
        \leq & \frac{1}{1- \frac{52\gamma^2 L^2 }{(1-\alpha)^2 (1-\sqrt{\rho})^2}} \left( T \frac{ \frac{8\gamma^2 }{\omega^4_{\min}} \left( C^2 + \sigma^2 d \right)}{(1-\alpha)^2 (1-\sqrt{\rho})^2} + T \frac{\gamma^2  \left( 14\zeta^2 + 26\kappa^2 \right)}{(1-\alpha)^2 (1-\sqrt{\rho})^2} \right)  \nonumber\\
        & + \frac{1}{1- \frac{52\gamma^2 L^2 }{(1-\alpha)^2 (1-\sqrt{\rho})^2}} \left( \frac{26\gamma^2}{(1-\alpha)^2 (1-\sqrt{\rho})^2} \sum^{T}_{t=1} \mathbb{E} \left[ \left\| \frac{1}{M} \sum^M_{i=1} \nabla f_i \left(x^{[t-1]}_i \right) \right\|^2 \right] \right)  \nonumber\\
        \stackrel{\scriptsize{\circled{1}}} \leq & T \frac{ \frac{16\gamma^2 }{\omega^4_{\min}} \left( C^2 + \sigma^2 d \right)}{(1-\alpha)^2 (1-\sqrt{\rho})^2} + T \frac{4 \gamma^2  \left( 7\zeta^2 + 13\kappa^2 \right)}{(1-\alpha)^2 (1-\sqrt{\rho})^2}  
        + \frac{52\gamma^2}{(1-\alpha)^2 (1-\sqrt{\rho})^2} \sum^{T}_{t=1} \mathbb{E} \left[ \left\| \frac{1}{M} \sum^M_{i=1} \nabla f_i \left( x^{[t-1]}_i \right) \right\|^2 \right] 
    \end{align}
    where $\scriptsize{\circled{1}}$ follows because $\gamma \leq \frac{(1-\alpha)(1-\sqrt{\rho})}{2\sqrt{26} L}$ is chosen to ensure $\left( 1 - \frac{52\gamma^2 L^2}{(1-\alpha)^2 (1-\sqrt{\rho})^2} \right)^{-1} \leq 2$.
 \end{proof}

 According to the above \textbf{Lemmas}, we now give the main proof of \textbf{Theorem}~\ref{thm:convergence-main}.
 Based on the $L$-smoothness of $\mathcal{F}$, we have
 \begin{align} \label{eq:fbars}
    \mathbb{E} \left[ \mathcal{F} \left(\bar{S}^{[t]} \right) \right]
    \leq \mathbb{E} \left[ \mathcal{F} \left(\bar{S}^{[t-1]} \right) \right] + \mathbb{E} \left[ \left\langle \nabla \mathcal{F} \left(\bar{S}^{[t-1]} \right), \bar{S}^{[t]} - \bar{S}^{[t-1]} \right\rangle \right] + \frac{L}{2} \mathbb{E} \left[ \left\| \bar{S}^{[t]} - \bar{S}^{[t-1]} \right\|^2 \right]
 \end{align}
 Based on \textbf{Lemma}~\ref{lem:seqdiff}, we have
 \begin{align} \label{eq:bars-bars}
    \mathbb{E} \left[ \left\langle \nabla \mathcal{F} \left( \bar{S}^{[t-1]} \right), \bar{S}^{[t]} - \bar{S}^{[t-1]} \right\rangle \right]
    = & \frac{-\gamma}{1-\alpha} \mathbb{E} \left[ \left\langle \nabla \mathcal{F} \left( \bar{S}^{[t-1]} \right),\frac{1}{M}\sum^M_{i=1} \bar{g}_i^{[t]} \right\rangle \right]  \nonumber\\
    = & \frac{-\gamma}{1-\alpha} \mathbb{E} \left[ \left\langle \nabla \mathcal{F} \left( \bar{S}^{[t-1]} \right) - \nabla \mathcal{F} \left( \bar{x}^{[t-1]} \right) ,\frac{1}{M}\sum^M_{i=1} \bar{g}_i^{[t]} \right\rangle \right]  \nonumber\\
    & + \frac{-\gamma}{1-\alpha} \mathbb{E} \left[ \left\langle \nabla \mathcal{F} \left( \bar{x}^{[t-1]} \right), \frac{1}{M}\sum^M_{i=1} \bar{g}_i^{[t]} \right\rangle \right]  \nonumber\\
    = & \underbrace{ \frac{-\gamma}{1-\alpha} \mathbb{E} \left[ \left\langle \nabla \mathcal{F} \left( \bar{S}^{[t-1]} \right) - \nabla \mathcal{F} \left( \bar{x}^{[t-1]} \right) ,\frac{1}{M}\sum^M_{i=1} \bar{g}_i^{[t]} \right\rangle \right] }_{\mathrm{\MakeTextUppercase{\romannumeral 3}}}  \nonumber\\
    & + \underbrace{ \frac{-\gamma}{1-\alpha} \mathbb{E} \left[ \left\langle \nabla \mathcal{F} \left(\bar{x}^{[t-1]} \right), \frac{1}{M}\sum^M_{i=1} \left( \bar{g}_i^{[t]} - g^{[t]}_{i,i} \right) \right\rangle \right] }_{\mathsf{\MakeTextUppercase{\romannumeral 4}}} \nonumber\\
    & + \underbrace{ \frac{-\gamma}{1-\alpha} \mathbb{E} \left[ \left\langle \nabla \mathcal{F} \left(\bar{x}^{[t-1]} \right), \frac{1}{M}\sum^M_{i=1} g^{[t]}_{i,i} \right\rangle \right] }_{\mathrm{\MakeTextUppercase{\romannumeral 5}}}
 \end{align}
 which can be bounded by bounding $\mathrm{\MakeTextUppercase{\romannumeral 3}}$, $\mathrm{\MakeTextUppercase{\romannumeral 4}}$ and $\mathrm{\MakeTextUppercase{\romannumeral 5}}$. For $\mathrm{\MakeTextUppercase{\romannumeral 3}}$, we have
 \begin{align} \label{eq:term-iii}
    & \frac{-\gamma}{1-\alpha} \mathbb{E} \left[ \left\langle \nabla \mathcal{F} \left( \bar{S}^{[t-1]} \right) - \nabla \mathcal{F} \left(\bar{x}^{[t-1]} \right) ,\frac{1}{M}\sum^M_{i=1} \bar{g}_i^{[t]} \right\rangle \right]  \nonumber\\
    \stackrel{\scriptsize{\circled{1}}} \leq & \frac{1-\alpha}{2{\alpha L}} \mathbb{E} \left[ \left\| \nabla \mathcal{F} \left( \bar{S}^{[t-1]} \right) - \nabla \mathcal{F} \left( \bar{x}^{[t-1]} \right) \right\|^2 \right] + \frac{\alpha L \gamma^2}{2(1-\alpha)^3} \mathbb{E} \left[ \left\| \frac{1}{M}\sum^M_{i=1} \bar{g}_i^{[t]} \right\|^2 \right]  \nonumber\\
    \stackrel{\scriptsize{\circled{2}}} \leq & \frac{(1-\alpha)L}{2{\alpha}} \mathbb{E} \left[ \left\| \bar{S}^{[t-1]} - \bar{x}^{[t-1]} \right\|^2 \right] + \frac{\alpha L \gamma^2}{2(1-\alpha)^3} \mathbb{E} \left[ \left\| \frac{1}{M}\sum^M_{i=1} \bar{g}_i^{[t]} \right\|^2 \right]
 \end{align}
 where $\scriptsize{\circled{1}}$ follows from the basic inequality $\langle a,b \rangle \leq \frac{1}{2} \|a\|^2 + \frac{1}{2} \|b\|^2$ with $a = \frac{\sqrt{1-\alpha}}{\sqrt{\alpha L}} \left( \nabla \mathcal{F} \left( \bar{S}^{[t-1]} \right) - \nabla \mathcal{F} \left( \bar{x}^{[t-1]} \right) \right)$ and $b = \frac{-\gamma \sqrt{\alpha L}}{(1-\alpha)^{\frac{3}{2}}} \left(\frac{1}{M} \sum^M_{i=1} \bar{g}_i^{[t]}\right)$; and $\scriptsize{\circled{2}}$ based on the smoothness of function $\mathcal{F}(\cdot)$.
 For $\mathrm{\MakeTextUppercase{\romannumeral 4}}$, we have
 \begin{align} \label{eq:term-iv}
    & \frac{-\gamma}{1-\alpha} \mathbb{E} \left[ \left\langle \nabla \mathcal{F} \left( \bar{x}^{[t-1]} \right), \frac{1}{M}\sum^M_{i=1} \left( \bar{g}_i^{[t]} - g^{[t]}_{i,i} \right) \right\rangle \right]  \nonumber\\
    \leq & \frac{1-\alpha}{2\alpha} \mathbb{E} \left[ \left\| \nabla \mathcal{F} \left(\bar{x}^{[t-1]} \right) \right\|^2 \right] + \frac{\alpha \gamma^2}{2(1-\alpha)^3} \mathbb{E} \left[ \left\| \frac{1}{M}\sum^M_{i=1} \left( \bar{g}_i^{[t]} - g^{[t]}_{i,i} \right) \right\|^2 \right]  
 \end{align}
 by applying $\langle a,b \rangle \leq \frac{1}{2} \|a\|^2 + \frac{1}{2} \|b\|^2$ with $a = \frac{-\sqrt{(1-\alpha)}}{\sqrt{\alpha}} \nabla \mathcal{F}\left( \bar{x}^{[t-1]} \right) $ and $b = \frac{\gamma \sqrt{ \alpha }}{(1-\alpha)^\frac{3}{2}} \left(\frac{1}{M}\sum^M_{i=1} \left( \bar{g}_i^{[t]} - g^{[t]}_{i,i} \right) \right)$.
 Then, for $\mathrm{\MakeTextUppercase{\romannumeral 5}}$, we have
 \begin{align} \label{eq:term-v-00}
    \mathbb{E} \left[ \left\langle \nabla \mathcal{F} \left(\bar{x}^{[t-1]} \right), \frac{1}{M}\sum^M_{i=1} g^{[t]}_{i,i} \right\rangle \right] = \mathbb{E} \left[ \left\langle \nabla \mathcal{F} \left(\bar{x}^{[t-1]} \right), \frac{1}{M}\sum^M_{i=1}  \nabla f_i \left( x^{[t-1]}_i \right) \right\rangle \right]
 \end{align}
 Therein,
 \begin{align} \label{eq:term-v-01}
    & \left\langle \nabla \mathcal{F} \left( \bar{x}^{[t-1]} \right), \frac{1}{M}\sum^M_{i=1} \nabla f_i \left( x^{[t-1]}_i \right) \right\rangle   \nonumber\\
    \stackrel{\scriptsize{\circled{1}}} = & \frac{1}{2} \left( \left\| \nabla \mathcal{F} \left( \bar{x}^{[t-1]} \right) \right\|^2 + \left\| \frac{1}{M}\sum^M_{i=1} \nabla f_i \left( x^{[t-1]}_i \right) \right\|^2 - \left\| \nabla \mathcal{F} \left(\bar{x}^{[t-1]} \right) - \frac{1}{M}\sum^M_{i=1}  \nabla f_i \left( x^{[t-1]}_i \right) \right\|^2 \right)  \nonumber\\
    \stackrel{\scriptsize{\circled{2}}} \geq & \frac{1}{2} \left( \left\| \nabla \mathcal{F} \left( \bar{x}^{[t-1]} \right) \right\|^2 + \left\| \frac{1}{M}\sum^M_{i=1} \nabla f_i \left( x^{[t-1]}_i \right) \right\|^2 - \frac{L^2}{M} \sum^M_{i=1} \left\| \bar{x}^{[t-1]} - x^{[t-1]}_i \right\|^2 \right)
 \end{align}
 where we have $\scriptsize{\circled{1}}$ according to $\langle a, b \rangle = \frac{1}{2} \left( \|a\|^2 + \|b\|^2 - \|a-b\|^2 \right)$ with $a=\nabla \mathcal{F}(\bar{x}^{[t-1]})$ and $b=\frac{1}{M}\sum^M_{i=1} \nabla f_i( x^{[t-1]}_i)$; $\scriptsize{\circled{2}}$ since
 \begin{align*}
    \left\| \nabla \mathcal{F}(\bar{x}^{[t-1]}) - \frac{1}{M}\sum^M_{i=1} \nabla f_i(x^{[t-1]}_i) \right\|^2 = & \left\| \frac{1}{M} \sum^M_{i=1} \nabla f_i(\bar{x}^{[t-1]}) - \frac{1}{M}\sum^M_{i=1} \nabla f_i(x^{[t-1]}_i) \right\|^2 \nonumber\\
    \stackrel{\scriptsize{\circled{1}}} \leq & \frac{1}{M} \sum^M_{i=1} \left\| \nabla f_i(\bar{x}^{[t-1]}) -  \nabla f_i(x^{[t-1]}_i) \right\|^2 \nonumber\\
    \stackrel{\scriptsize{\circled{2}}} \leq & \frac{1}{M} \sum^M_{i=1} L^2 \left\| \bar{x}^{[t-1]} - x^{[t-1]}_i \right\|^2
 \end{align*}
 where ${\scriptsize{\circled{1}}}$ due to the convexity of $\|\cdot\|^2$ and \textit{Jensen's Inequality}; and ${\scriptsize{\circled{2}}}$ according to the smoothness of $f_i(\cdot)$.
 By substituting (\ref{eq:term-v-01}) into (\ref{eq:term-v-00}), we obtain the bound of $\mathrm{\MakeTextUppercase{\romannumeral 5}}$ as follows
 \begin{align} \label{eq:term-v-02}
    \frac{-\gamma}{1-\alpha} \mathbb{E} \left[ \left\langle \nabla \mathcal{F} \left(\bar{x}^{[t-1]} \right), \frac{1}{M}\sum^M_{i=1} g^{[t]}_{i,i} \right\rangle \right] 
    \leq & \frac{-\gamma}{2(1-\alpha)} \mathbb{E} \left[ \left\| \nabla \mathcal{F} \left(\bar{x}^{[t-1]} \right) \right\|^2 \right]   \nonumber\\
    & + \frac{-\gamma}{2(1-\alpha)} \mathbb{E} \left[ \left\| \frac{1}{M}\sum^M_{i=1} \nabla f_i \left(x^{[t-1]}_i \right) \right\|^2 \right]  \nonumber\\
    & - \frac{-\gamma L^2}{2M(1-\alpha)} \sum^M_{i=1} \mathbb{E} \left[ \left\| \bar{x}^{[t-1]} - x^{[t-1]}_i \right\|^2 \right]
 \end{align}
 By combining the bounds of $\mathrm{\MakeTextUppercase{\romannumeral 3}}$, $\mathrm{\MakeTextUppercase{\romannumeral 4}}$, and $\mathrm{\MakeTextUppercase{\romannumeral 5}}$ (see (\ref{eq:term-iii}), (\ref{eq:term-iv}) and (\ref{eq:term-v-02}), respectively), we obtain the bound of (\ref{eq:bars-bars}), then we continue to bound $\mathbb{E} \left[ \mathcal{F} \left(\bar{S}^{[t]} \right) \right]$ based on (\ref{eq:fbars}) as follows
 \begin{align} \label{eq:fbarx}
    \mathbb{E} \left[ \mathcal{F} \left(\bar{S}^{[t]} \right) \right]  
    \leq & \mathbb{E} \left[ \mathcal{F} \left(\bar{S}^{[t-1]} \right) \right] + \frac{(1-\alpha)L}{2{\alpha}} \mathbb{E} \left[ \left\| \bar{S}^{[t-1]} - \bar{x}^{[t-1]} \right\|^2 \right] \nonumber\\
    & + \left( \frac{\alpha L \gamma^2}{2(1-\alpha)^3} + \frac{L\gamma^2}{2(1-\alpha)^2} \right) \mathbb{E} \left[ \left\| \frac{1}{M}\sum^M_{i=1} \bar{g_i}^{[t]} \right\|^2 \right] \nonumber\\
    & + \left( \frac{1-\alpha}{2\alpha} - \frac{\gamma}{2(1-\alpha)} \right) \mathbb{E} \left[ \left\| \nabla \mathcal{F} \left(\bar{x}^{[t-1]} \right) \right\|^2 \right]   \nonumber\\
    & + \frac{\alpha \gamma^2}{2(1-\alpha)^3} \mathbb{E} \left[ \left\| \frac{1}{M}\sum^M_{i=1} \left( \bar{g_i}^{[t]} - g^{[t]}_{i,i} \right) \right\|^2 \right]  \nonumber\\
    & - \frac{\gamma}{2(1-\alpha)} \mathbb{E} \left[ \left\| \frac{1}{M}\sum^M_{i=1}  \nabla f_i \left(x^{[t-1]}_i \right) \right\|^2 \right] \nonumber\\
    & + \frac{\gamma L^2}{2(1-\alpha)} \frac{1}{M} \sum^M_{i=1} \mathbb{E} \left[ \left\| \bar{x}^{[t-1]} - x^{[t-1]}_i \right\|^2 \right]  
 \end{align}
 By rearranging (\ref{eq:fbarx}), we have
 \begin{align} \label{eq:fbarx-01}
    \mathbb{E} \left[ \left\| \nabla \mathcal{F} \left( \bar{x}^{[t-1]} \right) \right\|^2 \right]
    \leq & \frac{1}{m_1} \left( \mathbb{E} \left[ \mathcal{F}\left(\bar{S}^{[t-1]}\right) \right] - \mathbb{E} \left[ \mathcal{F}\left(\bar{S}^{[t]}\right) \right] \right) \nonumber\\
    & + m_2 \mathbb{E} \left[ \left\| \frac{1}{M}\sum^M_{i=1} \bar{g_i}^{[t]} \right\|^2 \right] + m_3 \mathbb{E} \left[ \left\| \bar{S}^{[t-1]} - \bar{x}^{[t-1]} \right\|^2 \right]  \nonumber\\
    & + m_4  \mathbb{E} \left[ \left\| \frac{1}{M}\sum^M_{i=1} \left( \bar{g_i}^{[t]} - g^{[t]}_{i,i} \right) \right\|^2 \right]  \nonumber\\
    & + \frac{m_5}{M} \sum^M_{i=1} \mathbb{E} \left[ \left\| \bar{x}^{[t-1]} - x^{[t-1]}_i \right\|^2  \right] \nonumber\\
    & - m_6 \mathbb{E} \left[ \left\| \frac{1}{M}\sum^M_{i=1} \nabla f_i \left( x^{[t-1]}_i \right) \right\|^2 \right] 
 \end{align}
 where $m_1 = \frac{\gamma}{2(1-\alpha)} - \frac{1-\alpha}{2\alpha}$, $m_2 = \frac{1}{m_1} \left( \frac{\alpha L \gamma^2}{2(1-\alpha)^3} + \frac{L \gamma^2}{2(1-\alpha)^2} \right)$, $m_3 = \frac{L(1-\alpha)}{2 m_1 \alpha}$, $m_4 = \frac{\alpha \gamma^2}{2m_1(1-\alpha)^3} $, $m_5 = \frac{L^2 \gamma}{2 m_1 (1-\alpha)}$, $m_6 = \frac{\gamma}{2 m_1 (1-\alpha)}$.
 Summing over $t = 1, \cdots, T$, we have
 \begin{align} \label{eq:sumfbarx-00}
    \sum^{T}_{t=1} \mathbb{E} \left[ \left\| \nabla \mathcal{F}(\bar{x}^{[t-1]}) \right\|^2 \right]
    \leq & \frac{1}{m_1} \left( \mathbb{E} \left[ \mathcal{F} \left( \bar{S}^{[0]} \right) \right] - \mathbb{E} \left[ \mathcal{F}(\bar{S}^{[T]}) \right] \right) \nonumber\\
    & + m_2 \sum^{T}_{t=1} \mathbb{E} \left[ \left\| \frac{1}{M}\sum^M_{i=1} \bar{g_i}^{[t]} \right\|^2 \right] \nonumber\\
    & + m_3 \sum^{T}_{t=1} \mathbb{E} \left[ \left\| \bar{S}^{[t-1]} - \bar{x}^{[t-1]} \right\|^2 \right] \nonumber\\
    & + m_4 \sum^{T}_{t=1} \mathbb{E} \left[ \left\| \frac{1}{M}\sum^M_{i=1} \left( \bar{g}_i^{[t]} - g^{[t]}_{i,i} \right) \right\|^2 \right] \nonumber\\
    & + m_5 \sum^{T}_{t=1} \frac{1}{M} \sum^M_{i=1} \mathbb{E} \left[ \left\| \bar{x}^{[t-1]} - x^{[t-1]}_i \right\|^2  \right] \nonumber\\
    & - m_6 \sum^{T}_{t=1} \mathbb{E} \left[ \left\| \frac{1}{M}\sum^M_{i=1} \nabla f_i \left( x^{[t-1]}_i \right) \right\|^2 \right]
 \end{align}
 By combining $\textbf{Lemma}~\ref{lem:bdavggrad}$, $\textbf{Lemma}~\ref{lem:diffhatgrad}$, $\textbf{Lemma}~\ref{lem:sandx}$ and $\textbf{Lemma}~\ref{lem:barx-x}$ in (\ref{eq:sumfbarx-00}), we have,
 \begin{align} \label{eq:sumfbarx-01}
    & \sum^{T}_{t=1} \mathbb{E} \left[ \left\| \nabla \mathcal{F}(\bar{x}^{[t-1]}) \right\|^2 \right]  \nonumber\\
    \leq & \frac{1}{m_1} \left( \mathbb{E} \left[ \mathcal{F} \left( \bar{S}^{[0]} \right) \right] - \mathbb{E} \left[ \mathcal{F} \left( \bar{S}^{[T]} \right) \right] \right)  \nonumber\\
    & + T \left( m_2 + m_3 \frac{\gamma^2 \alpha^2}{(1-\alpha)^4} + m_4 \right) \left( \frac{4 C^2}{\omega^4_{\min}} + \frac{4 \sigma^2 d}{\omega^4_{\min}} + \frac{2 \zeta^2}{M} \right) \nonumber\\
    & + T m_5 \left( \frac{ \frac{16\gamma^2 }{\omega^4_{\min}} \left( C^2 + \sigma^2 d \right)}{(1-\alpha)^2 (1-\sqrt{\rho})^2} + \frac{4 \gamma^2  \left( 7\zeta^2 + 13\kappa^2 \right)}{(1-\alpha)^2 (1-\sqrt{\rho})^2} \right) \nonumber\\
    & - \left( m_6 - 2m_4 - m_5 \frac{52\gamma^2}{(1-\alpha)^2 (1-\sqrt{\rho})^2} \right) \sum^{T}_{t=1} \mathbb{E} \left[ \left\| \frac{1}{M} \sum^M_{i=1} \nabla f_i \left( x^{[t-1]}_i \right) \right\|^2 \right] 
 \end{align}
 Dividing both sides of (\ref{eq:sumfbarx-01}) by $T$ and employing the fact that $\bar{S}^{[0]} = \bar{x}^{[0]}$, we have
 \begin{align} \label{eq:finalbound}
    & \frac{1}{T} \sum^{T}_{t=1} \mathbb{E} \left[ \left\| \nabla \mathcal{F} \left( \bar{x}^{[t-1]} \right) \right\|^2 \right] \nonumber\\
    \leq & \frac{1}{m_1 T} \left( \mathcal{F} \left( \bar{x}^{[0]} \right) -  \mathcal{F}^* \right)  \nonumber\\
    & + \left( m_2 + m_3 \frac{\gamma^2 \alpha^2}{(1-\alpha)^4} + m_4 \right) \left( \frac{4 C^2}{\omega^4_{\min}} + \frac{4 \sigma^2 d}{\omega^4_{\min}} + \frac{2 \zeta^2}{M} \right) \nonumber\\
    & + m_5 \left( \frac{ \frac{16\gamma^2 }{\omega^4_{\min}} \left( C^2 + \sigma^2 d \right)}{(1-\alpha)^2 (1-\sqrt{\rho})^2} + \frac{4 \gamma^2 \left( 7\zeta^2 + 13\kappa^2 \right)}{(1-\alpha)^2 (1-\sqrt{\rho})^2} \right)
 \end{align}
 when 
 \begin{equation} \label{eq:condition-gamma}
    m_6 - 2m_4 - m_5 \frac{52\gamma^2}{(1-\alpha)^2 (1-\sqrt{\rho})^2}  \geq 0
 \end{equation}
 To ensure the inequality (\ref{eq:condition-gamma}), we have
 \begin{align}
    m_1 = \frac{\gamma}{2(1-\alpha)} - \frac{1-\alpha}{2\alpha} > 0
 \end{align}
 and
 \begin{align}
    1 - \frac{ 2\alpha \gamma}{(1-\alpha)^2} - \frac{52L^2\gamma^2}{(1-\alpha)^2 (1-\sqrt{\rho})^2} \geq 0
 \end{align}
 therefore, the step size $\gamma$ can be defined as
    \begin{align}
      \frac{(1-\alpha)^2}{\alpha} < \gamma \leq \min \left\{ \frac{(1-\alpha)(1-\sqrt{\rho})}{2\sqrt{26}L}, \frac{(1-\sqrt{\rho})\sqrt{52L^2(1-\alpha)^2 + \alpha^2(1-\sqrt{\rho})^2}}{52L^2} - \frac{\alpha(1-\sqrt{\rho})^2}{52L^2} \right\}
    \end{align}

 We continue to present the proof of \textbf{Corollary}~\ref{cor:main}. If letting $\gamma = \mathcal{O}\left(\sqrt{\frac{1}{T}}\right)$, we have the constants in (\ref{eq:main-constants}) represented by
 \begin{align}
      m_1 = \mathcal{O}\left(\sqrt{\frac{1}{T}}\right),~
      m_2 = \mathcal{O}\left(\sqrt{\frac{1}{T}}\right),~ 
      m_3 = \mathcal{O}\left(\sqrt{T}\right),~ 
      m_4 = \mathcal{O}\left(\sqrt{\frac{1}{T}}\right),~ 
      m_5 = \mathcal{O}\left(1\right),~
      \text{and}~m_6 = \mathcal{O}\left(1\right)
 \end{align}
 Then, for each term in the right hand side of (\ref{eq:finalbound}), we have
 \begin{align}
     \frac{1}{m_1 T} \left( \mathcal{F} \left( \bar{x}^{[0]} \right) -  \mathcal{F}^* \right) = \mathcal{O}\left(\frac{1}{\sqrt{T}}\right), 
 \end{align}
 \begin{align}
     \left( m_2 + \frac{ m_3 \gamma^2 \alpha^2}{(1-\alpha)^4} + m_4 \right) \left( \frac{4 C^2}{\omega^4_{\min}} + \frac{4 \sigma^2 d}{\omega^4_{\min}} + \frac{2 \zeta^2}{M} \right) = \mathcal{O}\left(\frac{1}{\sqrt{T}}\right) +  \mathcal{O}\left(\frac{\sigma^2 d}{\sqrt{T}}\right) + \mathcal{O}\left(\frac{1}{M\sqrt{T}}\right),
 \end{align}
 and
 \begin{align}
     m_5 \left( \frac{ \frac{16\gamma^2 }{\omega^4_{\min}} \left( C^2 + \sigma^2 d \right)}{(1-\alpha)^2 (1-\sqrt{\rho})^2} + \frac{4 \gamma^2  \left( 7\zeta^2 + 13\kappa^2 \right)}{(1-\alpha)^2 (1-\sqrt{\rho})^2} \right) = \mathcal{O}\left(\frac{\sigma^2 d}{T}\right) + \mathcal{O}\left(\frac{1}{T}\right) 
 \end{align}
 by combining which, we can conclude that there exists constant $K$ such that
 \begin{align}
     \frac{1}{T} \sum^{T}_{t=1} \mathbb{E} \left[ \left\| \nabla \mathcal{F} \left( \bar{x}^{[t-1]} \right) \right\|^2 \right] \leq K \left( \frac{1}{\sqrt{T}} + \frac{\sigma^2 d}{\sqrt{T}} + \frac{1}{M\sqrt{T}} + \frac{1}{T} + \frac{\sigma^2 d}{T} \right)
 \end{align}
 If we fix $M$ and $\sigma$, let $T$ be sufficiently large, the convergence rate of \textbf{Algorithm}~\ref{alg:pdsl} can be characterized by $\mathcal{O}\left(\frac{1}{\sqrt{T}}\right)$.

\end{document}